\documentclass[twoside,11pt]{article}

\usepackage[preprint]{jmlr2e}
\usepackage{natbib}
\setcitestyle{round}

\usepackage[utf8]{inputenc} %
\usepackage{xcolor}
\hypersetup{
    colorlinks,
    linkcolor={red!50!black},
    citecolor={blue!50!black},
    urlcolor={blue!80!black}
}

\renewcommand{\paragraph}[1]{\vspace{-0.1em}\noindent \textbf{#1}}

\usepackage{inconsolata}
\usepackage{url}            %
\usepackage{booktabs}       %
\usepackage{amsfonts}       %
\usepackage{nicefrac}       %
\usepackage{microtype}      %
\usepackage{marginalia}
\usepackage{array,amssymb,latexsym,amssymb,epsfig,epsf,amsmath,cite,accents,float,graphicx,natbib,caption,mathrsfs,verbatim,bm,color,enumitem}
\graphicspath{{figures/}}
\usepackage[linesnumbered,noend]{algorithm2e}
\usepackage{appendix}

\SetKwInput{KwEn}{Encoding}
\SetKwInput{KwDe}{Decoding}
\newcommand{\defn}[1]{\textbf{#1}}

\newcommand{\NUQSGD}{\mathrm{{NUQSGD}}}
\newcommand{\QSGD}{\mathrm{QSGD}}
\newcommand{\NUQ}{\mathrm{{NUQ}}}
\newcommand{\QINF}{\mathrm{Qinf}}
\newcommand{\mathbbm}[1]{\text{\usefont{U}{bbm}{m}{n}#1}} %

\newcommand{\ENCODE}{\mathrm{ENCODE}}
\newcommand{\DECODE}{\mathrm{DECODE}}
\newcommand{\Elias}{\mathrm{ERC}}
\newcommand{\level}[1]{\tilde s(#1)}

\def\reals{\mathbb{R}}

\def\integers{\mathbb{Z}}

\def\rhobf{\boldsymbol \rho }

\def\hbf{{\bf h}}

\def\qbf{{\bf q}}

\def\ubf{{\bf u}}
\def\vbf{{\bf v}}
\def\wbf{{\bf w}}

\def\ybf{{\bf y}}
\def\zbf{{\bf z}}

\def\ybf{{\bf y}}

\def\Pbf{{\bf P}}

\def\Dc{{\cal D}}

\def\Ic{{\cal I}}

\def\Lc{{\cal L}}

\def\Pc{{\cal P}}
\def\Qc{{\cal Q}}
\def\Rc{{\cal R}}
\def\Sc{{\cal S}}

\def\nn{\nonumber}
\def\beq{\begin{equation}}
\def\eeq{\end{equation}}
\def\beqa{\begin{eqnarray}}
\def\eeqa{\end{eqnarray}}
\def\balign{\begin{align}}
\def\ealign{\end{align}}
\def\bpr{\begin{proof}}
\def\epr{\end{proof}}
\def\bth{\begin{theorem}}
\def\eth{\end{theorem}}
\def\blm{\begin{lemma}}
\def\elm{\end{lemma}}
\def\bprop{\begin{proposition}}
\def\eprop{\end{proposition}}
\def\bcr{\begin{corollary}}
\def\ecr{\end{corollary}}

\def\ie{{\it i.e.,\ \/}}
\def\eg{{\it e.g.,\ \/}}
\def\defeq{\triangleq}

\def\st {{\rm subject~to~}}
\def\E{\mathbb{E}}

\def\and {{\rm and}}

\firstpageno{1}

\usepackage{lastpage}
\jmlrheading{22}{2021}{1-\pageref{LastPage}}{3/20; Revised
1/21}{4/21}{20-255}{Ali Ramezani-Kebrya, Fartash Faghri, Ilya Markov, Vitalii Aksenov, Dan Alistarh, and Daniel M. Roy}
\ShortHeadings{NUQSGD}{Ramezani-Kebrya, Faghri,  Markov,  Aksenov, Alistarh, and Roy}

\begin{document}

\title{NUQSGD: Provably Communication-efficient Data-parallel SGD via Nonuniform Quantization}

\author{%
       \name Ali Ramezani-Kebrya\thanks{Work performed at the Vector Institute and University of Toronto.} 
       \email ali.ramezani@epfl.ch\\
       \addr \'{E}cole Polytechnique F\'{e}d\'{e}rale de Lausanne\\
       Route Cantonale, 1015 Lausanne, Switzerland
       \AND
       \name Fartash Faghri \email faghri@cs.toronto.edu\\
       \addr Dept.\ of Computer Science, Univ.\ of Toronto and Vector Institute\\
       Toronto, ON M5T 3A1, Canada
       \AND
       \name Ilya Markov \email ilia.markov@ist.ac.at\\
       \addr Institute of Science and Technology Austria\\
       3400 Klosterneuburg, Austria
       \AND
       \name Vitalii Aksenov \email vitalii.aksenov@ist.ac.at\\
       \addr Institute of Science and Technology Austria\\
       3400 Klosterneuburg, Austria
       \AND
       \name Dan Alistarh \email dan.alistarh@ist.ac.at\\
       \addr Institute of Science and Technology Austria\\
       3400 Klosterneuburg, Austria
       \AND
       \name Daniel M. Roy \email daniel.roy@utoronto.ca\\
       \addr Dept.\ of Statistical Sciences, Univ.\ of Toronto and Vector Institute\\
       Toronto, ON M5S 3G3, Canada}           

\editor{Sathiya Keerthi}

\maketitle

\begin{abstract}As the size and complexity of models and datasets grow, so does the need for communication-efficient variants of stochastic gradient descent that can be deployed to perform parallel model training. 
One popular communication-compression method for data-parallel SGD is QSGD~\citep{QSGD}, which quantizes and encodes gradients to reduce communication costs. 
The baseline variant of QSGD provides strong theoretical guarantees, however, for practical purposes, the authors proposed a heuristic variant which we call QSGDinf, which demonstrated impressive empirical gains for distributed training of large neural networks. 
In this paper, we build on this work to propose a new gradient quantization scheme, and show that it has both stronger theoretical guarantees than QSGD, and matches and exceeds the empirical performance of the QSGDinf heuristic and of other compression methods. 
\end{abstract}


\section{Introduction}\label{sec:intro}

Deep learning is booming  thanks to enormous datasets and very large models, leading to the fact that the largest datasets and models can no longer be trained on a single machine. One common solution to this problem is to use distributed systems for training. The most common algorithms underlying deep learning are stochastic gradient descent (SGD) and its variants, which led to a significant amount of research on building and understanding distributed versions of SGD.

Implementations of SGD on distributed systems and data-parallel versions of SGD are scalable and take advantage of multi-GPU systems. Data-parallel SGD, in particular, has received significant attention due to its excellent scalability properties \citep{Zinkevich,Scaleup,Recht11,Dean12,Coates,Projectadam,Li14,Duchi15,Petuum,Zhang15,QSGD}. In data-parallel SGD, a large dataset is partitioned among $K$ processors. These processors work together to minimize an objective function. Each processor has access to the current parameter vector of the model. At each SGD iteration, each processor computes an updated stochastic gradient using its own local data. It then shares the gradient update with its peers. The processors collect and aggregate stochastic gradients to compute the updated parameter vector.      
 
Increasing the number of processing machines reduces the computational costs significantly. However, the communication costs to share and synchronize huge gradient vectors and parameters increases dramatically as the size of the distributed systems grows. Communication costs may thwart the anticipated benefits of reducing computational costs. Indeed, in practical scenarios, the communication time required to share stochastic gradients and parameters is the main performance bottleneck \citep{Recht11,Li14,Seide14,Strom15,QSGD}. Reducing communication costs in data-parallel SGD is an important problem.

One promising solution to the problem of reducing communication costs of data-parallel SGD is gradient compression, \eg through gradient quantization \citep{Dean12,Seide14,Sa,Gupta,Abadi,Dorefa,QSGD,TernGrad,signSGD}. 
(This should not be confused with weight quantization/sparsification, as studied by \citet{Wen16,Hubara,Park17,TernGrad}, which we do not discuss here.)
Unlike full-precision data-parallel SGD, where each processor is required to broadcast its local gradient in full-precision, 
\ie transmit and receive huge full-precision vectors at each iteration, quantization requires each processor to transmit only a few communication bits per iteration for each component of the stochastic gradient.

One popular such proposal for communication-compression is quantized SGD (QSGD), due to \citet{QSGD}.
In QSGD, stochastic gradient vectors are normalized to have unit $L^2$ norm, 
and then compressed by quantizing each element to a uniform grid of quantization levels using a randomized method.
While most lossy compression schemes do not provide convergence guarantees, QSGD's quantization scheme is designed to be unbiased, which implies that the quantized stochastic gradient is itself a stochastic gradient, only with higher variance determined by the dimension and number of quantization levels.
As a result, a number of theoretical guarantees are established for QSGD, 
including that it converges under standard assumptions. By changing the number 
of quantization levels, QSGD allows the user to trade-off communication 
bandwidth and convergence time.

Despite their theoretical guarantees based on quantizing after $L^2$ normalization, 
\citeauthor{QSGD} opt to present empirical results using $L^\infty$ normalization. We call this variation QSGDinf.
While the empirical performance of QSGDinf is strong, their theoretical guarantees on the number of bits transmitted no longer apply.
Indeed, in our own empirical evaluation of QSGD, we find the variance induced by quantization is substantial,
and the performance is far from that of SGD and QSGDinf.

Given the popularity of this scheme, it is natural to ask one can obtain guarantees as strong as those of QSGD 
while matching the practical performance of the QSGDinf heuristic. 
In this work, we answer this question in the affirmative 
by providing a new quantization scheme which fits into QSGD in a way that allows us to establish stronger theoretical guarantees on the variance, bandwidth, and cost to achieve a prescribed gap.
Instead of QSGD's uniform quantization scheme, we use an unbiased nonuniform logarithmic scheme, reminiscent of those introduced in telephony systems for audio compression \citep{History}.  
We call the resulting algorithm \emph{nonuniformly quantized stochastic gradient descent} (NUQSGD).
Like QSGD, NUQSGD is a quantized data-parallel SGD algorithm with strong theoretical guarantees 
that allows the user to trade off communication costs with convergence speed.
Unlike QSGD, NUQSGD has strong empirical performance on deep models and large datasets, matching that of QSGDinf. 
Beyond just the stronger theoretical guarantees, NUQSGD also allows us to employ non-trivial coding to the quantized gradients, as its code-length guarantees also hold in practice. 
Specifically, we provide a new efficient implementation for these schemes using a modern computational framework (Pytorch), and benchmark it on classic large-scale image classification tasks. 
Results showcase the practical performance of NUQSGD, which can surpass that of QSGDinf and of SignSGD with Error Feedback (EF)~\citep{karimireddy19a} when employing gradient coding, both in terms of communication-compression and end-to-end training time.

The intuition behind the nonuniform quantization scheme underlying NUQSGD
is that, after $L^2$ normalization, many elements of the normalized stochastic gradient will be near-zero.
By concentrating quantization levels near zero, we are able to establish stronger bounds on the excess variance.
These bounds decrease rapidly as the number of quantization levels increases.
In fact, we provide a lower bound showing that our variance bound is \emph{tight} w.r.t. the model dimension. 
We establish convergence guarantees for NUQSGD under standard assumptions for convex and nonconvex problems. Given the importance of momentum-based methods, we establish convergence guarantees for communication-efficient variants of SGD with momentum. 
We have derived a tight worst-case variance upper bound for a fixed set of arbitrary levels, expressed as the solution to an integer program with quadratic constraints. We can relax the program to obtain a quadratic program. A coarser analysis yields an upper bound expressed as the solution to a linear program, which is more amenable to analysis.
Finally, by combining the variance bound with a bound on the expected code-length, we obtain a bound on the total communication costs of achieving an expected suboptimality gap. The resulting bound is stronger than the one provided by QSGD.

To study how quantization affects convergence on state-of-the-art  deep models, 
we compare NUQSGD, QSGD,  QSGDinf, and  EF-SignSGD
focusing on training loss, variance, and test accuracy on standard deep models and large datasets.
Using the same number of bits per iteration, 
experimental results show that NUQSGD has smaller variance than QSGD, as predicted. 
The smaller variance also translates to improved optimization performance, in terms of both training loss and test accuracy. 
We also observe that NUQSGD matches the performance of QSGDinf in terms of variance and loss/accuracy. 
Further, our distributed implementation shows that the resulting algorithm considerably reduces communication cost of distributed training, without adversely impacting accuracy. 
Our empirical results show that NUQSGD can provide faster end-to-end parallel training relative to data-parallel SGD, QSGD, and EF-SignSGD on the ImageNet dataset \citep{ImageNet}, in particular when combined with non-trivial coding of the quantized gradients.

\subsection{Summary of Contributions}
\begin{itemize}[leftmargin=*,itemsep=0ex]
\item We propose a non-uniform gradient quantization method and establish strong theoretical guarantees for its excess variance and communication costs. 
These bounds are strictly stronger than those known for QSGD. In addition, we establish a lower bound on the variance that shows our bound is tight. 
\item We proceed to establish stronger convergence guarantees for NUQSGD for convex and nonconvex problems, under standard assumptions. 
We establish convergence guarantees for communication-efficient variants of SGD with momentum.
\item For generally spaced levels, we derive tight worst-case variance upper bounds expressed as an integer quadratic program and present several relaxations of this bound. 
\item We demonstrate that NUQSGD has strong empirical performance on deep models and large datasets, both in terms of accuracy and scalability. 
Thus, NUQSGD closes the gap between the theoretical guarantees of QSGD and the empirical performance of QSGDinf.
\end{itemize}

\subsection{Related Work}  
\citet{Seide14} proposed SignSGD, an efficient heuristic scheme to reduce communication costs drastically by quantizing each gradient component to two values. 
(This scheme is sometimes also called 1bitSGD~\citep{Seide14}.)
\citet{signSGD} later provided convergence guarantees for a variant of SignSGD. 
Note that the quantization employed by SignSGD is not unbiased, and so a new analysis was required.
As the number of levels is fixed, SignSGD does not provide any trade-off between communication costs and convergence speed.
\citet{karimireddy19a} proposed EF-SignSGD, which is an improved version of SignSGD. We compare NUQSGD and EF-SignSGD empirically.

\citet{Sa} introduced Buckwild!, a lossy compressed SGD with convergence guarantees.
The authors provided bounds on the error probability of SGD, assuming convexity, gradient sparsity, and unbiased quantizers. 

\citet{TernGrad} proposed TernGrad, a stochastic quantization scheme with three levels.
TernGrad also significantly reduces communication costs and obtains reasonable accuracy with a small degradation to performance compared to full-precision SGD. %
TernGrad can be viewed as a special case of QSGDinf with three quantization levels.

NUQSGD uses a logarithmic quantization scheme.\footnote{After the completion of this work, we became aware of earlier, independent work by \citet{Samuel}, which introduces gradient quantization to exponentially spaced levels (powers of $1/2$). They devise variance bounds for $L^p$ normalization. We obtain tighter variance bound for $L^2$ normalization, and extend our consideration to any arbitrary sequence of levels beyond powers of $1/2$.} Such schemes have long been used, e.g. in telephony systems for audio compression \citep{History}. 
Logarithmic quantization schemes have appeared in other contexts recently:
\citet{Hou18} studied weight distributions of long short-term memory networks and proposed to use logarithm quantization for network compression. \citet{Miyashita} and \citet{LogNet} proposed logarithmic encodings to represent weights and activations. \citet{Li19} obtained dimension-free bounds for logarithmic quantization of weights.
\citet{ZipML} proposed a gradient compression scheme and introduced an optimal quantization scheme, 
but for the setting where the points to be quantized are known in advance. As a result, their scheme is not applicable to the communication setting of quantized data-parallel SGD.

\section{Preliminaries: Data-parallel SGD and Convergence}\label{sec:conv}
We consider a high-dimensional machine learning model, parametrized by a vector $\wbf\in\reals^d$. 
Let $\Omega\subseteq\reals^d$ denote a closed and convex set. Our objective is to minimize $f:\Omega\rightarrow \reals$, 
which is an unknown, differentiable, and $\beta$-smooth function. The following summary is based on \citep{QSGD}.

Recall that a function $f$ is \defn{$\beta$-smooth} if, for all $\ubf,\vbf\in\Omega$, we have $\|\nabla f(\ubf)-\nabla f(\vbf)\|\leq \beta\|\ubf-\vbf\|$, where $\|\cdot\|$ denotes the Euclidean norm.
Let $(\Sc,\Sigma,\mu)$ be a probability space (and let $\E$ denote expectation). %
Assume we have access to stochastic gradients of $f$, \ie we have access to a function $g:\Omega\times\Sc\rightarrow\reals^d$ such that, if $s\sim\mu$, then $\E[g(\wbf,s)]=\nabla f(\wbf)$ for all $\wbf\in\Omega$. In the rest of the paper, we let $g(\wbf)$ denote the stochastic gradient for notational simplicity.
The update rule for conventional full-precision projected SGD is %
$
\wbf_{t+1}=\Pbf_\Omega(\wbf_t-\alpha g(\wbf_t)),
$
where $\wbf_t$ is the current parameter input, $\alpha$ is the learning rate, and $\Pbf_\Omega$ is the Euclidean projection onto $\Omega$. 

We say the stochastic gradient has a \defn{second-moment upper bound} $B$ when $\E[\|g(\wbf)\|^2]\leq B$ for all $\wbf\in \Omega$. 
Similarly, the stochastic gradient has a \defn{variance upper bound} $\sigma^2$ when $\E[\|g(\wbf)-\nabla f(\wbf)\|^2]\leq \sigma^2$ for all $\wbf\in \Omega$.   
Note that a second-moment upper bound implies a variance upper bound, because the stochastic gradient is unbiased.

We have classical convergence guarantees for conventional full-precision SGD given access to stochastic gradients at each iteration:

\begin{theorem}[{\citealt[Theorem 6.3]{Bubeck}}]\label{convbound}
Let $f:\Omega\rightarrow \reals$ denote a convex and $\beta$-smooth function and let $R^2 \defeq \sup_{\wbf\in\Omega}\|\wbf-\wbf_0\|^2$.
Suppose that the projected SGD update %
is executed for $T$ iterations with $\alpha=1/({\beta+1/\gamma})$ 
where $\gamma= R\sqrt{1/T}/\sigma$.
Given repeated and independent access to stochastic gradients with a variance upper bound $\sigma^2$, projected SGD satisfies 
\begin{align}\label{convSGDbound}
\E\Big[f\big(\frac{1}{T}\sum_{t=0}^T\wbf_t\big)\Big]-\min_{\wbf\in\Omega}f(\wbf)\leq R\sqrt{\frac{\sigma^2}{T}}+\frac{\beta R^2}{2T}. 
\end{align}     
\end{theorem}

Following \citep{QSGD}, we consider data-parallel SGD, a synchronous distributed framework consisting of $K$ processors that partition a large dataset among themselves. This framework models real-world systems with multiple GPU resources. Each processor keeps a local copy of the parameter vector and has access to independent and private stochastic gradients of $f$.

At each iteration, each processor computes its own stochastic gradient based on its local data and then broadcasts it to all peers. Each processor receives and aggregates the stochastic gradients from all peers to obtain the updated parameter vector. In detail, the update rule for full-precision data-parallel SGD is $\wbf_{t+1}=\Pbf_\Omega(\wbf_t-\frac{\alpha}{K}\sum_{l=1}^K\overline g_l(\wbf_t))$ where $\overline g_l(\wbf_t)$ is the stochastic gradient computed and broadcasted by processor $l$. Provided that $\overline g_l(\wbf_t)$ is a stochastic gradient with a variance upper bound $\sigma^2$ for all $l$, then $\frac{1}{K}\sum_{l=1}^K\overline g_l(\wbf_t)$ is a stochastic gradient with a variance upper bound $\frac{\sigma^2}{K}$. Thus, aggregation improves convergence of SGD by reducing the first term of the upper bound in \eqref{convSGDbound}. Assume each processor computes a mini-batch gradient of size {$J$}. Then, this update rule is essentially a mini-batched update with size {$JK$}.

Data-parallel SGD is described in Algorithm \ref{NUQSGDalg}. Full-precision data-parallel SGD is a special case of Algorithm \ref{NUQSGDalg} with identity encoding and decoding mappings. Otherwise, the decoded stochastic gradient $\hat g_i(\wbf_t)$ is likely to be different from the original local stochastic gradient $g_i(\wbf_t)$. 

By Theorem~\ref{convbound}, we have the following convergence guarantees for full-precision data-parallel SGD:
\bcr[{\citealt[Corollary 2.2]{QSGD}}]\label{dataparbound} Let $f$, $R$, and $\gamma$ be as defined in Theorem \ref{convbound} and let $\epsilon>0$. Suppose that the projected SGD update 
is executed for $T$ iterations with $\alpha=1/({\beta+\sqrt{K}/\gamma})$ on $K$ processors, each with access to independent stochastic gradients of $f$ with a second-moment bound $B$. The smallest $T$ for the full-precision data-parallel SGD that guarantees $\E\big[f(\frac{1}{T}\sum_{t=0}^T\wbf_t)\big]-\min_{\wbf\in\Omega}f(\wbf)\leq \epsilon$ is $T_{\epsilon}=O\big(R^2\max(\frac{B}{K\epsilon^2},\frac{\beta}{2\epsilon})\big)$.   
\ecr

\section{Nonuniformly Quantized Stochastic Gradient Descent} %
\label{sec:nqsgd}

\begin{algorithm}[t]
\SetAlgoLined
\KwIn{local data, local copy of the parameter vector $\wbf_t$, learning rate $\alpha$, and $K$}
\For{$t=1$ {\bfseries to} $T$}{
	\For(\tcp*[h]{each transmitter processor (in parallel)}){$i=1$ {\bfseries to} $K$}{
	Compute $g_i(\wbf_t)$ \tcp*[l]{stochastic gradient} %
	Encode $c_{i,t}\leftarrow \ENCODE\big(g_i(\wbf_t)\big)$\;
	Broadcast $c_{i,t}$ to all processors\;
	}
	\For(\tcp*[h]{each receiver processor (in parallel)}){$l=1$ {\bfseries to} $K$ }{ 
	\For(\tcp*[h]{each transmitter processor}){$i=1$ {\bfseries to} $K$ }{ 
	Receive $c_{i,t}$ from processor $i$ for each $i$\;
  	Decode $\hat g_i(\wbf_t)\leftarrow \text{DECODE}\big(c_{i,t}\big)$\;
	}
  	Aggregate $\wbf_{t+1}\leftarrow \Pbf_\Omega(\wbf_t-\frac{\alpha}{K}\sum_{i=1}^K\hat g_i (\wbf_t))$\;\label{step:agg}
	}
}

\caption{Data-parallel (synchronized) SGD.}
\label{NUQSGDalg}
\end{algorithm}
Data-parallel SGD reduces computational costs significantly. However, the communication costs of broadcasting stochastic gradients is the main performance bottleneck in large-scale distributed systems. 
In order to reduce communication costs and accelerate training, \citet{QSGD} introduced a compression scheme that produces a compressed and unbiased stochastic gradient, suitable for use in SGD.

At each iteration of QSGD,
each processor broadcasts an encoding of its own compressed stochastic gradient,
decodes the stochastic gradients received from other processors,
and sums all the quantized vectors to produce a stochastic gradient.
In order to compress the gradients, 
every coordinate (with respect to the standard basis) of the stochastic gradient is normalized by the Euclidean norm of the gradient and then stochastically quantized to one of a small number quantization levels distributed uniformly in the unit interval. 
The stochasticity of the quantization is necessary to not introduce bias.

\citet{QSGD} give a simple argument that provides a \emph{lower} bound on the number of coordinates that are quantized to zero in expectation. 
Encoding these zeros efficiently provides communication savings at each iteration.
However, the cost of their scheme is greatly increased variance in the gradient, 
and thus slower overall convergence.
In order to optimize overall performance, 
we must balance communication savings with variance.

By simple counting arguments,
the distribution of the (normalized) coordinates cannot be uniform. Indeed, this is the basis of the lower bound on the number of zeros.
These arguments make no assumptions on the data distribution, and rely entirely on the fact that the quantities being quantized are the coordinates of a unit-norm vector.
Uniform quantization does not capture the properties of such vectors,
leading to substantial gradient variance.   

\subsection{Nonuniform Quantization}\label{sec:nonunif}
\begin{figure}[t]
\centerline{\includegraphics[width=0.5\columnwidth]{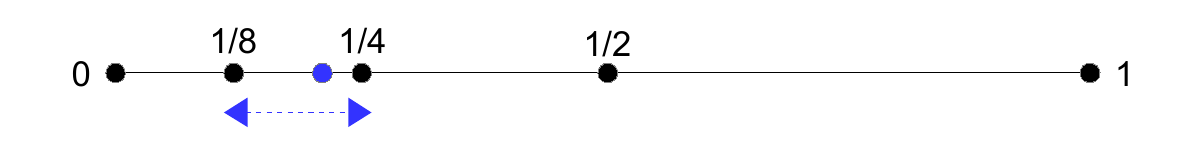}}
\caption{An example of nonuniform stochastic quantization with $s=3$. The point between the arrows represents the value of the normalized coordinate. It will be quantized to either 1/8 or 1/4. In this case, the point is closer to 1/4, and so will be more likely to be quantized to 1/4. The  probabilities are chosen so that the mean of the quantization is the unquantized coordinate's value.}
\label{NUQ}
\end{figure}
In this paper, we propose and study a new scheme to quantize normalized gradient vectors.
Instead of uniformly distributed quantization levels, as proposed by \citet{QSGD},
we consider quantization levels that are nonuniformly distributed in the unit interval, as depicted in Figure \ref{NUQ}.
In order to obtain a quantized gradient that is suitable for SGD, we need
the quantized gradient to remain unbiased. \citet{QSGD} achieve this 
via a randomized quantization scheme, which can be easily generalized to the case of
nonuniform quantization levels.

Using a carefully parametrized generalization of the unbiased quantization scheme introduced by \citeauthor{QSGD},
we can control both the cost of communication and the variance of the gradient.
Compared to a uniform quantization scheme, our scheme reduces quantization error and variance by better matching the properties of normalized vectors. In particular, by increasing the number of quantization levels near zero, we obtain a stronger variance bound.
Empirically, our scheme also better matches the distribution of normalized coordinates observed on real datasets and networks. %

We now describe the nonuniform quantization scheme:
Let $s\in \{1,2,\cdots\}$ be the number of internal quantization levels,
and let $\Lc=(l_0,l_1,\cdots,l_{s+1})$ denote the sequence of quantization levels, 
where $l_0=0 < l_1 < \dotsm < l_{s+1}=1$.  
For $r\in[0,1]$, let $\level{r}$ and $p(r)$ satisfy $l_{\level{r}}\leq r\leq l_{\level{r}+1}$ and $r=\big(1-p(r)\big)l_{\level{r}}+p(r)l_{\level{r}+1}$,  respectively. 
Define $\tau(r)=l_{\level{r}+1}-l_{\level{r}}$. Note that $\level{r}\in\{0,1,\cdots,s\}$.  

\begin{definition}The nonuniform quantization of a vector $\vbf\in\reals^d$ is 
\begin{align}
Q_s(\vbf)\defeq [Q_s(v_1),\cdots,Q_s(v_d)]^T \quad\text{where}\quad
Q_s(v_i)=\|\vbf\|\cdot \mathrm{sign}(v_i)\cdot h_i(\vbf,s)
\end{align} 
where,
letting $r_i=|v_i|/\|\vbf\|$,
the $h_i(\vbf,s)$'s are independent random variables such that
$h_i(\vbf,s)= l_{\level{r_i}}$ with probability $1-p(r_i)$ and 
$h_i(\vbf,s)= l_{\level{r_i}+1}$ otherwise.
\end{definition}

We note that the distribution of $h_i(\vbf,s)$ satisfies $\E[h_i(\vbf,s)]=r_i$ and achieves the minimum variance over all distributions that satisfy $\E[h_i(\vbf,s)]=r_i$ with support $\Lc$.
We first focus on a special case of nonuniform quantization with 
$\hat\Lc=(0,1/{2^s},\cdots,{2^{s-1}}/{2^s},1)$ as the quantization levels.  In Section \ref{sec:QPLP}, we extend our consideration to any arbitrary sequence of levels.

The intuition behind this quantization scheme is that it is very unlikely to observe large values of $r_i$ in the stochastic gradient vectors of machine learning models. Stochastic gradients are observed to be dense vectors \citep{signSGD}. Hence, it is natural to use fine intervals for small $r_i$ values to reduce quantization error and control the variance.

After quantizing the stochastic gradient with a small number of discrete levels, each processor must encode its local gradient into a binary string for broadcasting. We describe this encoding in Appendix~\ref{app:elias}.

\section{Theoretical Guarantees}

In this section, we provide theoretical guarantees for NUQSGD, 
giving variance and code-length bounds, 
and using these in turn to compare NUQSGD and QSGD.
Please note that the proofs of Theorems \ref{thm:varbound}, \ref{thm:codebound}, and \ref{thm:totalbits} are provided in Appendices~\ref{app:pr_var}, \ref{app:pr_code}, and \ref{app:pr_totalbits} respectively.

\bth[Variance bound]\label{thm:varbound}
Let $\vbf\in\reals^d$. The nonuniform quantization of $\vbf$ satisfies $\E[Q_s(\vbf)]=\vbf$. Then we have
\begin{align}\label{varbound}
\E[\|Q_s(\vbf)-\vbf\|^2]\leq\epsilon_Q\|\vbf\|^2
\end{align} where $\epsilon_Q=(1/8+2^{-2s-2}d)\mathbbm{1}\{d<2^{2s+1}\}+(2^{-s}\sqrt{d}-7/8)\mathbbm{1}\{d\geq 2^{2s+1}\}$ and $\mathbbm{1}$ denotes the indicator function. Provided that $s \le \log(d)/2$, then it also holds that $\E[\|Q_s(\vbf)-\vbf\|^2]\leq\hat\epsilon_Q\|\vbf\|^2$ where $\hat\epsilon_Q=\min\{2^{-2s}/4(d-2^{2s}),2^{-s}\sqrt{d-2^{2s}}\}+O(s)$. 
\eth

The result in Theorem \ref{thm:varbound} implies that if $g(\wbf)$ is a stochastic gradient with a second-moment bound $\eta$, then $Q_s(g(\wbf))$ is a stochastic gradient with a variance upper bound $\epsilon_Q\eta$. Note that the variance upper bound decreases with the number of quantization levels. In the range of $s = o(\log(d))$, $\hat \epsilon_Q$ decreases with $s$, which is because the first term in the upper bound decreases exponentially fast in $s$. To obtain $\hat \epsilon_Q$, we establish upper bounds on the number of coordinates of $\vbf$ falling into intervals defined by $\hat\Lc$. %
Our bound is tighter than the bound of \citet{Samuel}.

\bth[Code-length bound]\label{thm:codebound}
Let $\vbf\in\reals^d$. Provided  $d$ is large enough to ensure $2^{2s}+\sqrt{d}2^s\leq d/e$, 
the expectation $\E[|\ENCODE(\vbf)|]$
 of the number of communication bits needed to transmit $Q_s(\vbf)$ is bounded above by
\begin{align}
\label{codebound}\textstyle
N_Q = C+3n_{s,d} &+(1+o(1))n_{s,d}\log\Big(\frac{d}{n_{s,d}}\Big)+(1+o(1))n_{s,d}\log\log\Big(\frac{8(2^{2s}+d)}{n_{s,d}}\Big)  
\end{align}
where $C=b-(1+o(1))$ and $n_{s,d}=2^{2s}+2^s\sqrt{d}$.\footnote{In practice, we use standard 32-bit floating point encoding, \ie $b=32$.} 
\eth
Theorem \ref{thm:codebound} provides a bound on the expected number of communication bits to encode the quantized stochastic gradient. 
Note that $2^{2s}+\sqrt{d}2^s\leq d/e$ is a mild assumption in practice.
As one would expect, the bound, \eqref{codebound}, increases monotonically in $d$ and $s$.
In the sparse case, if we choose $s=o(\log d)$ levels, then the upper bound on the expected code-length is $O\big(2^s\sqrt{d}\log\big(\frac{\sqrt{d}}{2^s}\big)\big)$.

Combining the upper bounds above on the variance and code-length, 
Corollary \ref{dataparbound} implies the following guarantees for NUQSGD:  
\bth[NUQSGD for smooth convex optimization]\label{thm:NUQSGDconvbound}
Let $f$ and $R$ be defined as in Theorem~\ref{convbound},
let $\epsilon_Q$ be defined as in Theorem~\ref{thm:varbound},
let $\epsilon > 0$, $\hat B=(1+\epsilon_Q) B$, and let $\gamma > 0$ be given by $\gamma^2=R^2/({\hat B}T)$.
With $\ENCODE$ and $\DECODE$ defined as in Appendix~\ref{app:elias}, 
suppose that Algorithm \ref{NUQSGDalg} is executed for $T$ iterations
with a learning rate $\alpha=1/({\beta+\sqrt{K}/\gamma})$
on $K$ processors, 
each with access to independent stochastic gradients of $f$ with a second-moment bound $B$.
Then $T_{\epsilon}=O\big(\max\big(\frac{\hat B}{K\epsilon^2},\frac{\beta}{2\epsilon}\big)R^2\big)$ iterations suffice to guarantee
$\E\left[f\left(\frac{1}{T}\sum_{t=0}^T\wbf_t\right)\right]-\min_{\wbf\in\Omega}f(\wbf)\leq \epsilon.$ 
In addition, NUQSGD requires at most $N_Q$ communication bits per iteration in expectation. 
\eth  
\bpr
Let $g(\wbf)$ and $\hat g(\wbf)$ denote the full-precision and decoded stochastic gradients, respectively.
Then
\begin{align}\label{quantvarbound1}
\E[\|\hat g(\wbf)-\nabla f(\wbf)\|^2]\leq \E[\|g(\wbf)-\nabla f(\wbf)\|^2]+\E[\|\hat g(\wbf)-g(\wbf)\|^2] .
\end{align} 
By Theorem \ref{thm:varbound}, $\E[\|\hat g(\wbf)-g(\wbf)\|^2] \le \epsilon_Q \E[\|g(\wbf)\|^2]$.
By assumption, $\E[\|g(\wbf)\|^2]\leq B$. Noting $g(\wbf)$ is unbiased, $\E[\|\hat g(\wbf)-\nabla f(\wbf)\|^2]\leq (1+\epsilon_Q)B$.
The result follows by Corollary~\ref{dataparbound}.
\epr

In the following theorem, we show that for any given set of levels, there exists a distribution of points with dimension d such that the variance is in $\Omega(\sqrt{d})$, and so our bound is tight in $d$. 

\bth[Lower bound]\label{thm:lowerbound}
Let $d\in\integers^{>0}$ and let $(0,l_1,\cdots,l_s,1)$ denote an arbitrary sequence of quantization levels. Provided $d\geq (2/l_1)^2$, there exists a vector $\vbf\in\reals^d$ such that the variance of unbiased quantization of $\vbf$ is lower bounded by $\|\vbf\|^2l_1\sqrt{d}/2$, \ie the variance is in $\Omega(\sqrt{d})$. 
\eth
\bpr  
The variance of $Q_s(\vbf)$ for general sequence of quantization levels is given by  
\begin{align}\nn
\E[\|Q_s(\vbf)-\vbf\|^2]=\|\vbf\|^2 \sum_{i=1}^d\sigma^2(r_i).
\end{align} 

If $r\in[l_{\level{r}},l_{\level{r}+1}]$, the variance $\sigma^2(r)$ can be expressed as
\begin{align}\label{var_norm_coor}
\sigma^2(r)=\tau(r)^2p(r)\big(1-p(r)\big)=(l_{\level{r}+1}-r)(r-l_{\level{r}}). 
\end{align}
We consider $\vbf_0=[r,r,\cdots,r]^T$ for $r\neq 0$. The normalized coordinates is $\hat\vbf_0=[1/\sqrt{d},\cdots,1/\sqrt{d}]^T$.

Using \eqref{var_norm_coor} and noting $1/\sqrt{d}<l_1$, we have 
\begin{align}\label{var_lb_coor}
\sigma^2(r_0) = 1/\sqrt{d}\big(l_1-1/\sqrt{d}\big)\geq l_1/(2\sqrt{d}).
\end{align}

Summing variance of all coordinates and applying \eqref{var_lb_coor}, the variance of $Q_s(\vbf_0)$ is lower bounded by 
\begin{align}
\E[\|Q_s(\vbf_0)-\vbf_0\|^2]=\|\vbf_0\|^2d\sigma^2(r)\geq \|\vbf_0\|^2l_1\sqrt{d}/2.
\end{align}
\epr

We can obtain convergence guarantees to various learning problems where we have convergence guarantees for SGD under standard assumptions.  
On nonconvex problems, (weaker) convergence guarantees can be established along the lines of, e.g., \citep[Theorem 2.1]{Ghadimi}. In particular, NUQSGD is guaranteed to converge to a local minima for smooth general loss functions.

\bth[NUQSGD for smooth nonconvex optimization]\label{nonconvbound}
Let $f:\Omega\rightarrow \reals$ denote a possibly nonconvex and $\beta$-smooth function. Let $\wbf_0\in\Omega$ denote an initial point, $\epsilon_Q$ be defined as in Theorem~\ref{thm:varbound}, $T\in\integers^{>0}$, and $f^*=\inf_{\wbf\in\Omega}f(\wbf)$.
Suppose that Algorithm \ref{NUQSGDalg} is executed for $T$ iterations with a learning rate $\alpha<2/\beta$ on $K$ processors, 
each with access to independent stochastic gradients of $f$ with a second-moment bound $B$.
Then there exists a random stopping time $R\in\{0,\cdots,T\}$ such that NUQSGD guarantees $\E[\|\nabla f(\wbf_R)\|^2]\leq\beta(f(\wbf_0)-f^*)/T+2(1+\epsilon_Q) B/K$. 
\eth

\subsection{Worst-case Variance Analysis}\label{sec:QPLP}
In this section, we first derive a tight worst-case variance upper bound by optimizing over the distribution of normalized coordinates for an arbitrary sequence of levels, expressed as a solution to an integer program with quadratic constraints. We then relax the program to obtain a quadratically constrained quadratic program (QCQP). A coarser analysis yields an upper bound expressed as a solution to a linear program (LP), which is more amenable to analysis. We solve this LP analytically for the special case of $s=1$ and show the optimal level is at $1/2$. 

Then, for an exponentially spaced collection of levels of the form $(0,p^s,\cdots, p^2 ,p,1)$ for $p\in(0,1)$ and an integer number of levels, $s$, we write the expression of QCQP and solve it efficiently using standard solvers. We have a numerical method for finding the optimal value of $p$ that minimizes the worst-case variance, for any given $s$ and $d$. Through the worst-case analysis, we gain insight into the behaviour of the variance upper bound. We show that our current scheme is nearly optimal (in the worst-case sense) in some cases. Using these techniques we can obtain slightly tighter bounds numerically.

\subsubsection{Generally Spaced Levels}\label{sec:QPLPgen}
Let $\Lc=(l_0,l_1,\cdots,l_s,l_{s+1})$ denote an arbitrary sequence of quantization levels where $ l_0=0<l_1 < \dotsm < l_{s+1}=1$. Recall that, for $r\in[0,1]$, we define $\level{r}$ and $p(r)$ such that they satisfy $l_{\level{r}}\leq r\leq l_{\level{r}+1}$ and $r=\big(1-p(r)\big)l_{\level{r}}+p(r)l_{\level{r}+1}$, respectively. Define $\tau(r)=l_{\level{r}+1}-l_{\level{r}}$. Note that $\level{r}\in\{0,1,\cdots,s\}$. Then, $h_i(\vbf,s)$'s are defined in two cases based on which quantization interval $r_i$ falls into: 

1) If $r_i\in[0,l_1]$, then 
\begin{align}\label{genh_case1}\textstyle
h_i(\vbf,s)=
  \Bigg\{\begin{array}{ll}
        0 & \text{with probability}~1-p_1(r_i,\Lc);\\
       l_1 & \text{otherwise}
        \end{array}
\end{align} where $p_1\big(r,\Lc\big)=r/l_1.$

2) If $r_i\in[l_{j-1},l_j]$ for $j=1,\cdots,s+1$, then 
\begin{align}\label{genh_case2}\textstyle
h_i(\vbf,s)=
  \Bigg\{\begin{array}{ll}
        l_{j-1} & \text{with probability}~1-p_2(r_i,\Lc);\\
       l_j & \text{otherwise}
        \end{array}
\end{align}
where $p_2\big(r,\Lc\big)=(r-l_{j-1})/\tau_{j-1}.$ 

Let $\Sc_j$ denote the coordinates of vector $\vbf$ whose elements fall into the $(j+1)$-th bin, \ie $\Sc_j\defeq\{i:r_i\in[l_j,l_{j+1}]\}$ for $j=0,\cdots,s$. Let $d_j\defeq |\Sc_j|$.

Following Lemma \ref{lm:genvar} and steps in Theorem \ref{thm:varbound}, we can show that
\begin{align}\label{genvarbound_bad}
\E[\|Q_s(\vbf)-\vbf\|^2]&\leq\|\vbf\|^2(\min\{\tau_0^2d_0/4,\tau_0\sqrt{d_0}\}+\sum_{j=1}^s\min\{\tau_j^2d_j/4,\tau_j(\sqrt{d_j}-l_jd_j)\}).
\end{align}

\bth[QCQP bound]\label{thm:QP}
Let $\vbf\in\reals^d$. An upper bound on the nonuniform quantization of $\vbf$ is given by $\epsilon_{QP}\|\vbf\|^2$ where $\epsilon_{QP}$ is the optimal value of the following QCQP: 
\begin{align}
\Qc_1:\quad\max_{(d_0,\cdots,d_s,z_0,\cdots,z_s)}&~\sum_{j=0}^sz_j\nn\\
\st&d-d_0-\cdots-d_j\leq (1/l_{j+1})^2,~j=0,\cdots,s-1,\nn\\
&\sum_{j=0}^sd_j\leq d,~z_0\leq\tau_0^2d_0/4,~z_0^2\leq\tau_0^2d_0,\nn\\
&z_j\leq\tau_j^2d_j/4,~z_j^2+\tau_j^2l_j^2d_j^2+2\tau_jl_jd_jz_j\leq\tau_j^2d_j,~j=1,\cdots,s,\nn\\
&d_j\geq 0,~j=0,\cdots,s.\nn
\end{align}\eth
\bpr
Following Lemma \ref{lm:sparsity}, we have 
\begin{align}\label{genspar_cons}
d-d_0-d_1-\cdots-d_j\leq (1/l_{j+1})^2
\end{align} for $j=0,\cdots,s-1$.

The problem of optimizing $(d_0,\cdots,d_s)$ to maximize the variance upper bound \eqref{genvarbound_bad} subject to \eqref{genspar_cons} is given by 
\begin{align}
\Rc_1:\quad\max_{(d_0,\cdots,d_s)}&~\sum_{j=0}^s\min\{\tau_j^2d_j/4,\tau_j(\sqrt{d_j}-l_jd_j)\}\nn\\
\st &\eqref{genspar_cons},~j=0,\cdots,s-1,\nn\\
&\label{dim_cons}\sum_{j=0}^sd_j\leq d,\\
\label{int_cons}&d_j\in\integers^{\geq 0},~j=0,\cdots,s. 
\end{align} 
  
Let $z_j\defeq\min\{\tau_j^2d_j,\tau_j\sqrt{d_j}\}$ denote an auxiliary variable for $j=0,\cdots,s$. Problem $\Rc_1$ can be rewritten as 
\begin{align}
\Rc_2:\quad\max_{(d_0,\cdots,d_s,z_0,\cdots,z_s)}&~\sum_{j=0}^sz_j\nn\\
\st&\eqref{genspar_cons},~\eqref{dim_cons},~\and~\eqref{int_cons},\nn\\
\begin{split}\label{aux_cons}
&z_0\leq\tau_0^2d_0/4,~z_0^2\leq\tau_0^2d_0,\\
&z_j\leq\tau_j^2d_j/4,~z_j^2+\tau_j^2l_j^2d_j^2+2\tau_jl_jd_jz_j\leq\tau_j^2d_j,~j=1,\cdots,s.
\end{split}
\end{align} 
   
The variance optimization problem $\Rc_2$ is an integer nonconvex problem. We can obtain an upper bound on the optimal objective of problem $\Rc_2$ by relaxing the integer constraint as follows. The resulting QSQP is shown as follows: 
\begin{align}
\Qc_1:\quad\max_{(d_0,\cdots,d_s,z_0,\cdots,z_s)}&~\sum_{j=0}^sz_j\nn\\
\st &\label{q_cons}d_j\geq 0,~j=0,\cdots,s,\\
&\eqref{genspar_cons},~\eqref{dim_cons},~\and~\eqref{aux_cons}.\nn
\end{align} 
Note that problem $\Qc_1$ can be solved efficiently using standard standard interior point-based solvers, \eg CVX \citep{Convex}.\epr

In the following, we develop a coarser analysis that yields an upper bound expressed as the optimal value to an LP. 
\bth[LP bound]\label{thm:LP}
Let $\vbf\in\reals^d$. An upper bound on the nonuniform quantization of $\vbf$ is given by $\epsilon_{LP}\|\vbf\|^2$ where $\epsilon_{LP}$ is the optimal value of the following LP: 
\begin{align}
\Pc_1:\quad\max_{(d_0,\cdots,d_s)}&~\sum_{j=0}^s\tau_j^2d_j/4\nn\\
\st&d-d_0-\cdots-d_j\leq (1/l_{j+1})^2,~j=0,\cdots,s-1,\nn\\
&\sum_{j=0}^sd_j\leq d,\nn\\
&d_j\geq 0,~j=0,\cdots,s.\nn
\end{align}\eth
\bpr 
The proof follows the steps in the proof of Theorem \ref{thm:QP} for the problem of optimizing $(d_0,\cdots,d_s)$ to maximize the following upper bound

\begin{align}\label{genvarbound_coarse}
\E[\|Q_s(\vbf)-\vbf\|^2]&\leq\|\vbf\|^2\sum_{j=0}^s \tau_j^2d_j/4.
\end{align}
\epr

The LP bound can be solved exactly in some simple cases. In Appendix \ref{app:onelevel}, we present the optimal solution for the special case with $s=1$.

\subsubsection{Exponentially Spaced Levels}\label{sec:QPLPexp}
\begin{figure}[t]
\centering
   \includegraphics[width=0.49\textwidth]{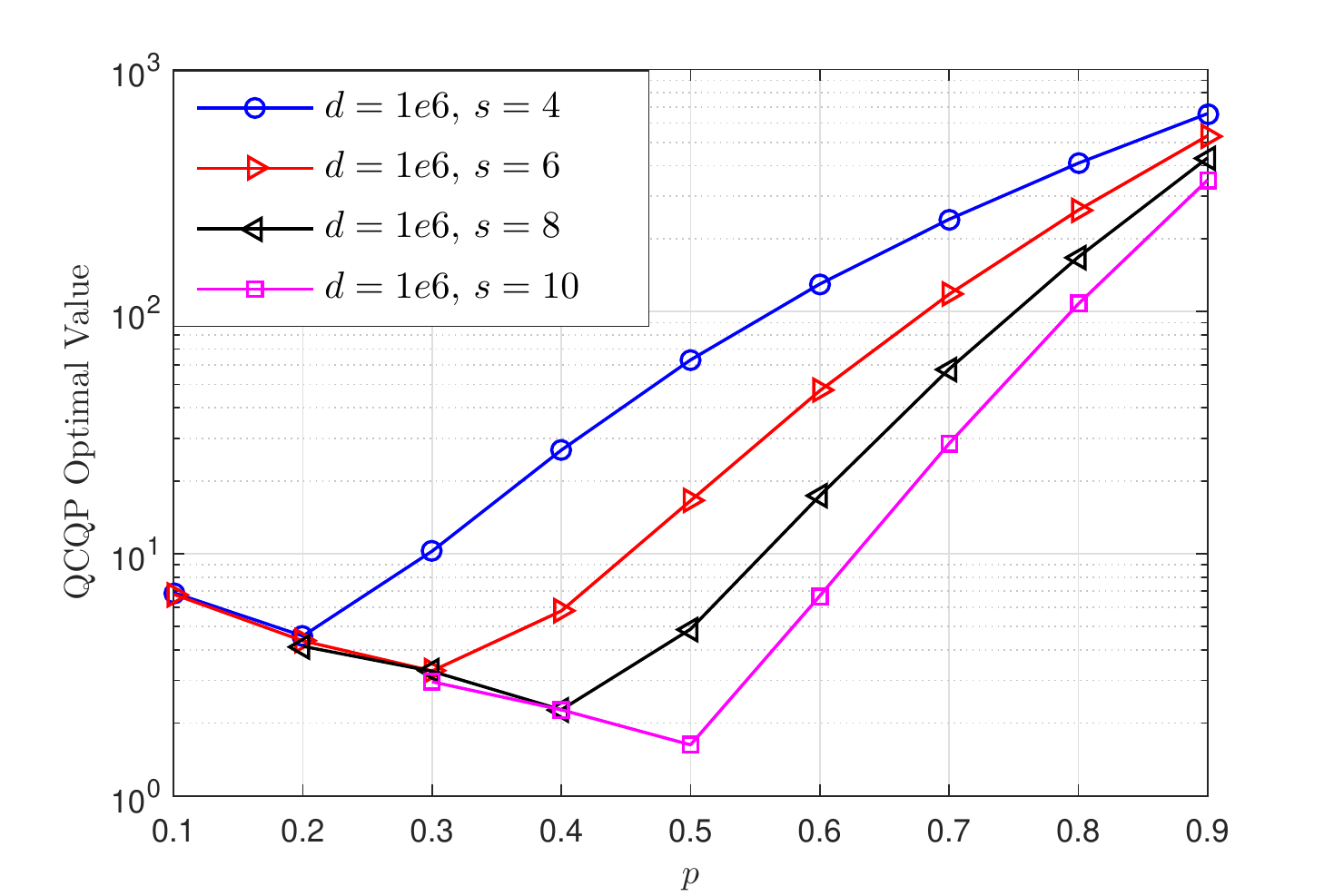}
    \hfill
    \includegraphics[width=0.49\textwidth]{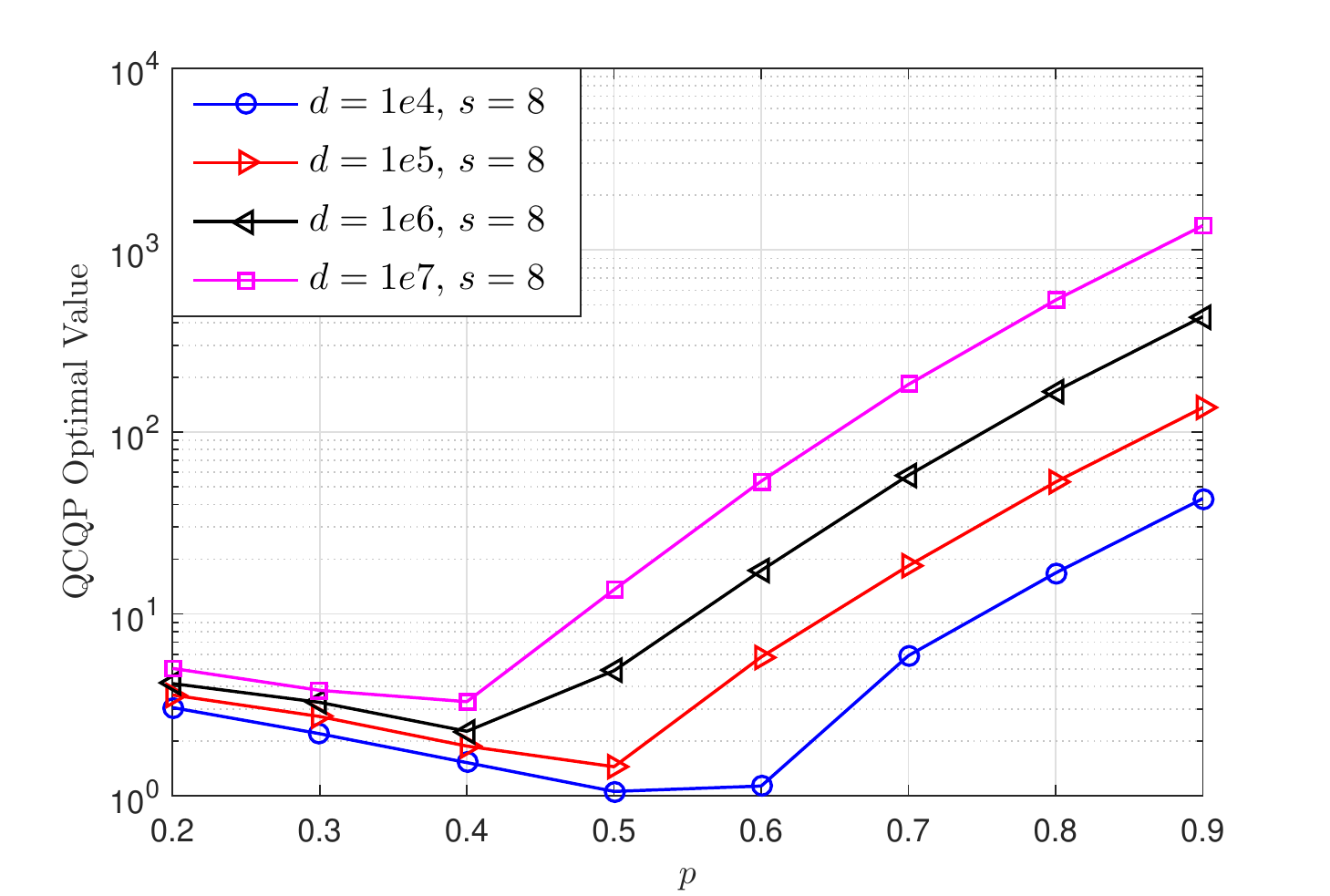}
    \caption{Optimal value of problem $\Qc_1$ versus $p\in [0,1]$ for 
    exponentially spaced collection of levels of the form $(0,p^s,\cdots, p^2 
    ,p,1)$.}
    \label{fig:QP}
\end{figure}

In this section, we focus on the special case of exponentially spaced collection of levels of the form $\Lc_p=(0,p^s,\cdots, p^2 ,p,1)$ for $p\in(0,1)$ and an integer number of levels, $s$. In this case, we have $\tau_0=p^s$ and $\tau_j=(1-p)p^{s-j}$ for $j=1,\cdots,s$.

For any given $s$ and $d$, we can solve the corresponding quadratic and linear programs efficiently to find the worst-case variance bound. As a bonus, we can find the optimal value of $p$ that minimizes the worst-case variance bound. In Figure \ref{fig:QP}, we show the numerical results obtained by solving QCQP $\Qc_1$ with $\Lc_p$ versus $p$ using CVX \citep{Convex}. In Figure \ref{fig:QP} (left), we fix $d$ and vary $s$, while in Figure \ref{fig:QP} (right), we fix $s$ and vary $d$. As expected, we note that the variance upper bound increases as $d$ increases and the variance upper bound decreases as $s$ increases. We observe that our current scheme is nearly optimal (in the worst-case sense) in some cases. Further, 
the optimal value of $p$ shifts to the right as $d$ increases and shifts to the left as $s$ increases.

\subsection{NUQSGD with Momentum}%
We have convergence guarantees for NUQSGD with momentum along the lines of, e.g., \citep{Unified} where convergence guarantees for momentum-based methods are established under standard assumptions.

The update rule for full-precision SGD with momentum is   
\begin{align}
\begin{split}\label{sgdm}
\ybf_{t+1}&=\wbf_t-\alpha g(\wbf_t)\\
\ybf^l_{t+1}&=\wbf_t-l\alpha g(\wbf_t)\\
\wbf_{t+1}&=\ybf_{t+1}+\mu\big(\ybf^l_{t+1}-\ybf^l_t\big)
\end{split}
\end{align} 
where $\wbf_t$ is the current parameter input and $\mu\in[0,1)$ is the momentum parameter \citep{Unified}. Note that the heavy-ball method \citep{Polyak} and Nesterov’s accelerated gradient method \citep{Nesterov} are obtained by substituting $l=0$ and $l=1$, respectively. 

One obtains data-parallel SGD with momentum by taking Algorithm \ref{NUQSGDalg} and replacing step \ref{step:agg} with \eqref{sgdm}. %
We first establish the convergence guarantees for convex optimization in the following theorem. 

\bth[NUQSGD with momentum for convex optimization]\label{convQUM}
Let $f:\reals^d\rightarrow \reals$ denote a convex function with $\|\nabla f(\wbf)\|\leq V$ for all $\wbf$. 
Let $\wbf_0$ denote an initial point, $\wbf^*=\arg\min f(\wbf)$, $\hat\wbf_T=1/T\sum_{t=0}^T\wbf_t$, and $\epsilon_Q$ be defined as in Theorem~\ref{thm:varbound}.

Suppose that  NUQSGD with momentum %
is executed for $T$ iterations with  a learning rate $\alpha>0$ on $K$ processors, 
each with access to independent stochastic gradients of $f$ with a second-moment bound $B$.
Then  NUQSGD with momentum satisfies 
\begin{align}\label{convNUQUMbound}
\begin{split}
\E[f(\hat\wbf_T)]-\min_{\wbf\in\Omega}f(\wbf)\leq&~\frac{\mu (f(\wbf_0)-f(\wbf^*))}{(1-\mu)(T+1)}+\frac{(1-\mu)\|\wbf_0-\wbf^*\|^2}{2\alpha(T+1)}\\
&+\frac{\alpha (1+2l\mu)(V^2+(1+\epsilon_Q) B/K)}{2(1-\mu)}.
\end{split}
\end{align}
\eth

On nonconvex problems, (weaker) convergence guarantees can be established for  NUQSGD with momentum.  In particular,  NUQSGD with momentum is guaranteed to converge to a local minima for smooth general loss functions.

\bth[NUQSGD with momentum for smooth nonconvex optimization]\label{nonconvQUM}
Let $f:\reals^d\rightarrow \reals$ denote a possibly nonconvex and $\beta$-smooth function with  $\|\nabla f(\wbf)\|\leq V$ for all $\wbf$. 
Let $\wbf_0$ denote an initial point, $\wbf^*=\arg\min f(\wbf)$, and $\epsilon_Q$ be defined as in Theorem~\ref{thm:varbound}.

Suppose that  NUQSGD with momentum %
is executed for $T$ iterations with $\alpha=\min\{(1-\mu)/(2\beta),C/\sqrt{T+1}\}$ for some $C>0$ on $K$ processors, 
each with access to independent stochastic gradients of $f$ with a second-moment bound $B$.
Then  NUQSGD with momentum satisfies 
\begin{align}\label{nonconvNUQUMbound}
\min_{t=0,\cdots,T}\E[\|\nabla f(\wbf_t)\|^2]\leq \frac{2(f(\wbf_0)-f(\wbf^*))(1-\mu)}{\alpha(T+1)}+\frac{C}{(1-\mu)^3\sqrt{T+1}}\tilde{V}
\end{align} where $\tilde{V}=\beta \big(\mu^2((1-\mu)l-1)^2+(1-\mu)^2\big)(V^2+(1+\epsilon_Q) B/K)$.
\eth

In Appendices~\ref{app:asynch} and \ref{app:decentralized}, we extend our analysis to asynchronous and decentralized settings.

\subsection{NUQSGD vs QSGD and QSGDinf}
\begin{figure}[t]
\centerline{\includegraphics[width=0.8\columnwidth]{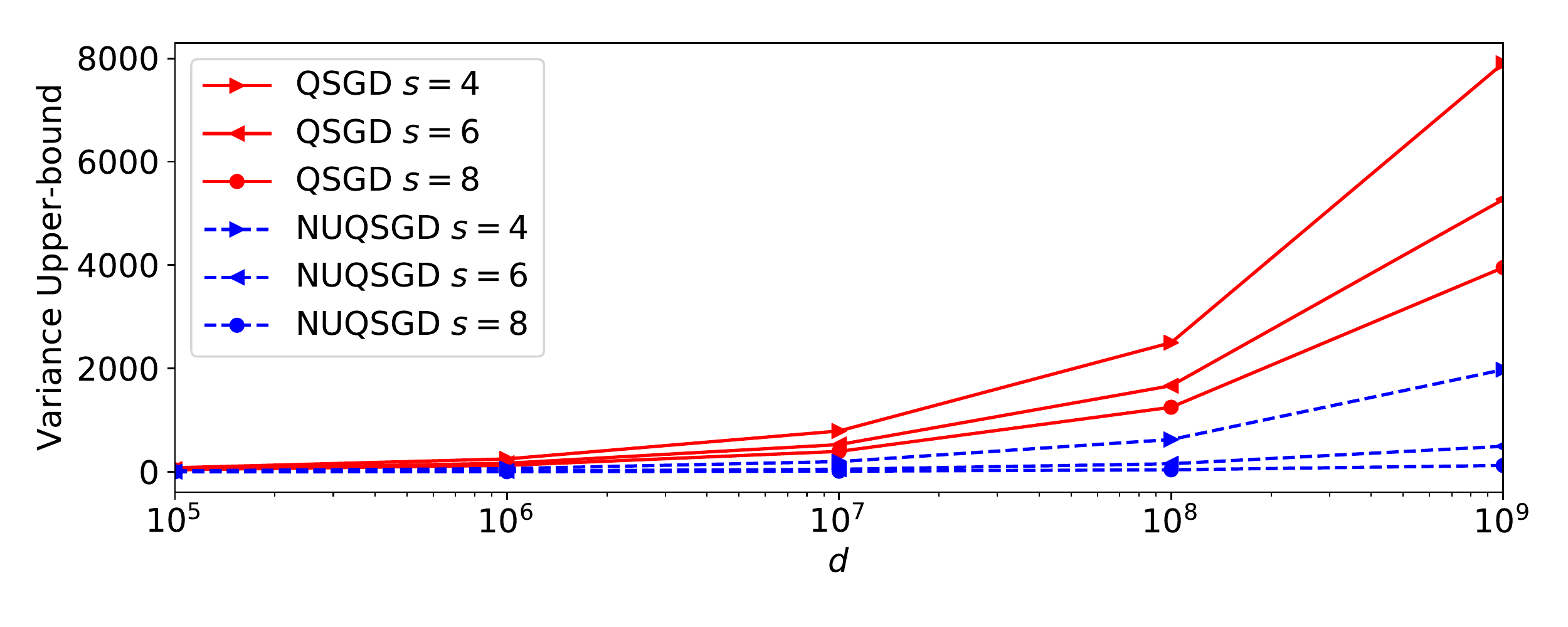}}
\caption{Variance upper bounds.}
\label{VarUB}
\end{figure}
How do QSGD and NUQSGD compare in terms of bounds on the expected number of communication bits required to achieve a given suboptimality gap $\epsilon$? The quantity that controls our guarantee on the convergence speed in both algorithms is the variance upper bound,
which in turn is controlled by the quantization schemes.
Note that the number of quantization levels, $s$, is usually a small number in practice. 
On the other hand, the dimension, $d$, can be very large, especially in overparameterized networks. 
In Figure \ref{VarUB}, we show that the quantization scheme underlying NUQSGD results in substantially smaller variance upper bounds 
for plausible ranges of $s$ and $d$. Note that these bounds do not make any assumptions on the dataset or the structure of the network. 

For any (nonrandom) number of iterations $T$, 
an upper bound, $\overline N_{A}$, holding uniformly over iterations $k \le T$ on the expected number of bits
used by an algorithm $A$ to communicate the gradient on iteration $k$,
yields an upper bound $T \overline N_{A}$, on the expected number of bits communicated over $T$ iterations by algorithm $A$.
Taking $T = T_{A,\epsilon}$ to be the (minimum) number of iterations needed to guarantee an expected suboptimality gap of $\epsilon$ based on the properties of $A$, 
we obtain an upper bound, $\zeta_{A, \epsilon} = T_{A, \epsilon} \overline N_{A}$, on the expected number of bits of communicated on a run expected to achieve a suboptimality gap of at most $\epsilon$.

\bth[Expected number of communication bits]\label{thm:totalbits}
Provided that $s= o(\log(d))$ and $\frac{2\hat B}{K\epsilon^2}>\frac{\beta}{\epsilon}$, 
$\zeta_{\NUQSGD,\epsilon} = O\big(\frac{1}{\epsilon^2}\sqrt{d(d-2^{2s})}\log\big(\frac{\sqrt{d}}{2^s}\big)\big)$ and 
$\zeta_{\QSGD,\epsilon} = O(\frac{1}{\epsilon^2}d\log\sqrt{d})$.
\eth

Focusing on the dominant terms in the expressions of overall number of communication bits required to guarantee a suboptimality gap of $\epsilon$, we observe that NUQSGD provides slightly stronger guarantees. Note that our stronger guarantees come without any assumption about the data. 

In Appendix~\ref{app:nuq_vs_qinf}, we show that there exist vectors for which the variance of quantization under NUQSGD is guaranteed to be smaller than that under QSGDinf.

\section{Experimental Evaluation}\label{sec:exp}

\begin{figure*}[t]
    \includegraphics[width=.49\textwidth]{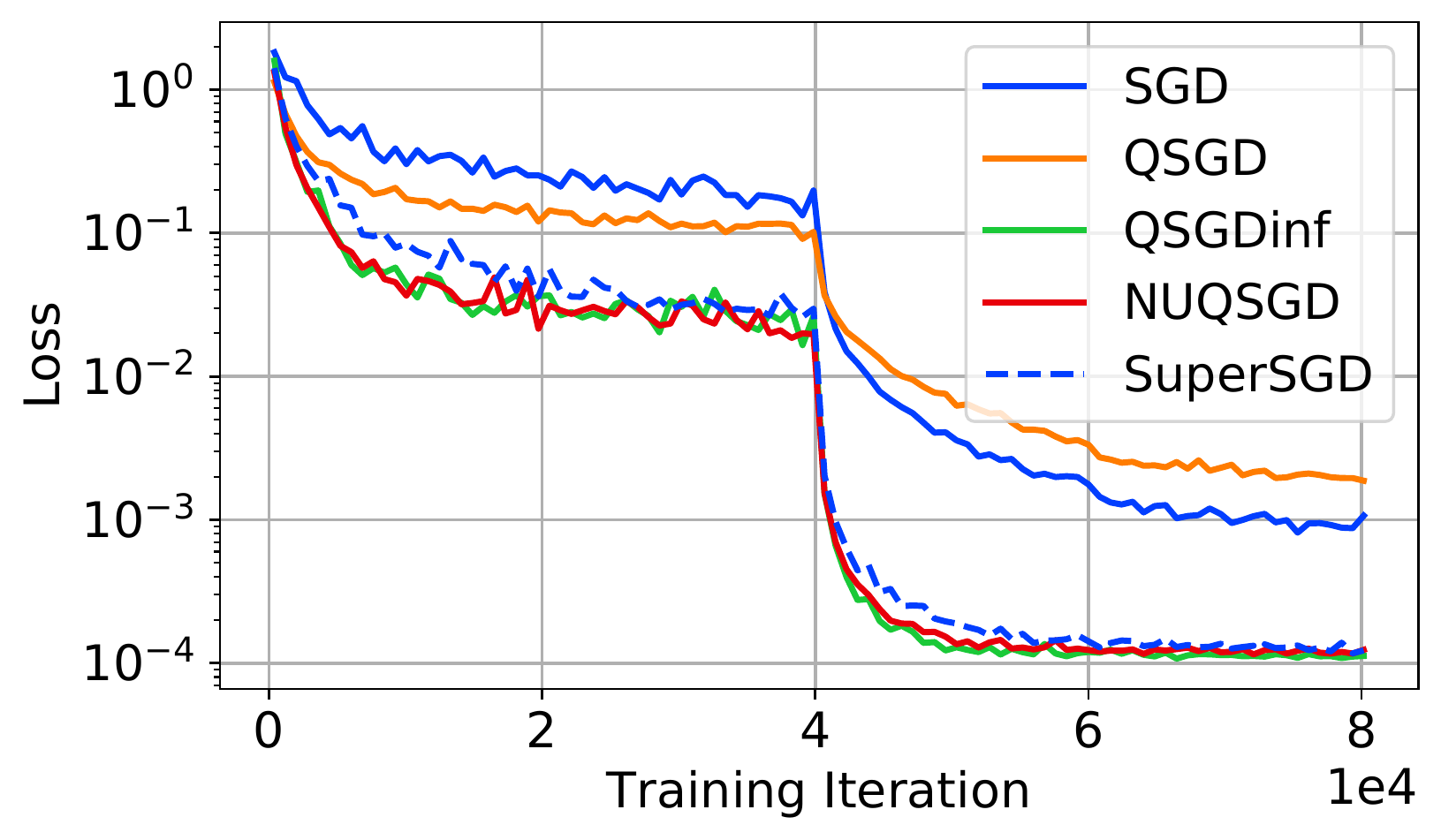}
    \hfill
    \includegraphics[width=.49\textwidth]{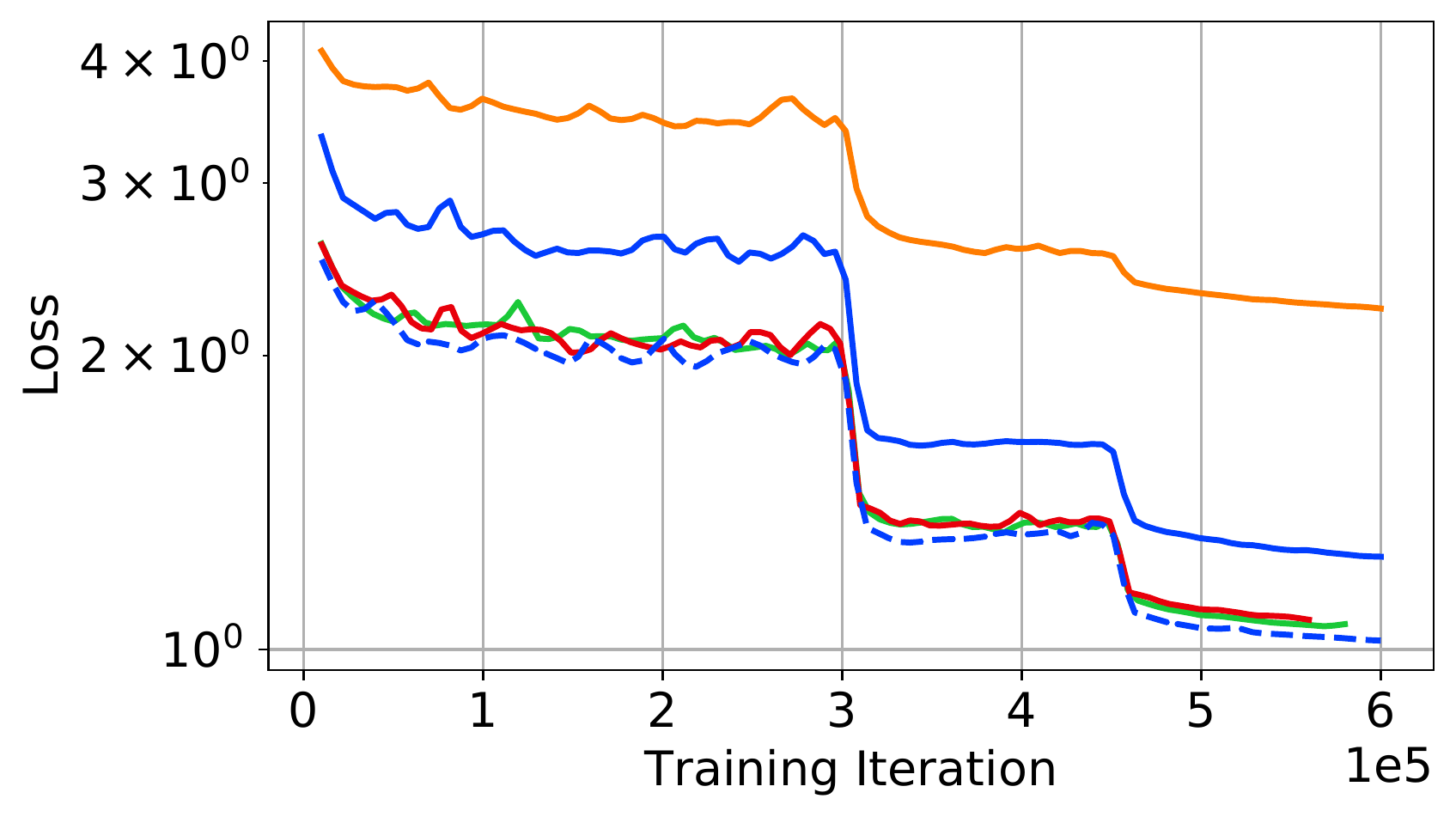}
    \caption{Training loss on CIFAR10 (left) and 
    ImageNet (right) for ResNet models. QSGD, QSGDinf, and NUQSGD are trained 
    by simulating the quantization and dequantizing of the gradients from 
    $8$-GPUs. On CIFAR10, SGD refers to the single-GPU training versus on 
    ImageNet it refers to $2$-GPU setup in the original ResNet paper.  SGD is 
    shown to highlight the significance of the gap between QSGD and QSGDinf.  
    SuperSGD refers to simulating full-precision distributed training without 
    quantization. SuperSGD is impractical in scenarios with limited bandwidth.}
    \label{fig:loss}
\end{figure*}

\begin{figure*}[t]
    \centerline{\includegraphics[width=.5\textwidth]{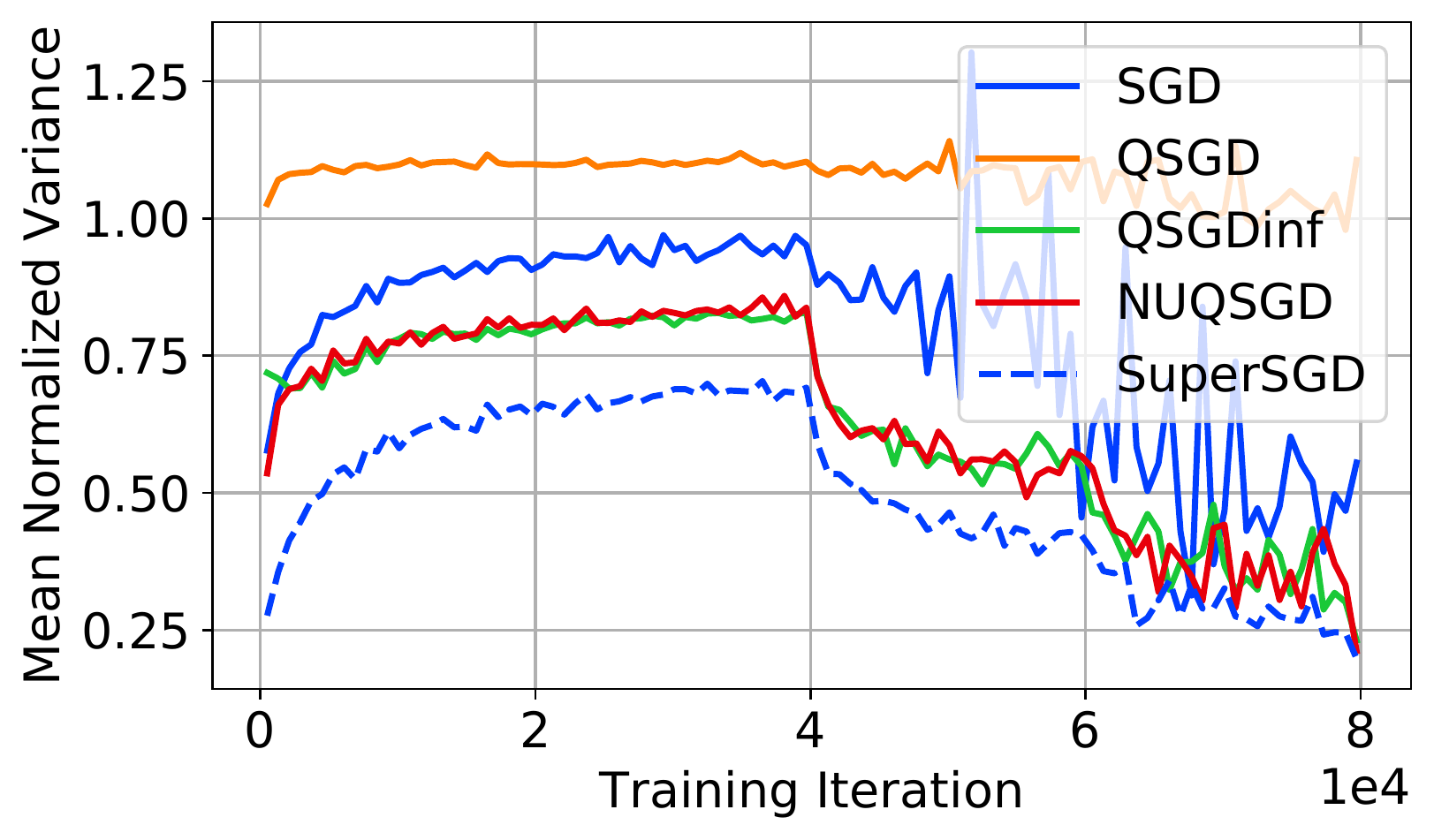}}
    \caption{Estimated normalized variance on CIFAR10 on the trajectory of 
    single-GPU SGD\@. Variance is measured for fixed model snapshots during 
    training.  Notice that the variance for NUQSGD and QSGDinf is lower than 
    SGD for almost all the training and it decreases after the learning rate 
    drops.}
    \label{fig:varCIFAR}
\end{figure*}

In this section, we examine the practical performance of NUQSGD in terms of both convergence (accuracy) and speedup.
The goal is to empirically examine whether NUQSGD can provide the similar performance and accuracy compared to the QSGDinf heuristic, which has no theoretical compression guarantees.\footnote{We also provide a generic implementation in Horovod~\citep{Horovod}, a communication back-end which can support a range of modern frameworks such as Tensorflow, Keras, Pytorch, and MXNet.}
For this, we implement and test these three methods (NUQSGD, QSGD, and QSGDinf), together with the distributed full-precision SGD baseline, which we call SuperSGD. 
Additionally, we will compare practical performance against a variant of SignSGD with EF~\citep{karimireddy19a}. 
We split our study across two axes: first, we validate our theoretical analysis by examining the variance induced by the methods, as well as their convergence in terms of loss/accuracy versus number of samples processed. 
Second, we provide an efficient implementation of all four methods in Pytorch using the Horovod communication back-end~\citep{Horovod}, a communication back-end supporting Pytorch, Tensorflow and MXNet. 
We adapted Horovod to efficiently support quantization and gradient coding, and examine speedup relative to the full-precision baseline. 
Further, we examine the effect of quantization on training performance by measuring loss, variance,  accuracy, and speedup   
for ResNet models~\citep{Resnet} applied to ImageNet and 
CIFAR10~\citep{CIFAR10}.

\paragraph{Convergence and Variance.} 
Our first round of experiments examine the impact of quantization on solution quality. 
We evaluate these methods on two image classification datasets: ImageNet and 
CIFAR10.  We train ResNet110 on CIFAR10 and ResNet18 on ImageNet with 
mini-batch size $128$ and base learning rate $0.1$.  In all experiments, 
momentum and weight decay are set to $0.9$ and $10^{-4}$, respectively.  The 
bucket size (quantization granularity) and the number of quantization bits are set to $8192$ and $4$, 
respectively. We observed similar trends in experiments with various 
bucket sizes and number of bits per entry.  
We simulate a scenario with $k$ GPUs for all 
three quantization methods by estimating the gradient from $k$ independent 
mini-batches and aggregating them after quantization and dequantization.

In Figure~\ref{fig:loss} (left and right), we show the training loss with $8$ 
GPUs. We observe that NUQSGD and  QSGDinf have lower training loss compared to 
QSGD on ImageNet. We observe a significant gap in training loss on CIFAR10, where 
the gap grows as training proceeds. We also observed similar performance gaps in 
test accuracy (provided in Appendix~\ref{app:exp}). In particular, unlike 
NUQSGD, QSGD does not achieve the test accuracy of full-precision SGD.  
Figure~\ref{fig:varCIFAR} shows the mean normalized variance of the 
gradient (defined formally in Appendix ~\ref{app:exp}) versus the training iterations, on the 
trajectory of single-GPU SGD on CIFAR10. These observations validate our 
theoretical results that NUQSGD has smaller variance for large models with 
small number of quantization bits, and support the intuition that this can impact solution quality. 

\begin{figure*}[t]
    \includegraphics[width=.49\textwidth]{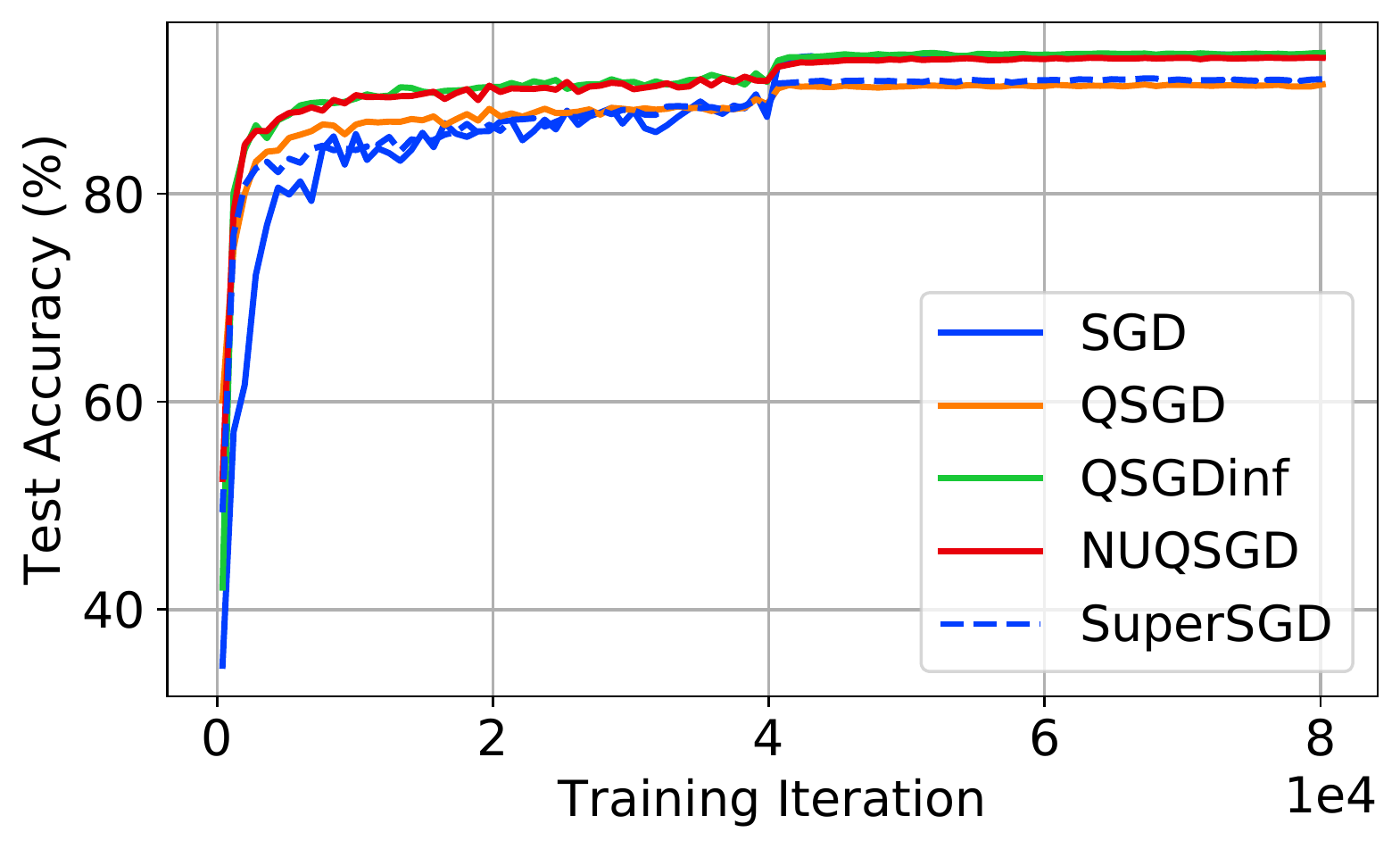}
    \hfill
    \includegraphics[width=.49\textwidth]{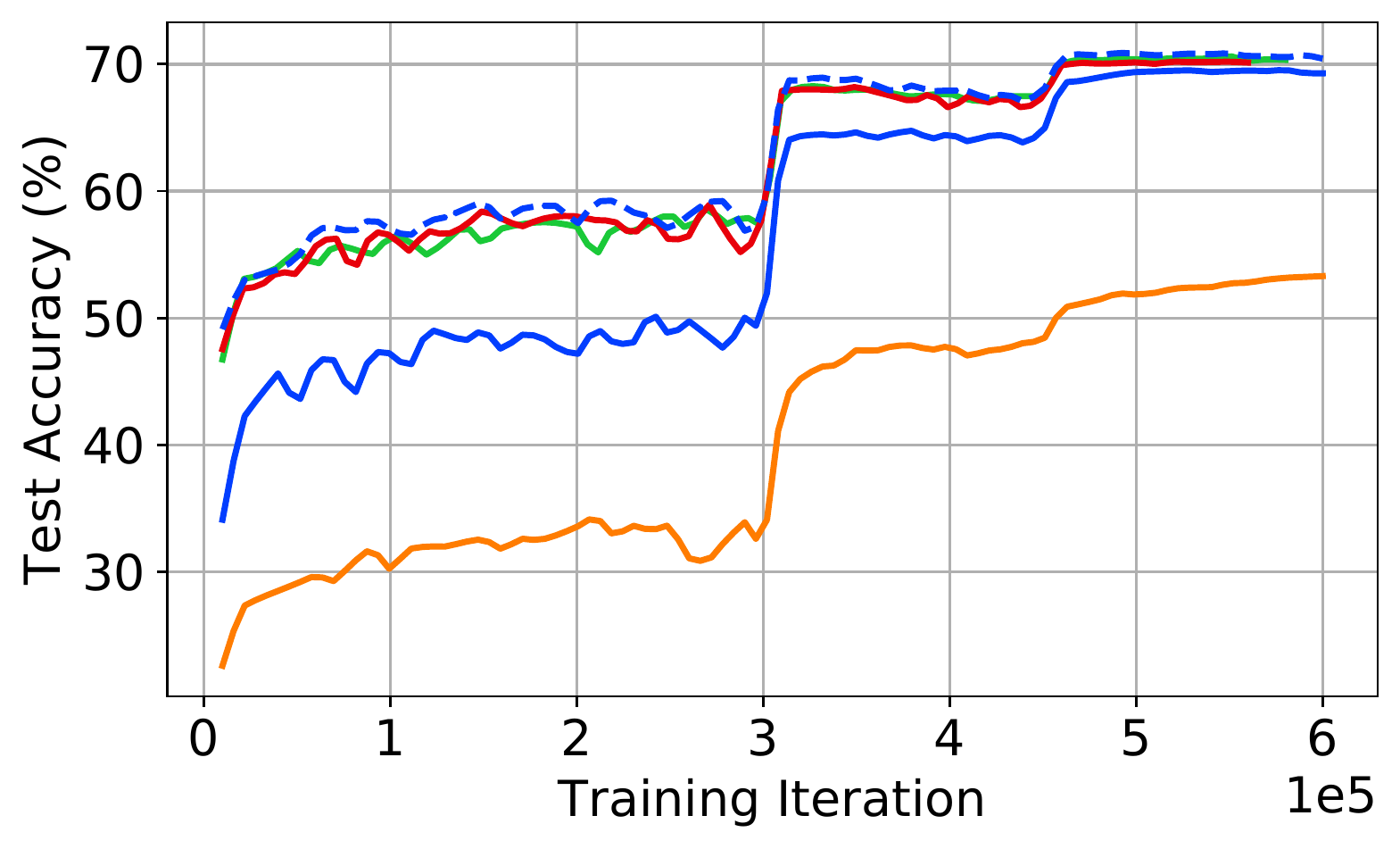}
    \caption{Accuracy on the hold-out set on CIFAR10 (left) and on ImageNet 
    (right) for training ResNet models from random initialization until 
    convergence. For CIFAR10, the hold-out set is the test set and for 
    ImageNet, the hold-out set is the validation set.}
    \label{fig:Vacc}
\end{figure*}

In Figure~\ref{fig:Vacc}, we show the test accuracy for training ResNet110 on 
CIFAR10 and validation accuracy for training ResNet34 on ImageNet from random 
initialization until convergence. Similar 
to the training loss performance, we observe that NUQSGD and QSGDinf outperform 
QSGD in terms of test accuracy in both experiments. In both experiments, unlike 
NUQSGD, QSGD does not recover the test accuracy of SGD\@. The gap between 
NUQSGD and QSGD on ImageNet is significant. We argue that this is achieved 
because NUQSGD and QSGDinf have lower variance relative to QSGD\@. It turns out 
both training loss and generalization error can benefit from the reduced 
variance.

\paragraph{Efficient Implementation and Speedup.}
To examine speedup behavior, we implemented all quantization methods in Horovod~\citep{Horovod}. 
Doing so efficiently requires non-trivial refactoring of this framework, since it does not support communication compression---our framework will be open-sourced upon publication. 
For performance reasons, our implementation diverges slightly from the theoretical analysis. 
First, the Horovod framework applies ``tensor fusion'' to multiple layers, by merging the resulting gradient tensors for more efficient transmission. 
This causes the gradients for different layers to be quantized together, which can lead to loss of accuracy (due to e.g. different normalization factors across the layers). 
We addressed this by tuning the way in which tensor fusion is applied to the layers such that it minimizes the accuracy loss. 
Second, we noticed that quantizing the gradients corresponding to the biases has an adverse effect on accuracy; since the communication impact of biases is negligible, we transmit them at full precision. 
We apply these steps to all methods. 
We implemented compression and de-compression via efficient CUDA kernels.

\paragraph{Efficient Encoding.}
One of the advantages of NUQSGD is that it provides both good practical accuracy, as well as bounds on the code-length of the transmitted gradients in actual executions. 
This should be contrasted with QSGD (which provides such bounds, but, as seen above,  loses accuracy), and QSGDinf, whose analysis no longer implies any bounds on the expected code-length. 
(The QSGD analysis is only performed for L2 normalization.)
We leverage this fact by employing Huffman coding on an initial sample of gradients, in order to determine a non-trivial encoding of the integer levels sent. We then employ this coding for the rest of the execution. This variant is called encoded NUQSGD.

\begin{figure*}[t]
    \includegraphics[width=.49\textwidth]{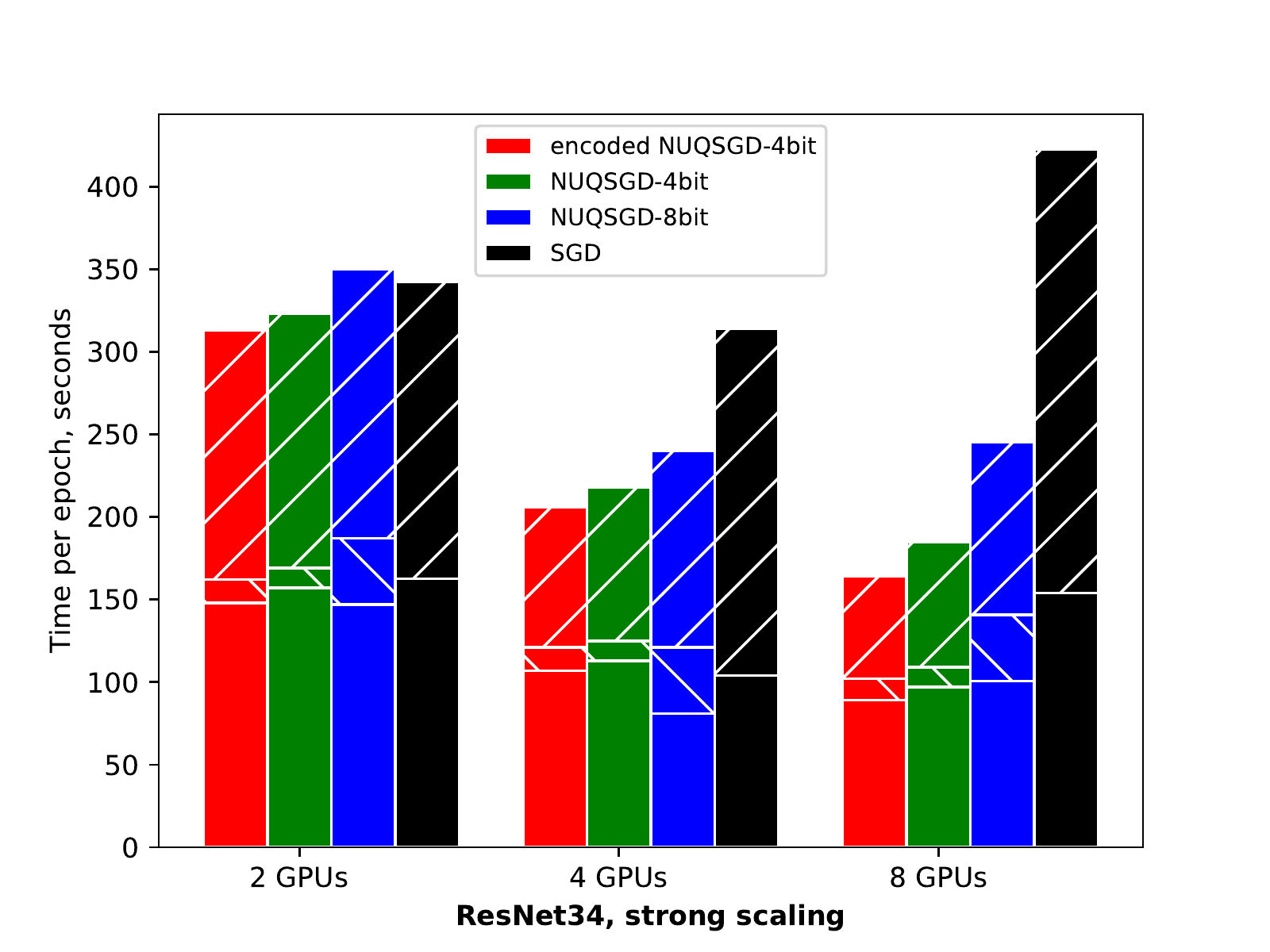}
    \hfill
    \includegraphics[width=.49\textwidth]{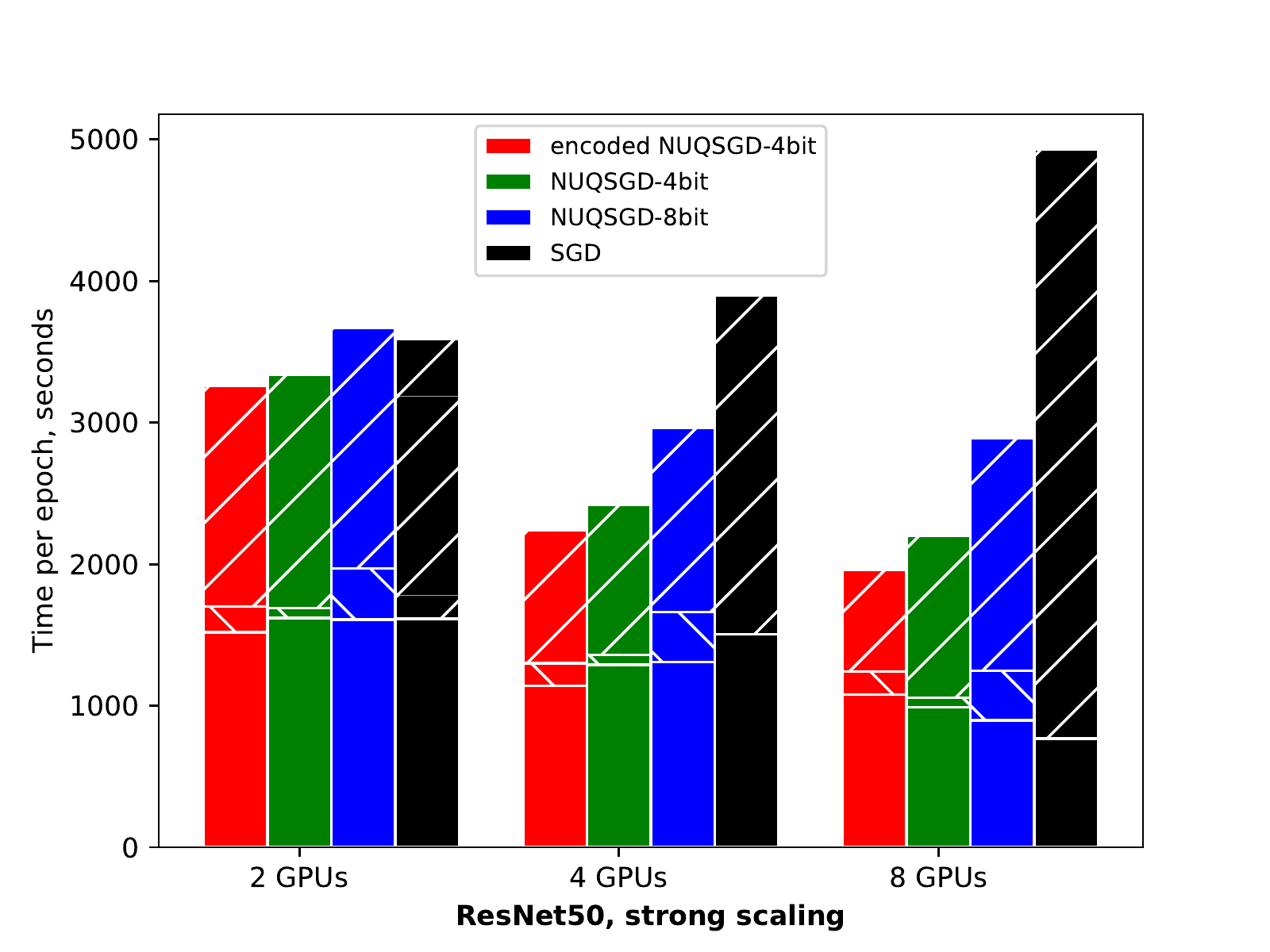}
    \caption{Scalability behavior for NUQSGD versus the full-precision baseline when training ResNet34 (left) and ResNet50 (right) on ImageNet. 
    Both experiments examine \emph{strong scaling}, splitting a global batch of size $256$ onto the available GPUs, for 2, 4, and 8 nodes. 
    Each time bar is split into computation (bottom), encoding cost (middle), and transmission cost (top). 
    Notice the negative scalability of the SGD baseline in both scenarios. By contrast, the 4-bit communication-compressed implementation achieves positive scaling, while the 8-bit variant stops scaling between 4 and 8 nodes due to the higher communication and encoding costs, while the encoding variant offers further compression.}
    \label{fig:scalability}
\end{figure*}

\begin{figure*}[t]
    \centerline{\includegraphics[width=.5\textwidth]{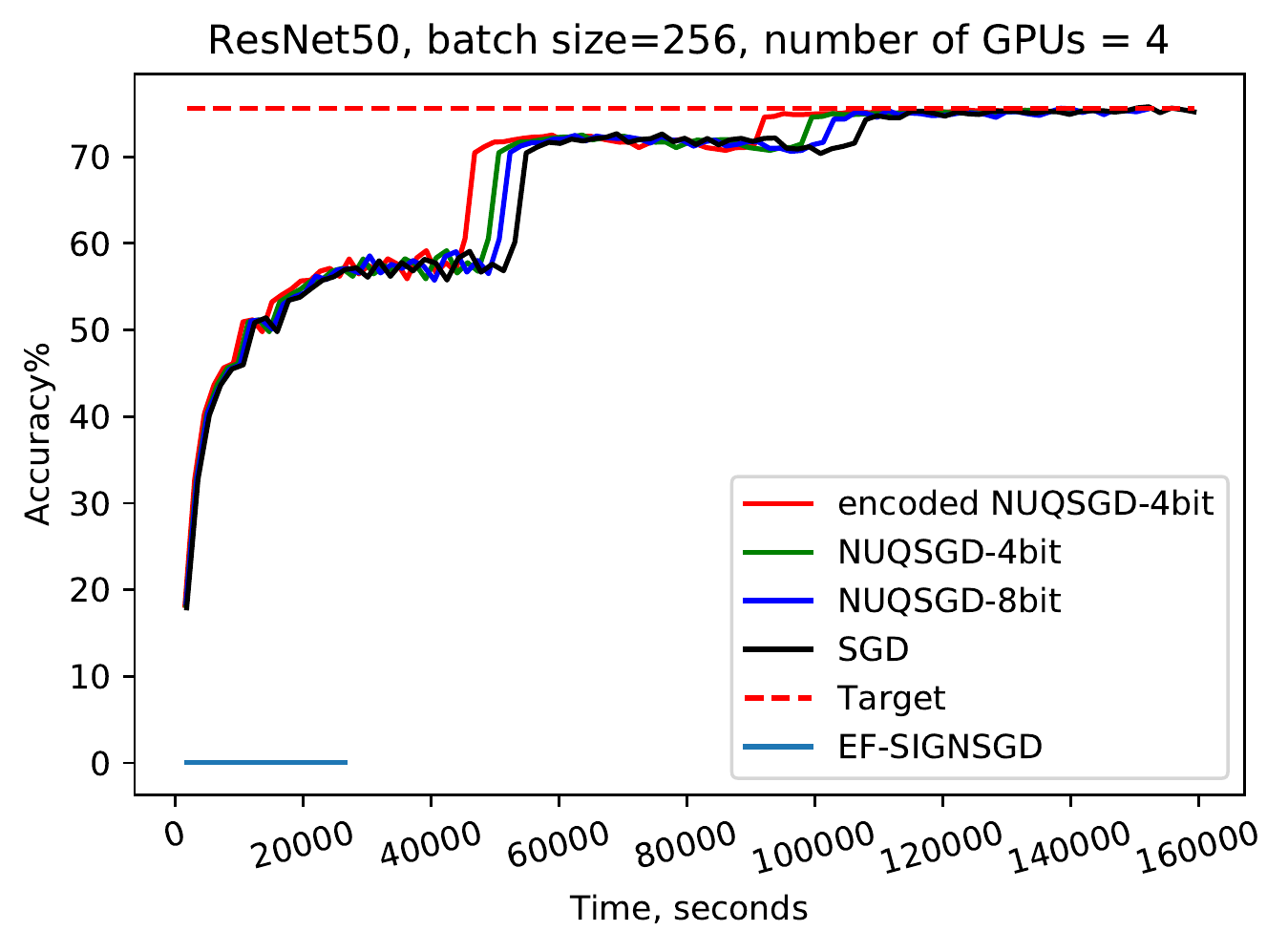}}
    \caption{End-to-end training time for ResNet50/ImageNet for NUQSGD and EF-SignSGD (which diverges) versus the SuperSGD baseline.} %
    \label{fig:end-to-end}
\end{figure*}

Our baselines are full-precision SGD (SuperSGD), EF-SignSGD~\citep{karimireddy19a}, and the QSGDinf heuristic, which we compare against the 4-bit and 8-bit NUQSGD variants executing the same pattern. 
The implementation of the QSGDinf heuristic provides virtually identical convergence numbers, and is sometimes omitted for visibility. QSGD yields inferior convergence on this dataset and is therefore omitted. 
All variants are implemented using a standard all-to-all reduction pattern. 
Figure~\ref{fig:scalability} (left and right) shows the execution time per epoch for ResNet34 and ResNet50 models on ImageNet, on a cluster machine with 8 NVIDIA 2080 Ti GPUs, for the hyper-parameter values quoted above.\footnote{We use the following hardware:  CPU information: Intel(R) Xeon(R) Silver 4214 CPU @ 2.20GHz, 24 cores. GPU2GPU Bandwidth: unidirectional 10GB/s, Bidirectional 15GB/s.}

The results confirm the efficiency and scalability of the compressed variant, mainly due to the reduced communication volume. We note that the overhead of compression and decompression is less than $1\%$ of the batch computation time for vanilla NUQSGD, and of $<4\%$ for the gradient coding variant. We also note that the smallest reduction times are obtained by encoded NUQSGD, which send on average approximately $2.7$ bits per component, a reduction of $67\%$ over standard NUQSGD. 

Figure~\ref{fig:end-to-end} presents end-to-end speedup numbers (time versus accuracy) for ResNet50 on ImageNet, executed on 4 GPUs, under the same hyperparameter settings as the full-precision baseline, with bucket size $512$.  
First, notice that all NUQSGD variants match the target accuracy of the 32-bit model, with non-trivial speedup over the standard data-parallel variant, directly proportional to the per-epoch speedup. 
The QSGDinf heuristic yields similar accuracy and performance, and is therefore omitted. 
We found that unfortunately EF-SignSGD does not converge under these standard hyperparameter settings. 
To address this issue, we performed a non-trivial amount of hyperparameter tuning for this algorithm: in particular, we found that the scaling factors and the bucket size must be carefully adjusted for convergence on ImageNet. 
We were able to recover full accuracy with EF-SignSGD on ResNet50, but at the cost of quantizing into buckets of size $64$. 
In this setting, the algorithm sends $32$ bits of scaling data for every $64$ entries, and the GPU implementation becomes less efficient due to error computation and reduced parallelism. 
The end-to-end speedup of this tuned variant is inferior to baseline 4-bit NUQSGD, and only slightly superior to that of 8-bit NUQSGD. 
Please see Figure~\ref{fig:RN50_sign} in the Appendix~\ref{app:exp} and the accompanying text for details. 

We therefore conclude that NUQSGD offers competitive or superior performance w.r.t. QSGDinf and EF-SignSGD, while providing strong convergence guarantees, and that it allows additional savings due to the fact that gradients can be encoded.

\paragraph{Comparison with DGC.} 
In this section, we make a comparison with DGC~\citep{DGC}. DGC is essentially a sparsification method, which leverages momentum correction and local gradient clipping to recover accuracy.
\begin{figure*}[t]
    \includegraphics[width=.49\textwidth]{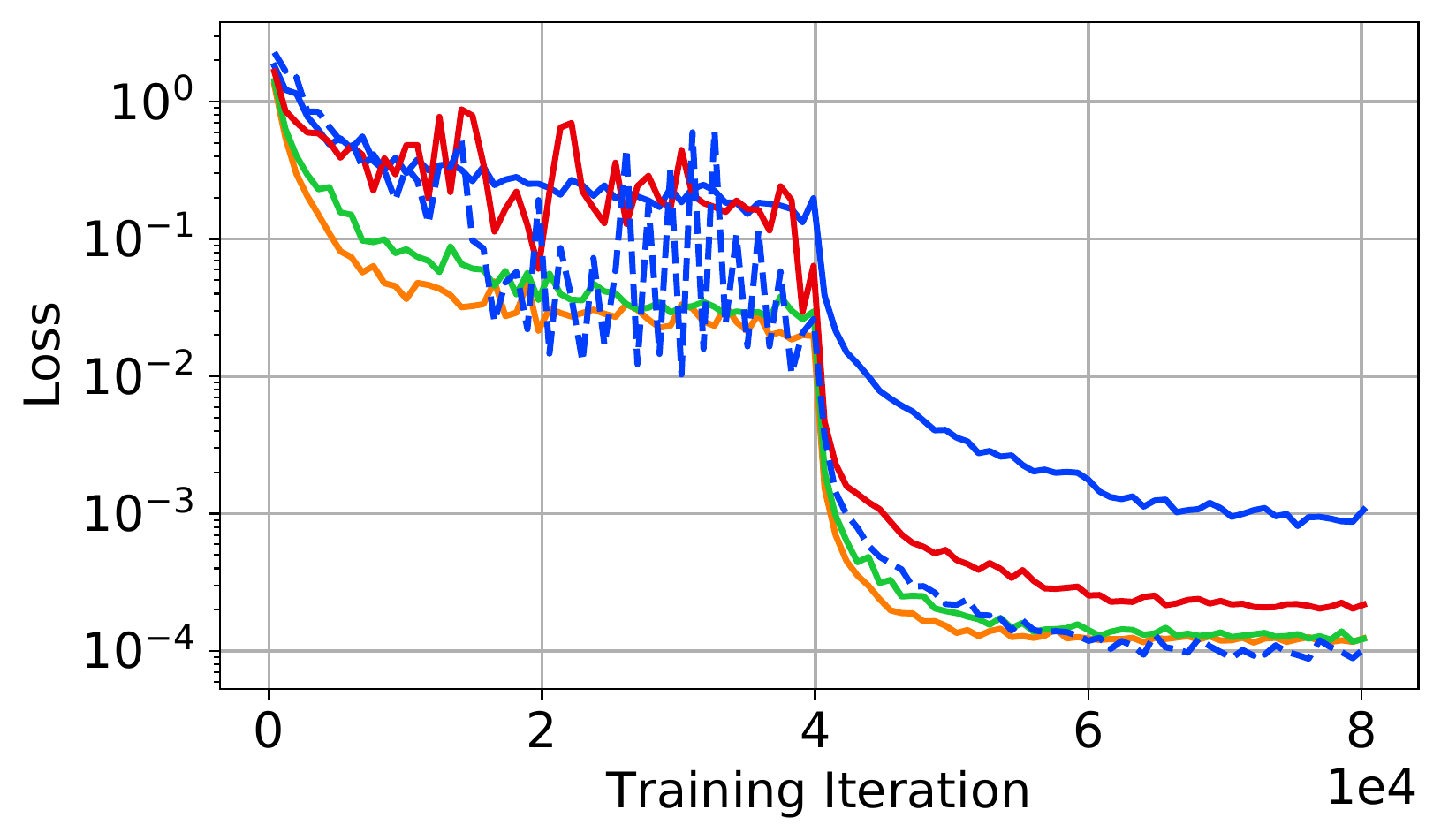}
    \hfill
    \includegraphics[width=.49\textwidth]{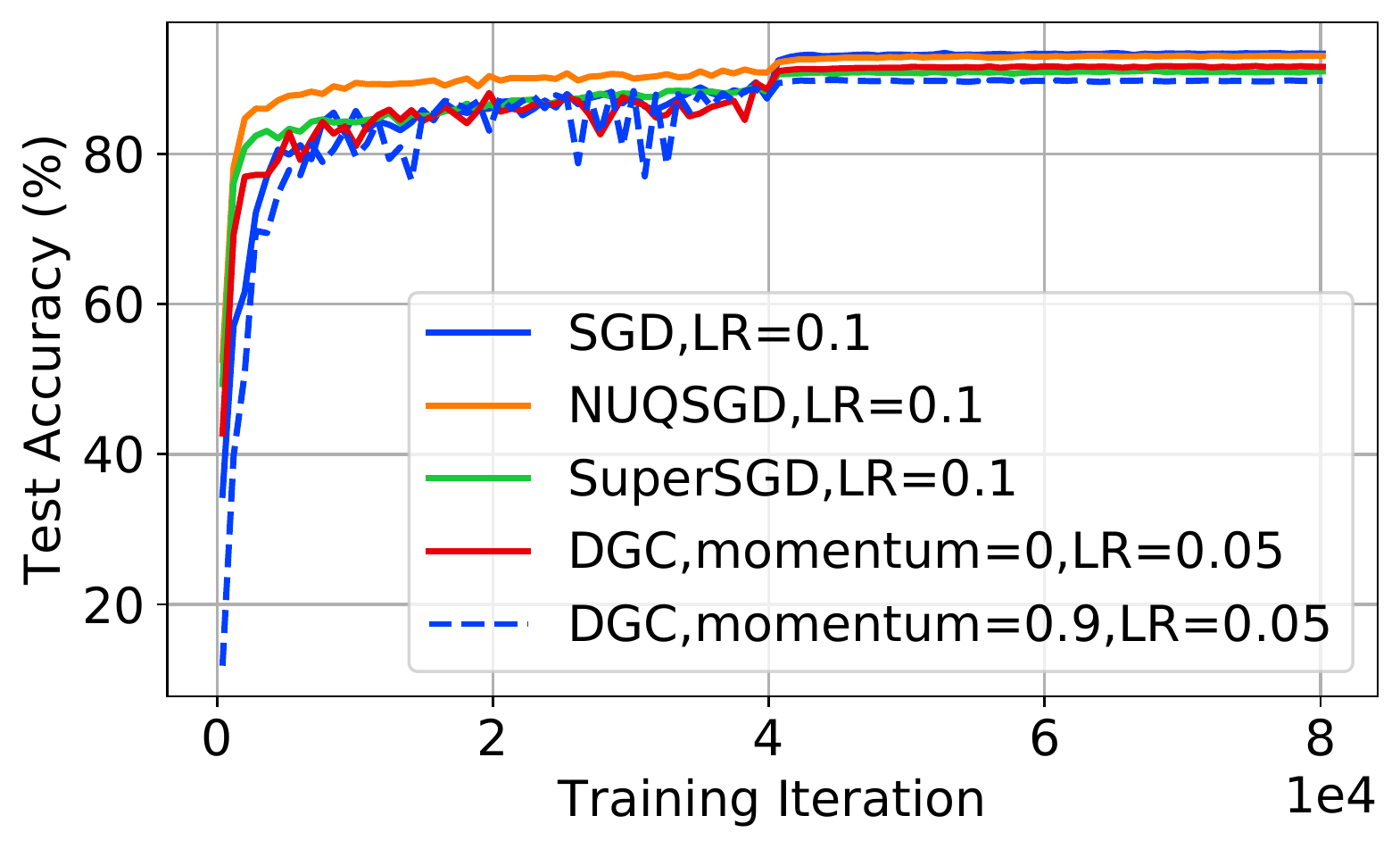}
    \caption{Training loss (left) and Test accuracy (right) on CIFAR10 for ResNet110. 
    We set the ratio of compression for DGC to be roughly the same as NUQSGD.
    In particular, we compare compression methods at compression ratio = $4/32$, \ie NUQSGD with 
    4 bits and DGC at 12.5\% compression. At this rate, both methods will have approximately the same 
    communication cost, \ie comparison in simulation is representative of real-time performance. 
    We tune the learning rate and momentum for DGC and show its best performance.}
    \label{fig:DGCsim}
\end{figure*}

\begin{figure*}[t]
    \includegraphics[width=.49\textwidth]{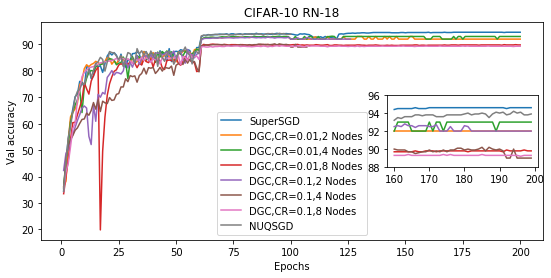}
    \hfill
    \includegraphics[width=.49\textwidth]{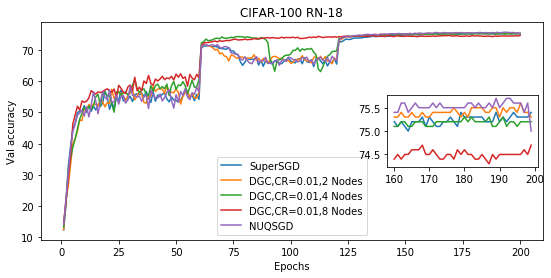}
    \caption{Test Accuracy on CIFAR10  (left) and CIFAR100 (right)  for ResNet18. 
    For NUQSGD and  SuperSGD, we use 8 GPUs. For NUQSGD, we use the same hyperparameters
    that are tuned for the full-precision baseline. However, we tune hyperparameters for DGC
    at 1\% and 10\% compression ratios.}
    \label{fig:DGCRN18}
\end{figure*}

In Figure~\ref{fig:DGCsim}, we show simulation results to compare convergence and generalization of DGC with those of NUQSGD and full-precision SGD. We set the ratio of compression for DGC to be roughly the same as NUQSGD. In particular, we compare compression methods at compression ratio = 4/32, \ie NUQSGD with 4 bits and DGC at 12.5\% compression. At this rate, both methods will have approximately the same communication cost, \ie comparison in simulation is representative of real-time performance. We show the results for NUQSGD, DGC, and full-precision baselines on CIFAR10.  We tune the learning rate and momentum for DGC and show its best performance. 

Our results show that by tuning DGC carefully, it can achieve full-precision performance albeit some noisy curves at the beginning. On the other hand, NUQSGD provides communication efficiency on the fly where practitioners can reuse the same hyperparameters that are tuned for full-precision schemes. NUQSGD is also easier to implement efficiently in practice. Furthermore, NUQSGD enjoys strong theoretical guarantees. Finally, we note that NUQSGD can be used as the encoding function on top of DGC to further reduce its communication costs. Hence, DGC and NUQSGD can be viewed as complementary methods for practitioners. Additional simulation results are presented in Appendix~\ref{app:exp}.

To further evaluate the performance of DGC over different numbers of GPUs, compression ratios, and datasets, in Figure~\ref{fig:DGCRN18}, we show validation accuracy when we train ResNet18 on CIFAR10 and CIFAR100. For NUQ and  the full-precision baseline (SuperSGD), we use 8 GPUs. For NUQSGD, we use the same hyperparameters that are tuned for the full-precision baseline. However, we tune hyperparameters for DGC at 1\% and 10\% compression ratios. We note that unlike NUQSGD, DGC’s performance degrades as we increase the number of GPUs. We observe DGC is less stable for simpler datasets such as CIFAR10. In sum, while NUQSGD offers lower \textit{peak} compression relative to DGC, NUQSGD is more practical. Additionally, NUQSGD offers strong theoretical guarantees relative to DGC.

\paragraph{Comparison with TernGrad and ATOMO.} We ported the original ATOMO~\citep{ATOMO} and TernGrad~\citep{TernGrad} code to our framework. In Table~\ref{Tab:compTRN_ATOMO}, we present results of training ResNet20 on CIFAR10 on Google Cloud (4 nodes in a bandwidth-constrained setting) under the standard training parameters. Our experiments show that NUQSGD achieves slightly higher accuracy than ATOMO at a slightly higher compression ratio. (We did not use Huffman coding here.) ATOMO’s time/step is high due to its computation cost (we have used its original implementation, but further optimizations might be possible). Unfortunately, TernGrad did not converge for the standard (sequential) hyperparameters. \citet{DGC} showed that TernGrad loses accuracy at ImageNet scale. Its average compression ratio (2 bits/entry) is similar to that of NUQSGD with Huffman coding (approximately 2.4 bits/entry).

\begin{table}
    \centering
     \begin{tabular}{c|c|c|c}%
\toprule
Method        &  Accuracy  & Time/step (s) & Compression\% \\
\midrule
   
SuperSGD & 91.5      &       205           &          100\\
NUQSGD (3bit) &    90.94     &       85            &          9\\ 
ATOMO (rank=3) &      90.5     &      297           &           17\\
TernGrad      & \multicolumn{3}{c}{did not converge for standard hyperparameters}\\
\bottomrule      

    \end{tabular}
    \captionof{table}{Training ResNet20 on CIFAR-10 using on 4 nodes under standard training parameters.}
    \label{Tab:compTRN_ATOMO} 
\end{table}

\section{Conclusions}

We study data-parallel and communication-efficient version of stochastic gradient descent. Building on QSGD \citep{QSGD}, we study a nonuniform quantization scheme. We establish upper bounds on the variance of nonuniform quantization and the expected code-length. The former decreases as the number of quantization levels increases, while the latter increases with the number of quantization levels. Thus, this scheme provides a trade-off between the communication efficiency and the convergence speed. We compare NUQSGD and QSGD in terms of their variance bounds and the expected number of communication bits required to meet a certain convergence error, and show that NUQSGD provides stronger guarantees.  
Experimental results are consistent with our theoretical results and confirm that NUQSGD matches the performance of QSGDinf when applied to practical deep models and datasets including ImageNet. Thus, NUQSGD closes the gap between the theoretical guarantees of QSGD and empirical performance of QSGDinf.
One limitation of our study which we aim to address in future work is that we focus on all-to-all reduction patterns, which interact easily with communication compression. 
In particular, we aim to examine the interaction between more complex reduction patterns, such as ring-based reductions~\citep{hannun2014deep}, which may yield superior performance in bandwidth-bottlenecked settings, but which interact with communication-compression in non-trivial ways, since they may lead a gradient to be quantized at each reduction step. 

\section*{Acknowledgement}
Ramezani-Kebrya was supported by an NSERC Postdoctoral Fellowship. Faghri was supported by an OGS Scholarship. Alistarh, Markov, and Aksenov were supported by the European Research Council (ERC) under the European Union's Horizon 2020 research and innovation
programme (grant agreement No 805223 ScaleML). Roy was supported by an NSERC Discovery Grant. Resources used in preparing this
research were provided, in part, by the Province of Ontario, the Government of Canada through CIFAR, and companies sponsoring the Vector Institute.\footnote{\url{www.vectorinstitute.ai/\#partners}}

\bibliography{RefNUQSGD}

\newpage

\begin{appendices}

\section{Encoding}\label{app:elias}
\begin{algorithm}[t]
\small
\SetAlgoLined
\KwEn{}
Place a 0 at the end of the string\;
	\eIf{$N==1$}{
	Stop\;
	}{
	Prepend $\text{binary}(N)$ to the beginning \tcp*[l]{Let $N'$ denote \# bits prepended minus 1}
	Encode $N'$ recursively\;
	}
\KwDe{}
Start with $N=1$\;
	\eIf{$\text{the next bit}==0$}{
	Stop and return $N$\;
	}{
	Read that bit plus $N$ following bits and update $N$\; 
	}
\caption{Elias recursive coding produces a bit string encoding of positive integers.}%
\label{Eliascode}
\end{algorithm}

By inspection, the quantized gradient $Q_s(\vbf)$ is determined by the tuple 
$(\|\vbf\|,\rhobf,\hbf)$, 
where $\|\vbf\|$ is the norm of the gradient,
$\rhobf\defeq[\text{sign}(v_1),\cdots,\text{sign}(v_d)]^T$ is the vector of signs of the coordinates $v_i$'s,
and 
$\hbf\defeq[h_1(\vbf,s),\cdots,h_d(\vbf,s)]^T$ are the quantizations of the normalized coordinates.
We can describe the $\ENCODE$ function (for Algorithm~\ref{NUQSGDalg}) 
in terms of the tuple $(\|\vbf\|,\rhobf,\hbf)$ and an
encoding/decoding scheme $\Elias : \{1,2,\cdots\} \to \{0,1\}^*$ and $\Elias^{-1} : \{0,1\}^* \to \{1,2,\cdots\}$ for encoding/decoding positive integers.
The encoding, $\ENCODE(\vbf)$, of a stochastic gradient is as follows:
We first encode the norm $\|\vbf\|$ using $b$ bits where, in practice, we use standard 32-bit floating point encoding.
We then proceed in rounds, $r=0,1,\cdots$. 
On round $r$, having transmitted all nonzero coordinates up to and including $t_r$, 
we transmit $\Elias(i_r)$ where $t_{r+1} = t_r + i_r$ is either 
(i) the index of the first nonzero coordinate of $\hbf$ after $t_r$ (with $t_0=0$) or
(ii) the index of the last nonzero coordinate.
In the former case, we then transmit one bit encoding the sign $\rho_{t_{r+1}}$, transmit 
$\Elias( \log(2^{s+1} h_{t_{r+1}}) )$, and proceed to the next round.
In the latter case, the encoding is complete after transmitting  $\rho_{t_{r+1}}$ and 
$\Elias( \log(2^{s+1} h_{t_{r+1}}))$.

The DECODE function (for Algorithm~\ref{NUQSGDalg}) simply reads $b$ bits to reconstruct $\|\vbf\|$. 
Using $\Elias^{-1}$, it decodes the index of the first nonzero coordinate, reads the bit indicating the sign, and then uses $\Elias^{-1}$ again to determines the quantization level of this first nonzero coordinate. 
The process proceeds in rounds, mimicking the encoding process, finishing when all coordinates have been decoded.

Like \citet{QSGD}, we use Elias recursive coding \citep[ERC]{Elias} to encode positive integers.
ERC is simple and has several desirable properties, including 
the property that the coding scheme assigns shorter codes to smaller values,
which makes sense in our scheme as they are more likely to occur. 
Elias coding is a universal lossless integer coding scheme with a recursive encoding and decoding structure.

The Elias recursive coding scheme is summarized in Algorithm \ref{Eliascode}. For any positive integer $N$, the following results are known for ERC \citep{QSGD}: 
\begin{enumerate}
\item $|\Elias(N)|\leq \big(1+o(1)\big)\log N+1$;
\item $\Elias(N)$ can be encoded and decoded in time $O(|\Elias(N)|)$;
\item Decoding can be done without knowledge of an upper bound on $N$. 
\end{enumerate}
 
\section{Proof of Theorem \ref{thm:varbound} (Variance Bound)}\label{app:pr_var}
We first find a simple expression of the variance of $Q_s(\vbf)$ for every arbitrary quantization scheme in the following lemma: 
\blm\label{lm:genvar}
Let $\vbf\in\reals^d$, $\Lc=(l_0,l_1,\cdots,l_{s+1})$, and fix $s\geq 1$. The variance of $Q_s(\vbf)$ for general sequence of quantization levels is given by 
\begin{align}\label{genvar}
\E[\|Q_s(\vbf)-\vbf\|^2]=\|\vbf\|^2\sum_{i=1}^d\tau^2(r_i) p(r_i)\big(1-p(r_i)\big)
\end{align} where $r_i=|v_i|/\|\vbf\|$ and $p(r),\level{r},\tau(r)$ are defined in Section \ref{sec:nonunif}.%
\elm 
\bpr
Noting the random quantization is i.i.d over elements of a stochastic gradient, we can decompose $\E[\|Q_s(\vbf)-\vbf\|^2]$ as: 
\begin{align}\label{genvar_dec}
\E[\|Q_s(\vbf)-\vbf\|^2]=\sum_{i=1}^d \|\vbf\|^2\sigma^2(r_i)  
\end{align} where $\sigma^2(r_i)=\E[(h_i(\vbf,s)-r_i)^2]$.
Computing the variance of $h_i(\vbf,s)$, we can show that $\sigma^2(r_i)=\tau^2(r_i) p(r_i)\big(1-p(r_i)\big)$. 
\epr

In the following, we consider NUQSGD algorithm with 
$\hat\Lc=(0,1/{2^s},\cdots,{2^{s-1}}/{2^s},1)$ as the quantization levels.
Then, $h_i(\vbf,s)$'s are defined in two cases based on which quantization interval $r_i$ falls into: 

1) If $r_i\in[0,2^{-s}]$, then 
\begin{align}\label{h_case1}\textstyle
h_i(\vbf,s)=
  \Bigg\{\begin{array}{ll}
        0 & \text{with probability}~1-p_1(r_i,s);\\
       2^{-s} & \text{otherwise}
        \end{array}
\end{align} where $p_1\big(r,s\big)=2^sr.$

2) If $r_i\in[2^{j-s},2^{j+1-s}]$ for $j=0,\cdots,s-1$, then 
\begin{align}\label{h_case2}\textstyle
h_i(\vbf,s)=
  \Bigg\{\begin{array}{ll}
        2^{j-s} & \text{with probability}~1-p_2(r_i,s);\\
       2^{j+1-s} & \text{otherwise}
        \end{array}
\end{align}
where $p_2\big(r,s\big)=2^{s-j}r-1.$ Note that $Q_s(\mathbf{0})=\mathbf{0}$. 
     
Let $\Sc_j$ denote the coordinates of vector $\vbf$ whose elements fall into the $(j+1)$-th bin, \ie $\Sc_0\defeq\{i:r_i\in[0,2^{-s}]\}$ and $\Sc_{j+1}\defeq\{i:r_i\in[2^{j-s},2^{j+1-s}]\}$ for $j=0,\cdots,s-1$. Let $d_j\defeq |\Sc_j|$.    
Applying the result of Lemma \ref{lm:genvar}, we have 
\begin{align}
\begin{split}\label{vareq}
\E[\|Q_s(\vbf)-\vbf\|^2]&=\|\vbf\|^2\tau_0^2\sum_{i\in\Sc_0}p_1(r_i,s)(1-p_1(r_i,s))\\
&\quad+\|\vbf\|^2\sum_{j=0}^{s-1}\tau_{j+1}^2\sum_{i\in\Sc_{j+1}}p_2(r_i,s)\big(1-p_2(r_i,s)\big)
\end{split}
\end{align}
where $\tau_j\defeq l_{j+1}-l_j$ for $j\in\{0,\cdots,s\}$.

The variance of $Q_s(\vbf)$ can be also expressed as   
\begin{align}\label{var_reform}
\E[\|Q_s(\vbf)-\vbf\|^2]=\|\vbf\|^2\big(\sum_{r_i\in\Ic_0}(2^{-s}-r_i)r_i+\sum_{j=0}^{s-1}\sum_{r_i\in\Ic_{j+1}}(2^{j+1-s}-r_i)(r_i-2^{j-s})\big).
\end{align} where $\Ic_0\defeq [0,2^{-s}]$ and $\Ic_{j+1}\defeq[2^{j-s},2^{j+1-s}]$ for $j=0,\cdots,s-1$.

Inspired by the proof in \citep{Samuel}, we can find $k$ that satisfies $(2^{j+1-s}-r)(r-2^{j-s})\leq kr^2$ for $r\in \Ic_{j+1}$. Expressing $r=2^{j-s}\theta$, we can find $k$ through solving 
\begin{align}\label{k_for_r2}
k = \max_{1\leq\theta\leq 2} (2-\theta)(\theta-1)/\theta^2=1/8. 
\end{align}

Substituting \eqref{k_for_r2} into \eqref{var_reform}, an upper bound on $\E[\|Q_s(\vbf)-\vbf\|^2]$ is given by 
\begin{align}\nn
\E[\|Q_s(\vbf)-\vbf\|^2]\leq \|\vbf\|^2(1/8+\sum_{r_i\in\Ic_0}(2^{-s}-r_i)r_i). 
\end{align}

In the following, we derive three different bounds on $\sum_{r_i\in\Ic_0}(2^{-s}-r_i)r_i$, each gives us an upper bound on   $\E[\|Q_s(\vbf)-\vbf\|^2]$. 

\blm\label{lm:K_p}
Let $p\in (0,1)$ and $r\in \Ic_0$. Then we have $r(2^{-s}-r)\leq K_p2^{(-2+p)s} r^p$ where 
\begin{align}\label{K_p}
K_p=\Big(\frac{1/p}{2/p-1}\Big)\Big(\frac{1/p-1}{2/p-1}\Big)^{(1-p)}. 
\end{align}
\elm 
\bpr
We can find $K_p$ through solving $K_p=2^{(2-p)s}\max_{r\in \Ic_0}r(2^{-s}-r)/r^p$. Expressing the optimization variable as $r=2^{-s}\theta^{1/p}$, $K_p$ can be obtained by solving this problem: 

\begin{align}\label{opt_K_p}
K_p=\max_{0\leq\theta\leq 1} \theta^{1/p-1}-\theta^{2/p-1}. 
\end{align}
We can solve \eqref{opt_K_p} and obtain the optimal solution $\theta^*=\big(\frac{1/p-1}{2/p-1}\big)^p $. Substituting $\theta^*$ into \eqref{opt_K_p}, we obtain \eqref{K_p}.  
\epr

Note that using H\"{o}lder's inequality, we have 
\begin{align}\nn
\sum_{r_i\in\Ic_0}r_i^p= \sum_{i\in\Sc_0}\frac{|v_i|^p}{\|\vbf\|^p}\leq \Big(\frac{\|\vbf\|_p}{\|\vbf\|}\Big)^p\leq d^{1-p/2}. 
\end{align}
This gives us an upper bound on $\E[\|Q_s(\vbf)-\vbf\|^2]$:  
\begin{align}\nn
\E[\|Q_s(\vbf)-\vbf\|^2]\leq \|\vbf\|^2(1/8+K_p2^{(-2+p)s}d^{1-p/2}). 
\end{align}
  
Furthermore, note that $r(2^{-s}-r)\leq 2^{-s}r$ and $r(2^{-s}-r)\leq 2^{-s}(2^{-s}-r)$ for $r\in \Ic_0$. This leads to the following upper bound on $\E[\|Q_s(\vbf)-\vbf\|^2]$:  
\begin{align}\nn
\E[\|Q_s(\vbf)-\vbf\|^2]\leq \|\vbf\|^2(1/8+\min(2^{-s}\sqrt{d},2^{-2s}d)). 
\end{align}  

The final upper bound is obtained using the following lemma.

\blm\label{lm:Sine}
Let $r\in \Ic_0$. Then we have $r(2^{-s}-r)\leq \hat K \sin\big(2^s\pi r\big)$ where $\hat K=2^{-2s}/\pi$. 
\elm 
\bpr
We can find $\hat K$ through solving $\hat K=\sup_{r\in (0,2^{-s})}r(2^{-s}-r)/\sin(2^s\pi r)$. Expressing the optimization variable as $r=2^{-s}\theta$, $\hat K$ can be obtained by solving this problem: 

\begin{align}\label{opt_Sine}
\hat K=2^{-2s}\sup_{0<\theta< 1} \theta(1-\theta)/\sin(\pi\theta)=2^{-2s}/\pi. 
\end{align}
\epr

Finally an upper bound on $\E[\|Q_s(\vbf)-\vbf\|^2]$ is given by:

\begin{align}\nn
\E[\|Q_s(\vbf)-\vbf\|^2]&\leq \|\vbf\|^2\Big(1/8+2^{-2s}|\Sc_0|/\pi \sin\big(2^s \pi/|\Sc_0|\sum_{i\in\Sc_0}\|v_i\|/\|\vbf\|\big)\Big)\\
&\leq \|\vbf\|^2\big(1/8+2^{-2s}d/\pi\big)
\end{align} where the first inequality follows from Jensen's inequality.

Combining these bounds, we have
\begin{align}\label{varboundwrtp}
\E[\|Q_s(\vbf)-\vbf\|^2]\leq\epsilon_Q\|\vbf\|^2
\end{align} where $\epsilon_Q=1/8+\inf_{0<p<1}K_p2^{(-2+p)s}d^{1-p/2}$  with $K_p=\big(\frac{1/p}{2/p-1}\big)\big(\frac{1/p-1}{2/p-1}\big)^{(1-p)}$. 

Note that the optimal $p$ to minimize $\epsilon_Q$ is obtained by minimizing:  
\begin{align}\nn
\Xi(p) =  \big(\frac{1/p}{2/p-1}\big)\big(\frac{1/p}{2/p-1}\big)^{1-p}\delta^{1-p}
\end{align} where $\delta = \sqrt{d}/2^s$. 

Differentiating $\Xi(p)$, the optimal $p^*$ is given by  
\begin{align}\label{optpcaes}
p^*=
\begin{cases}
\frac{\delta-2}{\delta-1},\quad\delta\geq 2 \\
0,\quad \delta< 2. 
\end{cases}
\end{align}
Substituting~\eqref{optpcaes} into~\eqref{varboundwrtp} gives \eqref{varbound}.

For the second part of theorem, substituting $\tau_0=2^{-s}$ and $\tau_{j}=2^{j-1-s}$ for $j\in\{1,\cdots,s\}$ into \eqref{vareq}, we have

\begin{align}\label{varineq}
\E[\|Q_s(\vbf)-\vbf\|^2]&=\|\vbf\|^22^{-2s}\sum_{i\in\Sc_0}p_1(r_i,s)(1-p_1(r_i,s))\nn\\
&\quad+\|\vbf\|^2\sum_{j=0}^{s-1}2^{2(j-s)}\sum_{i\in\Sc_{j+1}}p_2(r_i,s)\big(1-p_2(r_i,s)\big)\nn\\
\begin{split}
&\leq \|\vbf\|^22^{-2s}\sum_{i\in\Sc_0}p_1(r_i,s)\\
&\quad+\|\vbf\|^2\sum_{j=0}^{s-1}2^{2(j-s)}\sum_{i\in\Sc_{j+1}}p_2(r_i,s)
\end{split}
\end{align}
We first note that $\sum_{i\in\Sc_0}p_1(r_i,s)\leq d$ and $\sum_{i\in\Sc_{j+1}}p_2(r_i,s)\leq d$ for all $j$, \ie an upper bound on the variance of $Q_s(\vbf)$ is given by 
$\E[\|Q_s(\vbf)-\vbf\|^2]\leq \|\vbf\|^2 d/3(2^{-2s+1}+1)$. Furthermore, we have 
\begin{align}\label{p1bound}
\sum_{i\in\Sc_0}p_1\big(r_i,s\big)\leq\min\{d_0,2^s\sqrt{d_0}\}
\end{align} since  $\frac{\sum_{i\in\Sc_0}|v_i|}{\|\vbf\|}\leq\sqrt{d_0}$.
Similarly, we have 
\begin{align}\label{p2bound}
\sum_{i\in\Sc_{j+1}}p_2\big(r_i,s\big)\leq\min\{d_{j+1},2^{(s-j)}\sqrt{d_{j+1}}\}. 
\end{align}

Considering the variance expression \eqref{var_reform}, note that $(2^{-s}-r)r_i\leq 2^{-2s}/4$ for $r\in \Ic_0$ and $(2^{j+1-s}-r)(r-2^{j-s})\leq 2^{2j-2s}/4$ for $r\in \Ic_{j+1}$ for all $j$. This gives us an upper bound: 
\begin{align}\label{3bound} 
\E[\|Q_s(\vbf)-\vbf\|^2]\leq\|\vbf\|^2/4\big(2^{-2s}d_0+\sum_{j=0}^{s-1}2^{2j-2s}d_{j+1}\big).
\end{align}

Substituting the upper bounds in \eqref{p1bound}, \eqref{p2bound}, and \eqref{3bound} into \eqref{varineq}, an upper bound on the variance of $Q_s(\vbf)$ is given by 
\begin{align}
\begin{split}\label{varbound_bad}
\E[\|Q_s(\vbf)-\vbf\|^2]&\leq\min\{2^{-2s}d_0/4,2^{-s}\sqrt{d_0}\}\|\vbf\|^2\\
&\quad+\sum_{j=0}^{s-1}\min\{2^{2(j-s)}d_{j+1}/4,2^{j-s}\sqrt{d_{j+1}}\}\|\vbf\|^2.
\end{split}  
\end{align} 
The upper bound in \eqref{varbound_bad} cannot be used directly as it depends on $\{d_0,\cdots,d_s\}$. Note that $d_j$'s  depend on quantization intervals. In the following, we obtain an upper bound on $\E[\|Q_s(\vbf)-\vbf\|^2]$, which depends only on $d$ and $s$. To do so, we need to use this lemma inspired by  \citep[Lemma A.5]{QSGD}: Let $\|\cdot\|_0$ count the number of nonzero components.

\blm\label{lm:sparsity}
Let $\vbf\in\reals^d$. The expected number of nonzeros in $Q_s(\vbf)$ is bounded above by
\begin{align}%
\E[\|Q_s(\vbf)\|_0]\leq 2^{2s}+\sqrt{d_0}2^s.\nn 
\end{align}
\elm 
\bpr
Note that $d-d_0\leq 2^{2s}$ since 
\begin{align}
(d-d_0)2^{-2s}\leq \sum_{i\not\in \Sc_0}r_i^2\leq 1.
\end{align}
For each $i\in\Sc_0$, $Q_s(v_i)$ becomes zero with probability $1-2^sr_i$, which results in 
\begin{align}
\E[\|Q_s(\vbf)\|_0]&\leq d-d_0+\sum_{i\in\Sc_0}r_i2^{s}\nn\\
&\leq 2^{2s}+\sqrt{d_0}2^s.
\end{align}
\epr

Using a similar argument as in the proof of Lemma \ref{lm:sparsity}, we have 
\begin{align}
d-d_0-d_1-\cdots-d_j\leq 2^{2(s-j)}
\end{align} for $j=0,1,\cdots,s-1$. Define $b_j\defeq d-2^{2(s-j)}$ for $j=0,\cdots,s-1$. Then 
\begin{align}
b_0&\leq d_0\nn\\
b_1&\leq d_1+d_0\nn\\
\vdots&\qquad \vdots\nn\\
b_{s-1}&\leq d_0+\cdots+d_{s-1}.
\end{align} Note that $d_s=d-d_0-\cdots-d_{s-1}$. 

We define 
\begin{align}
\tilde d_0&\defeq b_0=d-2^{2s}\nn\\
\tilde d_1&\defeq b_1-b_0=3\cdot 2^{2(s-1)}\nn\\
\vdots&\qquad \vdots\nn\\
\tilde d_{s-1}&\defeq b_{s-1}-b_{s-2}=12\nn\\
\tilde d_s&\defeq d-\tilde d_0-\tilde d_1-\cdots-\tilde d_{s-1}=4.
\end{align}

Note that $\tilde d_0\leq d_0$, $\tilde d_1+\tilde d_0\leq d_1+d_0$, $\cdots$, $\tilde d_{s-1}+\cdots+\tilde d_0\leq d_{s-1}+\cdots+d_0$, and $\tilde d_s+\cdots+\tilde d_0= d_s+\cdots+d_0$. 

Noting that the coefficients of the additive terms in the upper bound in \eqref{varbound_bad} are monotonically increasing with $j$, we can find an upper bound on $\E[\|Q_s(\vbf)-\vbf\|^2]$ by replacing $(d_0,\cdots,d_s)$ with $(\tilde d_0,\cdots,\tilde d_s)$ in \eqref{varbound_bad}, which completes the proof.%

\section{Proof of Theorem \ref{thm:codebound} (Code-length Bound)}\label{app:pr_code}

Let $|\cdot|$ denote the length of a binary string.
In this section, we find an upper bound on $\E[|\ENCODE(\vbf)]$, \ie the expected number of communication bits per iteration. 
Recall from Appendix~\ref{app:elias} that the 
quantized gradient $Q_s(\vbf)$ is determined by the tuple 
$(\|\vbf\|,\rhobf,\hbf)$.
Write $i_1 < i_2 < \dots < i_{\|\hbf\|_0}$ for the indices of the $\|\hbf\|_0$ nonzero entries of $\hbf$. Let $i_0 = 0$.

\newcommand{\EncodingRuns}{R}
\newcommand{\EncodingEntries}{E}
The encoding produced by $\ENCODE(\vbf)$ can be partitioned into two parts, $\EncodingRuns$ and $\EncodingEntries$, such that,
for $j =1,\dots,\|\hbf\|_0$,
\begin{itemize}
\item $\EncodingRuns$ contains the codewords $\Elias(i_j-i_{j-1})$ encoding the runs of zeros; and
\item $\EncodingEntries$ contains the sign bits and codewords $\Elias(\log \{2^{s+1}h_{i_j}\})$ encoding the normalized quantized coordinates.
\end{itemize}

Note that $\|[i_1,i_2-i_1,\cdots,i_{\|\hbf\|_0}-i_{\|\hbf\|_0-1}]\|_1\leq d$.
Thus, by \citep[][Lemma A.3]{QSGD}, the properties of Elias encoding imply that
\begin{align}\label{c1bound}
|\EncodingRuns| \leq \|\hbf\|_0+(1+o(1))\|\hbf\|_0\log\Big(\frac{d}{\|\hbf\|_0}\Big).
\end{align}

We now turn to bounding $|\EncodingEntries|$. The following result in inspired by \citep[][Lemma A.3]{QSGD}.

\blm\label{lm:Eliasbound} 
Fix a vector $\qbf$ such that $\|\qbf\|_p^p\leq P$,
let $i_1 < i_2 < \dots i_{\|\qbf\|_0}$ be the indices of its $\|\qbf\|_0$ nonzero entries,
and assume each nonzero entry is of form of $2^k$, for some positive integer $k$.
Then
\begin{align}
\sum_{j=1}^{\|\qbf\|_0}|\Elias(\log(q_{i_j}))|\leq& 
(1+o(1))\log\big(\frac{1}{p}\big)+\|\qbf\|_0\nn\\
&+(1+o(1))\|\qbf\|_0\log\log\Big(\frac{P}{\|\qbf\|_0}\Big).\nn
\end{align}  
\elm 
\bpr
Applying property (1) for ERC (end of Appendix \ref{app:elias}), we have 
\begin{align}
\sum_{j=1}^{\|\qbf\|_0}|\Elias(\log(q_{i_j}))|
&\leq(1+o(1))\sum_{j=1}^{\|\qbf\|_0}\log\log q_{i_j}+\|\qbf\|_0\nn\\
&\leq(1+o(1))\log\big(\frac{1}{p}\big)+\|\qbf\|_0\nn\\
& \qquad +(1+o(1))\sum_{j=1}^{\|\qbf\|_0}\log\log q_{i_j}^p\nn\\
&\leq (1+o(1))\log\big(\frac{1}{p}\big)+\|\qbf\|_0 \nn\\
&\qquad +(1+o(1))\|\qbf\|_0\log\log\big(\frac{P}{\|\qbf\|_0}\big)\nn %
\end{align} 
where the last bound is obtained by Jensen's inequality. 
\epr

Taking $\qbf = 2^{s+1} \hbf$, we note that $\| \qbf \|^2=2^{2s+2}\|\hbf\|^2$ and
\begin{align}\label{norm2ineq}
\|\hbf\|^2&\leq \sum_{i=1}^d\Big(\frac{v_i}{\|\vbf\|}+\frac{1}{2^s}\Big)^2\nn\\
&\leq 2\sum_{i=1}^d\Big(\frac{v_i^2}{\|\vbf\|^2}+\frac{1}{2^{2s}}\Big)=2\big(1+\frac{d}{2^{2s}}\big).
\end{align}
By Lemma \ref{lm:Eliasbound} applied to $\qbf$ and the upper bound \eqref{norm2ineq},
\begin{align} 
\begin{split}\label{c2bound}
|\EncodingEntries|&\leq -(1+o(1))+2\|\hbf\|_0 \\
&\qquad\qquad+(1+o(1))\|\hbf\|_0\log\log\Big(\frac{2^{2s+2}\|\hbf\|^2}{\|\hbf\|_0}\Big).
\end{split}
\end{align}

Combining \eqref{c1bound} and \eqref{c2bound}, 
we obtain an upper bound on the expected code-length:
\begin{align}
\label{codeboundpre}
\E[|\ENCODE(\vbf)|]\leq N(\|\hbf\|_0)
\end{align} 
where 
\begin{align}
\begin{split}
N(\|\hbf\|_0)
&=b+3\|\hbf\|_0+(1+o(1))
\E\big[\|\hbf\|_0\log\big(\frac{d}{\|\hbf\|_0}\big)\big]\\
&\qquad-(1+o(1))+(1+o(1))\E\big[\|\hbf\|_0\log\log\big(\frac{8(2^{2s}+d)}{\|\hbf\|_0}\big)\big].
\end{split}
\end{align}

It is not difficult to show that, for all $k>0$, $g_1(x)\defeq x\log\big(\frac{k}{x}\big)$ is concave. 
Note that $g_1$ is an increasing function up to $x=k/e$. 

Defining $g_2(x)\defeq x\log\log\big(\frac{C}{x}\big)$ and taking the second derivative, we have 
\begin{align}
g''_2(x)=-\big(x\ln(2)\ln(C/x)\big)^{-1}\Big(1+\big(\ln(C/x)\big)^{-1}\Big).
\end{align} 
Hence $g_2$ is also concave on $x< C$. Furthermore, $g_2$ is increasing up to some $C/5<x^*<C/4$. 
We note that $\E[\|\hbf\|_0]\leq2^{2s}+\sqrt{d}2^s$ following Lemma \ref{lm:sparsity}. 
By assumption $2^{2s}+\sqrt{d}2^s\leq d/e$, and so, Jensen's inequality and \eqref{codeboundpre} lead us to \eqref{codebound}.

\section{Proof of Theorem \ref{thm:totalbits} (Expected Number of Communication Bits)}\label{app:pr_totalbits}
Assuming $\frac{2\hat B}{K\epsilon^2} > \frac{\beta}{\epsilon}$,
then $T_{\epsilon}=O\big(\frac{2\hat B}{K\epsilon^2} R^2\big)$. 
Ignoring all but terms depending on $d$ and $s$, we have $T_{\epsilon} = O(\hat B/\epsilon^2)$. 
Following Theorems \ref{thm:varbound} and \ref{thm:codebound} for NUQSGD, $\zeta_{\NUQSGD,\epsilon}=O(N_Q\epsilon_Q B/\epsilon^2)$.
For QSGD, following the results of \citet{QSGD}, $\zeta_{\QSGD,\epsilon}=O(\tilde N_Q\tilde\epsilon_Q B/\epsilon^2)$ where $\tilde N_Q=3(s^2+s\sqrt{d})+(\frac{3}{2}+o(1))(s^2+s\sqrt{d})\log\Big(\frac{2(s^2+d)}{s^2+\sqrt{d}}\Big)+b$ and $\tilde\epsilon_Q=\min\big(\frac{d}{s^2},\frac{\sqrt{d}}{s}\big)$. 

In overparameterized networks, where $d\geq 2^{2s+1}$, we have $\epsilon_Q=2^{-s}\sqrt{d-2^{2s}}+O(s)$ and $\tilde\epsilon_Q=\sqrt{d}/s$.  
Furthermore, for sufficiently large $d$, $N_Q$ and $\tilde N_Q$ are given by $O\big(2^s\sqrt{d}\log\big(\frac{\sqrt{d}}{2^s}\big)\big)$ and $O\big(s\sqrt{d}\log(\sqrt{d})\big)$, respectively.

\section{Optimal Level for the Special Case with $s=1$}\label{app:onelevel}
\bcr[Optimal level]\label{cr:onelevel}
For the special case with $s=1$, the optimal level to minimize the worst-case bound obtained from problem $\Pc_1$ is given by $l_1^*=1/2$.   
\ecr
\bpr 
For $s=1$, problem $\Pc_1$ is given by 
\begin{align}
\Pc_0:\max_{(d_0,d_1)}&~(\tau_0^2d_0+\tau_1^2d_1)/4\nn\\
\st&d-d_0\leq (1/l_1)^2,\nn\\
&d_0+d_1\leq d,\nn\\
&d_0\geq 0,~d_1\geq 0.\nn
\end{align}
Note that the objective of $\Pc_0$ is monotonically increasing in $(d_0,d_1)$. It is not difficult to verify that the optimal $(d^*_0,d^*_1)$ is a corner point on the boundary line of the feasibility region of $\Pc_0$. Geometrical representation shows that that candidates for an optimal solution are $(d-(1/l_1)^2,(1/l_1)^2)$ and $(d,0)$. Substituting into the objective of $\Pc_0$, the optimal value of $\Pc_0$ is given by 
\begin{align}\label{onelevelopt}
\epsilon_{LP}^*= \max\{\tau_0^2d,\tau_0^2d+\tau_1^2/\tau_0^2-1\}/4. 
\end{align} Finally, note that $\tau_0=\tau_1=1/2$ minimizes the optimal value of $\Pc_0$ \eqref{onelevelopt}. 
\epr

\section{Lower Bound in the Regime of Large $s$}\label{app:lowerbound_s}
In the following theorem, we show that for exponentially spaced levels with $p=0.5$, if $s$ is large enough, there exists a distribution of points such that the variance is in $1/8\|\vbf\|^2$. 

\bth[Lower bound for large $s$]\label{thm:lowerbound_s}
Let $d\in\integers^{>0}$ and let $(0,2^{-s},\cdots,1/2,1)$ denote the sequence of quantization levels. Provided $s\geq \log(\sqrt{d})$, there exists a vector $\vbf\in\reals^d$ such that the variance of unbiased quantization of $\vbf$ is given by $\|\vbf\|^2/8$. 
\eth
\bpr
Consider $\vbf_0=[r,r,\cdots,r]^T$ for $r\neq 0$. The normalized coordinates is $\hat\vbf_0=[1/\sqrt{d},\cdots,1/\sqrt{d}]^T$. Suppose $s$ is large enough such that $2^{j-s}=\frac{3}{4\sqrt{d}}$ for some $j=0,\cdots,s-1$. We can compute of the variance and obtain $\E[\|Q_s(\vbf)-\vbf\|^2]=\|\vbf\|^2/8$. 
\epr

\section{Asynchronous Variant of NUQSGD}\label{app:asynch}
Asynchronous parameter-server model is an important setting that provides additional flexibility for distributed training of large models. 
In this section, we extend our consideration to asynchronous variant of NUQSGD when a star-shaped parameter-server setting is used for training. The star machine is a master processor, while other processors serve as workers. The master aggregates stochastic gradients received from workers and updates the model parameters. We consider a consistent scenario where workers cannot read the values of model parameters during the update step. This is a valid model when computer clusters are used for distributed training \citep{Asynch}. 

On smooth and possibly nonconvex problems in a consistent parameter-server setting, we establish convergence guarantees for asynchronous variant of NUQSGD  along the lines of, \eg \citep[Theorem 1]{Asynch}:

\bth[NUQSGD for asynchronous smooth nonconvex optimization]\label{thm:asynchronous}
Let $f:\Omega\rightarrow \reals$ denote a possibly nonconvex and $\beta$-smooth function. Let $\wbf_0\in\Omega$ denote an initial point, $\epsilon_Q$ be defined as in Theorem~\ref{thm:varbound} and $f^*=\inf_{\wbf\in\Omega}f(\wbf)$. Suppose $K$ workers each compute an unbiased stochastic gradient with mini-batch size $J$ and second-moment bound $B$, compress the stochastic gradient using nonuniform quantization, and send to a master processor for $T$ iterations. Suppose the delay for model parameters used in evaluation of a stochastic gradient at each iteration is upper bounded by $\tau$.  Provided that the sequence of learning rates satisfies $\beta J\alpha_t+2\beta^2J^2\tau\alpha_t\sum_{i=1}^{\tau}\alpha_{t+i}\leq 1$, for all $t=1,2,\cdots$, we have  

\begin{align}\nn
\sum_{t=1}^T\frac{\alpha_t\E[\|\nabla f(\wbf_t)\|^2]}{\sum_{t=1}^T\alpha_t}\leq \frac{2(f(\wbf_0)-f^*)+\lambda (1+\epsilon_Q)B/K}{J\sum_{t=1}^T\alpha_t}
\end{align} where $\lambda=\sum_{t=1}^T\big(\alpha_t^2J\beta+2\beta^2J2\alpha_t\sum_{i=t-\tau}^{t-1}\alpha_i^2\big)$.

\eth

\section{NUQSGD for Decentralized Training}\label{app:decentralized}
In a networked distributed system, decentralized algorithms are deployed when either processors cannot establish a fully connected topology due to communication barriers or the network suffers from high latency. When the network is bandwidth-limited as well, communication-efficient variants of decentralized parallel SGD is a promising solution to tackle both high latency and limited bandwidth to train deep models in a networked system \citep{Decentralized}. In this section, we show that NUQSGD can be integrated with extrapolation compression decentralized parallel SGD (ECD-PSGD) proposed in  \citep{Decentralized}.\footnote{Similarly, NUQSGD can be integrated with difference compressions D-PSD (DCD-PSD). However, DCD-PSD requires stringent constraints on stochastic compression scheme, which may fail under an aggressive compression. Hence, we focus on  ECD-PSGD in this paper.} In decentralized optimization, we consider the following problem
\begin{align}
\min_{\wbf\in\reals^d}F(\wbf)=\frac{1}{K}\sum_{i=1}^Kf_i(\wbf) \nn
\end{align} where $f_i(\wbf)=\E_{\xi\sim\Dc_i}[f(\wbf;\xi)]$, $\Dc_i$ is the local data distribution stored in processor $i$, and $f(\wbf;\xi)$ is the loss of a model described by $\wbf$ on example (mini-batch) $\xi$.  

At iteration $t$ of D-PSGD, each computing processor computes its local stochastic gradient, \eg processor $i$ computes $g_i(\wbf_t^{(i)})$ with $\E_{\xi_t^{(i)}\sim \Dc_i}[g_i(\wbf_t^{(i)})]=\nabla f_i(\wbf_t^{(i)})$ where $\wbf_t^{(i)}$ and $\xi_t^{(i)}$ are the local parameter vector and samples on processor $i$, respectively. Processor $i$ then fetches its neighbours' parameter vectors and updates its local parameter vector using $\wbf_{t+1/2}^{(i)}=\sum_{j=1}^KW_{i,j}\wbf_t^{(j)}$ where $W\in\reals^{K\times K}$ is a symmetric doubly stochastic matrix, \ie $W=W^T$ and $\sum_{j=1}^K W_{i,j}=1$ for all $i$. Note that $W_{i,j}\geq 0$ in general and $W_{i,j}=0$ means that processors $i$ and $j$ are not connected. Finally, processor $i$ updates its local parameter vector using the update rule $\wbf_{t+1}^{(i)}\leftarrow \wbf_{t+1/2}^{(i)}-\alpha g_i(\wbf_t^{(i)})$. ECD-PSGD integrated with NUQSGD is described in Algorithm \ref{ECDalg}.

\begin{algorithm}[t]
\small
\SetAlgoLined
\KwIn{local data, weight matrix $W$, initial parameter vector $\wbf_1^{(i)}=\wbf_1$, initial estimate $\tilde\wbf_1^{(i)}=\wbf_1$, learning rate $\alpha$, and $K$}
\For{$t=1$ {\bfseries to} $T$}{
	\For(\tcp*[h]{each transmitter processor (in parallel)}){$i=1$ {\bfseries to} $K$}{
		Compute $g_i(\wbf_t^{(i)})$ \tcp*[l]{local stochastic gradient}  
		Compute $\wbf_{t+1/2}^{(i)}=\sum_{j=1}^KW_{i,j}\tilde\wbf_t^{(j)}$ \tcp*[l]{neighbourhood weighted average} 
		Update $\wbf_{t+1}^{(i)}\leftarrow \wbf_{t+1/2}^{(i)}-\alpha g_i(\wbf_t^{(i)})$\;
		Compute $\zbf_{t}^{(i)}=(1-t/2)\wbf_{t}^{(i)}+t/2\wbf_{t+1}^{(i)}$\;
		Encode $c_{t}^{(i)}\leftarrow \ENCODE\big(\zbf_{t}^{(i)}\big)$\;
		Broadcast $c_{t}^{(i)}$ to connected neighbours\;
		Receive $c_{t}^{(j)}$ from connected neighbours\;
		$\tilde\wbf_{t+1}^{(j)}\leftarrow (1-2/t)\tilde\wbf_t^{(j)}+2/t\text{DECODE}\big(c_{t}^{(j)}\big)$ \tcp*[l]{update estimates for connected neighbours}
	}
}
Aggregate $\overline\wbf_{T}\leftarrow 1/K\sum_{i=1}^K\wbf_{T}^{(i)}$\;
\caption{ECD-PSGD with NUQSGD.}
\label{ECDalg}
\end{algorithm}
In this section, we make the following assumptions: 
\begin{itemize}
\item Given $W$, $\rho<1$ where $\rho$ denotes the second largest eigenvalue of $W$. 
\item  The variance of local stochastic gradients are bounded. In particular, there are $\sigma^2,\zeta^2$ such that for all $\wbf$ and $i$: 
\begin{align}
\E[\|g_i(\wbf)-\nabla f_i(\wbf)\|^2]&\leq \sigma^2,\nn\\
\frac{1}{K}\sum_{i=1}^K\|\nabla f_i(\wbf)-\nabla F(\wbf)\|^2&\leq \zeta^2.\nn
\end{align}
\end{itemize}   

On smooth and possibly nonconvex problems and under assumptions above, we establish convergence guarantees for ECD-PSGD integrated with NUQSGD along the lines of, \eg \citep[Corollary 4]{Decentralized}:
\bth[NUQSGD for decentralized smooth nonconvex optimization]\label{thm:decentralized}
Let $f_i$ denote a possibly nonconvex and $\beta$-smooth function for all $i$. Let $\epsilon_Q$ be defined as in Theorem~\ref{thm:varbound}. Suppose that Algorithm \ref{ECDalg} is executed for $T$ iterations with a learning rate $\alpha<(12\beta/(1-\rho)+\sigma\sqrt{T/K}+\zeta^{2/3}T^{1/3})^{-1}$ on $K$ processors,  each with access to independent and local stochastic gradients with variance bound $\sigma^2$.  Then we have  

\begin{align}
\frac{1}{T}\sum_{t=1}^T\E\Big[\Big\|\sum_{i=1}^K\frac{\nabla f_i(\sum_{i=1}^K \wbf_t^{(i)}/K)}{K}\Big\|^2\Big]&\lesssim \frac{\sigma(1+(1+\epsilon_Q)R^2\log(T)/K)}{\sqrt{KT}}+\frac{(1+\epsilon_Q)R^2\log(T)}{T}\nn\\&\quad+\zeta^{2/3}(1+(1+\epsilon_Q)R^2\log(T)/K)/T^{2/3}+1/T\nn
\end{align} where $R^2=\max_{1\leq i\leq K}\max_{1\leq t\leq T}\|\zbf_{t}^{(i)}\|^2$.
\eth

\section{NUQSGD vs QSGDinf}\label{app:nuq_vs_qinf}
In this section, we show that there exist vectors for which the variance of quantization under NUQSGD is guaranteed to be smaller than that under QSGDinf. Intuitively, with the same communication budget, NUQSGD is guaranteed to outperform QSGDinf in terms of variance for dense  vectors with a unique dominant coordinate. 
\bth[NUQSGD vs QSGDinf]\label{thm:nuq_vs_qinf}

Let $\vbf\in\reals^d$ be such that $v_i=1$ and $v_j=\Theta(1/d)$ for $j\neq i$. Provided that $d$  and $s$ are large enough to ensure $\frac{K_1}{(d-1)\sqrt{1+K_2^2/(d-1)}}<2^{-s}$  and $(1+K_1^2/(d-1))K_1(0.25K_1/(d-1)+2^{-s})<K_2(1/s-K_1/(d-1))$ for some $K_2<K_1=o(d)$, NUQSGD is guaranteed to have smaller variance than QSGDinf. 
\eth
\bpr 
Our proof argument is based on establishing a lower bound on the variance of QSGDinf and an upper bound on the variance of NUQSGD. Denote the variance quantization of $\vbf$ using QSGDinf and NUQSGD by $\Delta_{\QINF}\defeq\E[\|Q^{\QINF}_s(\vbf)-\vbf\|^2]$ and $\Delta_{\NUQ}\defeq\E[\|Q^{\NUQ}_s(\vbf)-\vbf\|^2]$, respectively. Note that $\|\vbf\|_{\infty}=1$. A lower bound on $\Delta_{\QINF}$ is given by 
\begin{align}
\Delta_{\QINF}\geq K_2(1/s-K_1/(d-1)).
\end{align}

We can bound the Eulidean norm of $\vbf$ as: 
\begin{align}\nn
\sqrt{1+K_2^2/(d-1)}\leq\|\vbf\|\leq\sqrt{1+K_1^2/(d-1)}.
\end{align}
Note that the assumption $\frac{K_1}{(d-1)\sqrt{1+K_2^2/(d-1)}}<2^{-s}$ implies that the normalized coordinates $v_j/\|\vbf\|\in[0,2^{-s}]$ for $j\neq i$. Then an upper bound on $\Delta_{\NUQ}$ is given by  
\begin{align}
\begin{split}
\Delta_{\NUQ}&\leq \Big(1+\frac{K_1^2}{d-1}\Big)\Big(0.5\Big(1-\frac{1}{\sqrt{1+K_1^2/(d-1)}}\Big)+\frac{K_12^{-s}}{\sqrt{1+K_2^2/(d-1)}}\Big)\\
&\leq \Big(1+\frac{K_1^2}{d-1}\Big)K_1(0.25K_1/(d-1)+2^{-s})
\end{split} 
\end{align} where we used $\sqrt{1+K_1^2/(d-1)}\geq 1+0.5K_1^2/(d-1)$ in the second inequality. 
\epr

\section{Additional Experiments}\label{app:exp}
In this section, we present further experimental results in a similar setting 
to Section~\ref{sec:exp}.

We first measure the variance and normalized variance at fixed snapshots during 
training by evaluating multiple gradient estimates using each quantization 
method (discussed in Section~\ref{sec:exp}).  All methods are evaluated on the same trajectory traversed by the 
single-GPU SGD\@. These plots answer this specific question: What would the 
variance of the first gradient estimate be if one were to train using SGD for 
any number of iterations then continue the optimization using another method?  
The entire future trajectory may change by taking a single good or bad step. We 
can study the variance along any trajectory. However, the trajectory of SGD is 
particularly interesting because it covers a subset of points in the parameter 
space that is likely to be traversed by any first-order optimizer.  For 
multi-dimensional parameter space, we average the variance of each dimension.

\begin{figure*}[t]
    \includegraphics[width=.49\textwidth]{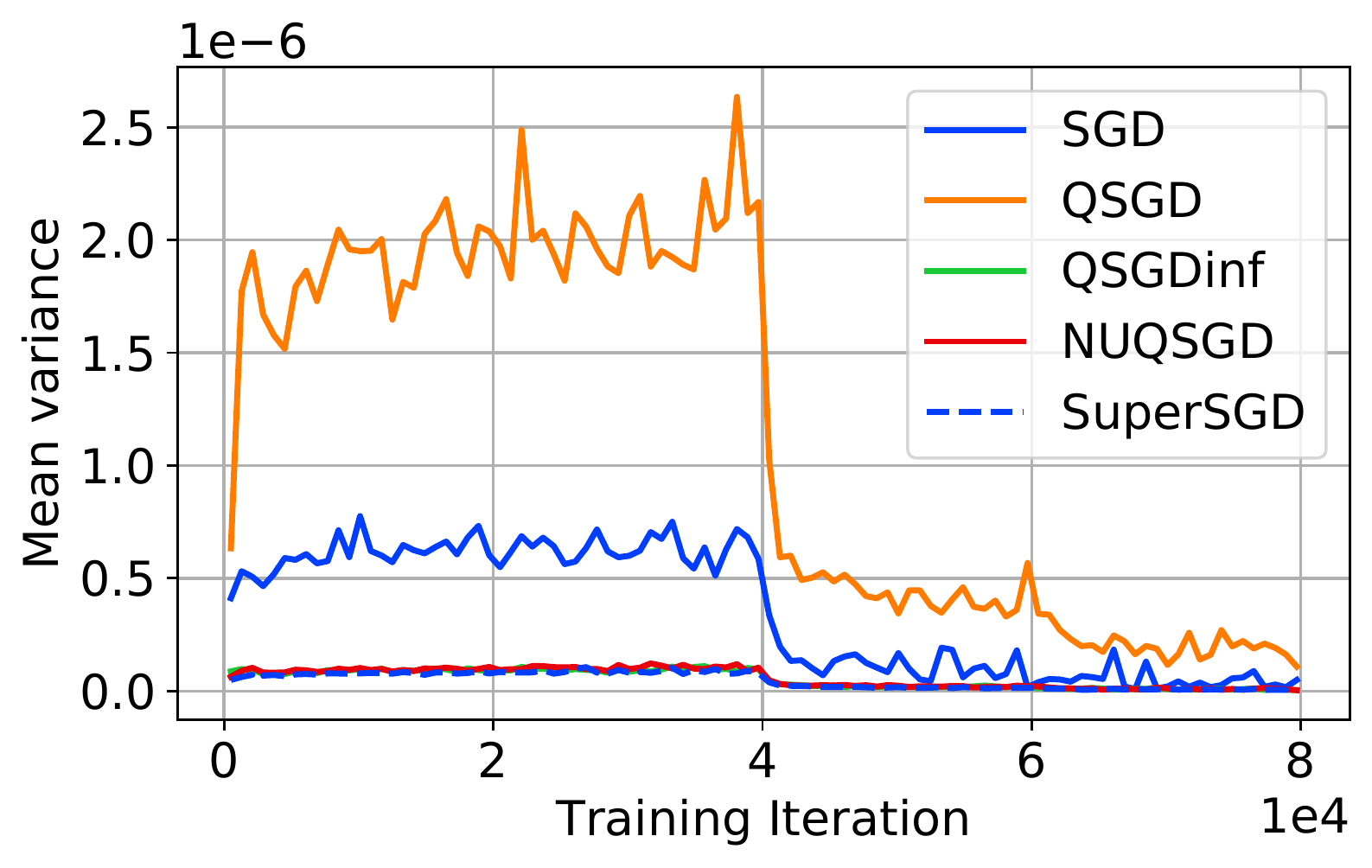}
    \hfill
    \includegraphics[width=.49\textwidth]{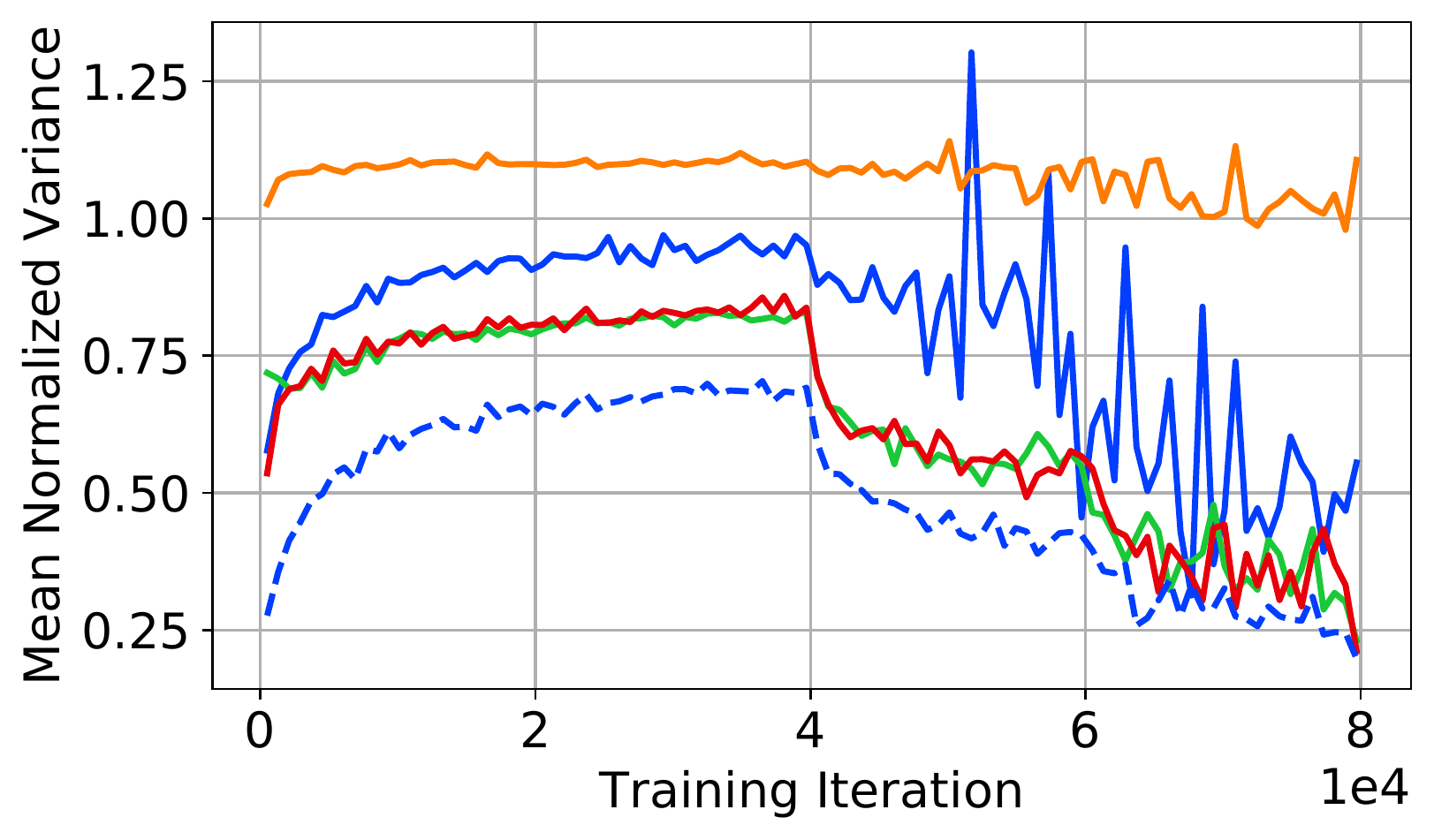}
    \caption{Estimated variance (left) and normalized variance (right) on 
    CIFAR10 on the trajectory of single-GPU SGD\@. Variance is measured for 
    fixed model snapshots during training.  Notice that the variance for NUQSGD 
    and QSGDinf is lower than SGD for almost all the training and it decreases 
    after the learning rate drops.  All methods except SGD simulate training 
    using $8$ GPUs.  SuperSGD applies no quantization to the gradients and represents the lowest variance we could hope to achieve.}
    \label{fig:var}
\end{figure*}

Figure~\ref{fig:var} (left), shows the variance of the gradient estimates on 
the trajectory of single-GPU SGD on CIFAR10.  We observe that QSGD has 
particularly high variance, while QSGDinf and NUQSGD have lower variance 
than single-GPU SGD.  

We also propose another measure of stochasticity, normalized variance, that is 
the variance normalized by the norm of the gradient. The mean normalized 
variance can be expressed as
\[
\frac{\E_i[V_A[ \partial l(\wbf;\zbf)/ \partial w_i]]}
{\E_i[\E_A[ (\partial l(\wbf;\zbf)/ \partial w_i)^2]]}
\]
where $l(\wbf;\zbf)$ denotes the loss of the model parametrized by $\wbf$ on  
sample $\zbf$ and subscript $A$ refers to randomness in the algorithm, \eg 
randomness in sampling and quantization. Normalized variance can be interpreted 
as the inverse of Signal to Noise Ratio (SNR) for each dimension. We argue that 
the noise in optimization is more troubling when it is significantly larger 
than the gradient. For sources of noise such as quantization that stay constant 
during training, their negative impact might only be observed when the norm of 
the gradient becomes small.

Figure~\ref{fig:var} (right) shows the mean normalized variance of the gradient 
versus training iteration. Observe that the normalized variance for QSGD stays 
relatively constant while the unnormalized variance of QSGD drops after 
the learning rate drops. It shows that the quantization noise of QSGD can cause 
slower convergence at the end of the training than at the beginning.

\begin{figure*}[t]
    \includegraphics[width=.49\textwidth]{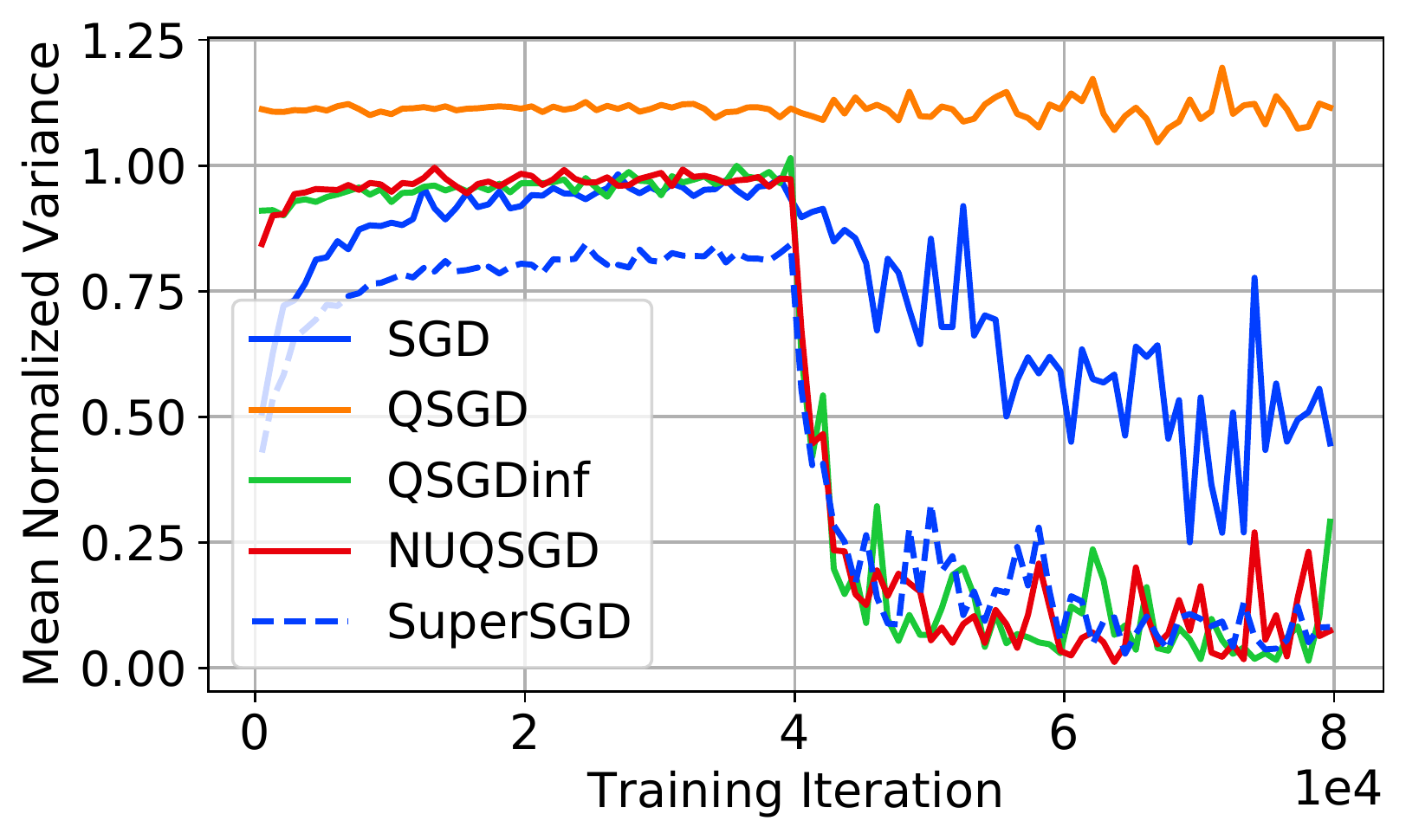}
    \hfill
    \includegraphics[width=.49\textwidth]{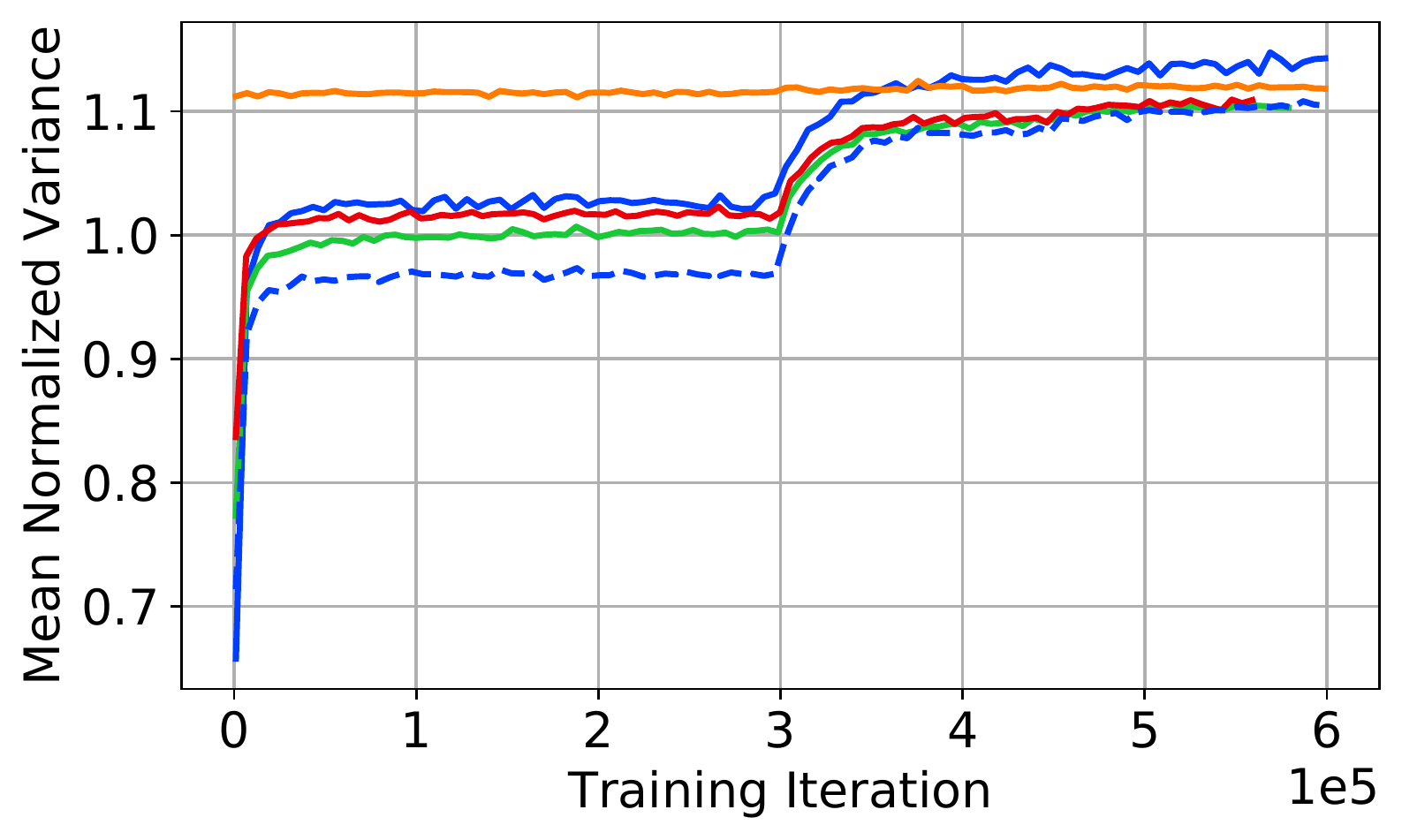}
    \caption{Estimated normalized variance on CIFAR10 (left) and ImageNet 
    (right). For different methods, the variance is measured on their own trajectories. 
    Note that the normalized variance of NUQSGD and QSGDinf is lower than SGD for 
    almost the entire training. It decreases on CIFAR10 after the learning rate 
    drops and does not grow as much as SGD on ImageNet. Since the variance 
    depends on the optimization trajectory, these curves are not directly 
    comparable. Rather the general trend should be studied.}
    \label{fig:var_n}
\end{figure*}

In Figure~\ref{fig:var_n}, we show the mean normalized variance of the gradient versus training iteration on CIFAR10 and ImageNet. 
For different methods, the variance is measured on their own trajectories. Since the variance depends on the optimization trajectory, these curves are not directly comparable. Rather the general trend should be studied.

\begin{figure*}[t]
    \includegraphics[width=.32\textwidth]{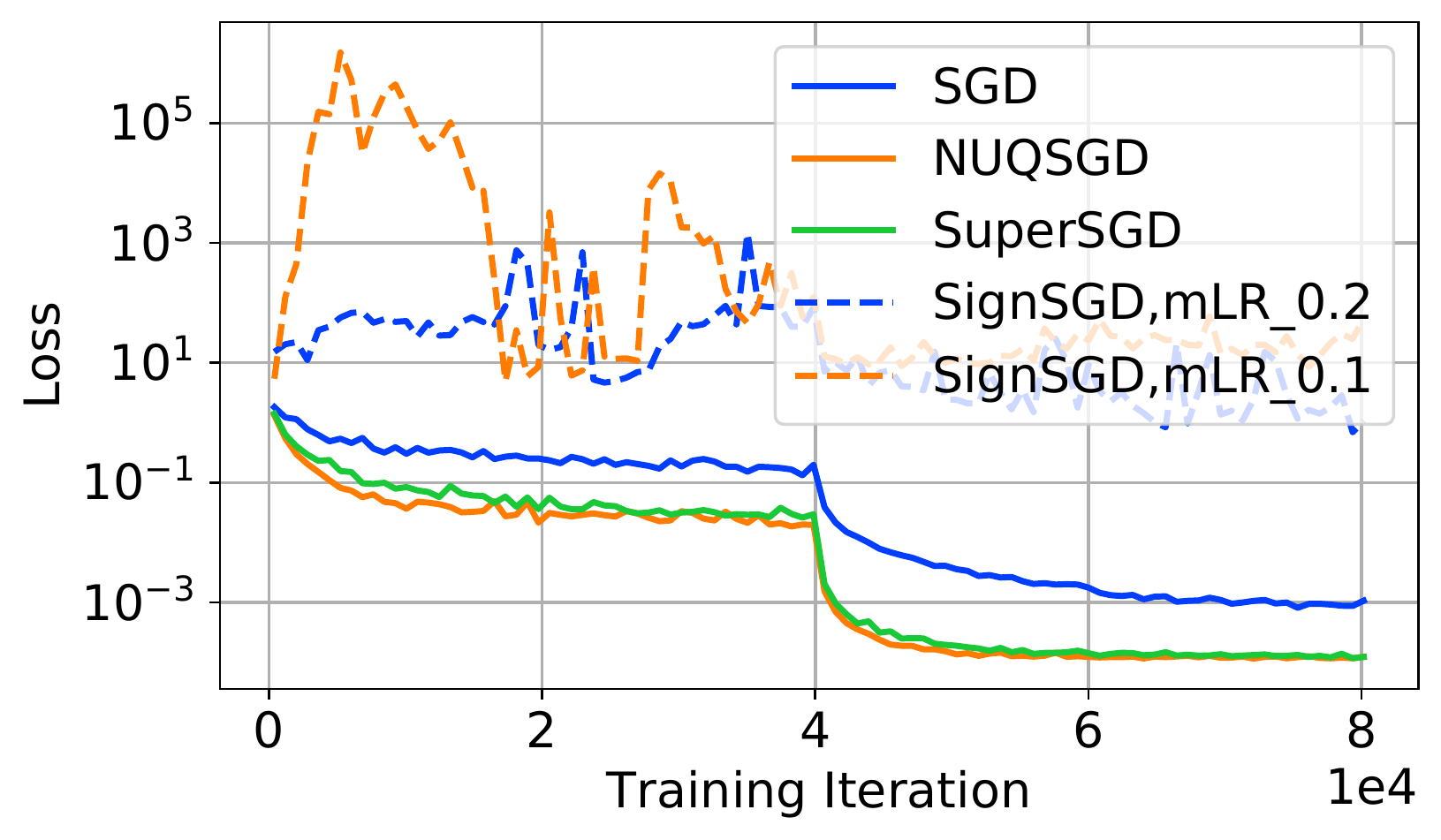}
    \hfill
    \includegraphics[width=.32\textwidth]{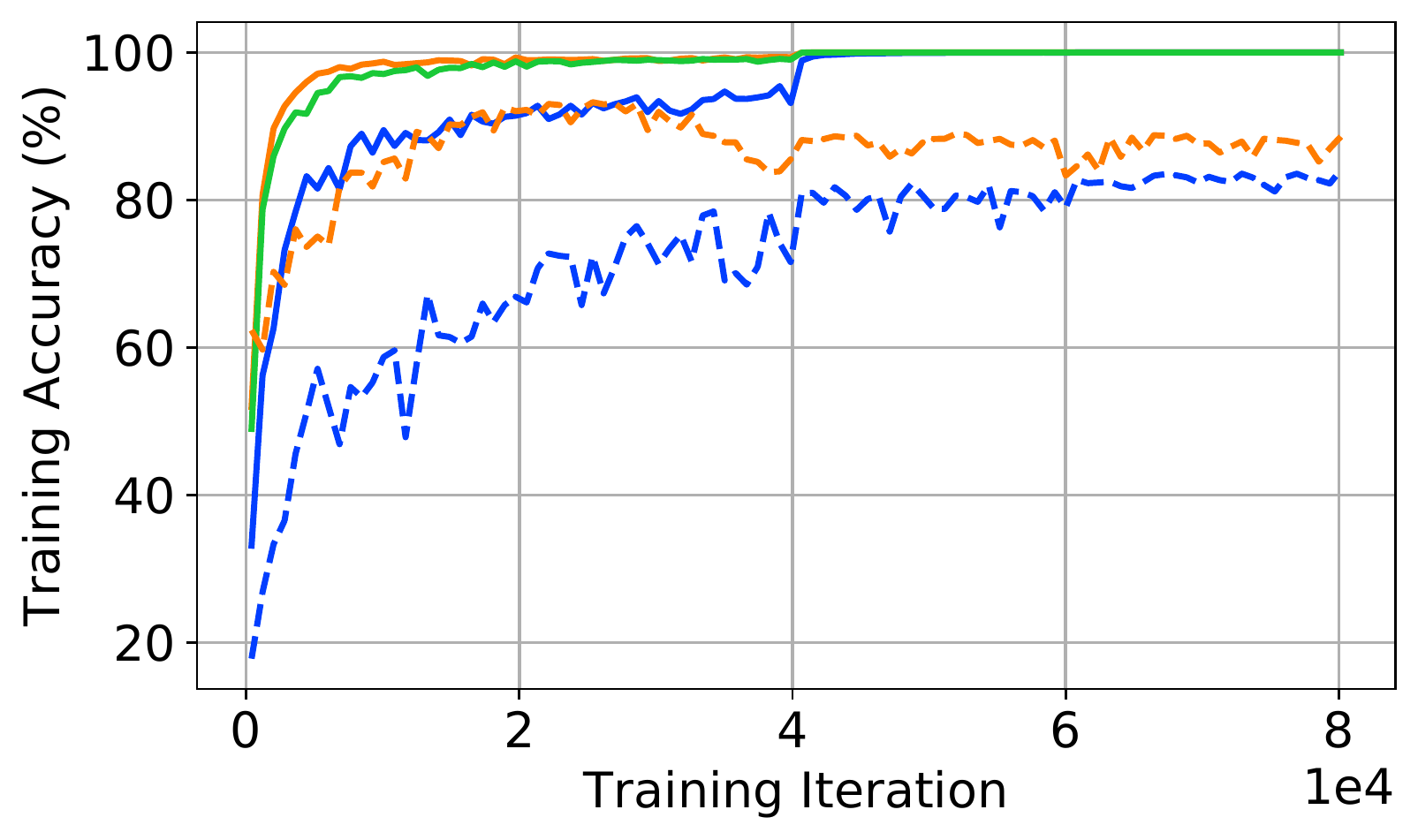}
    \hfill
    \includegraphics[width=.32\textwidth]{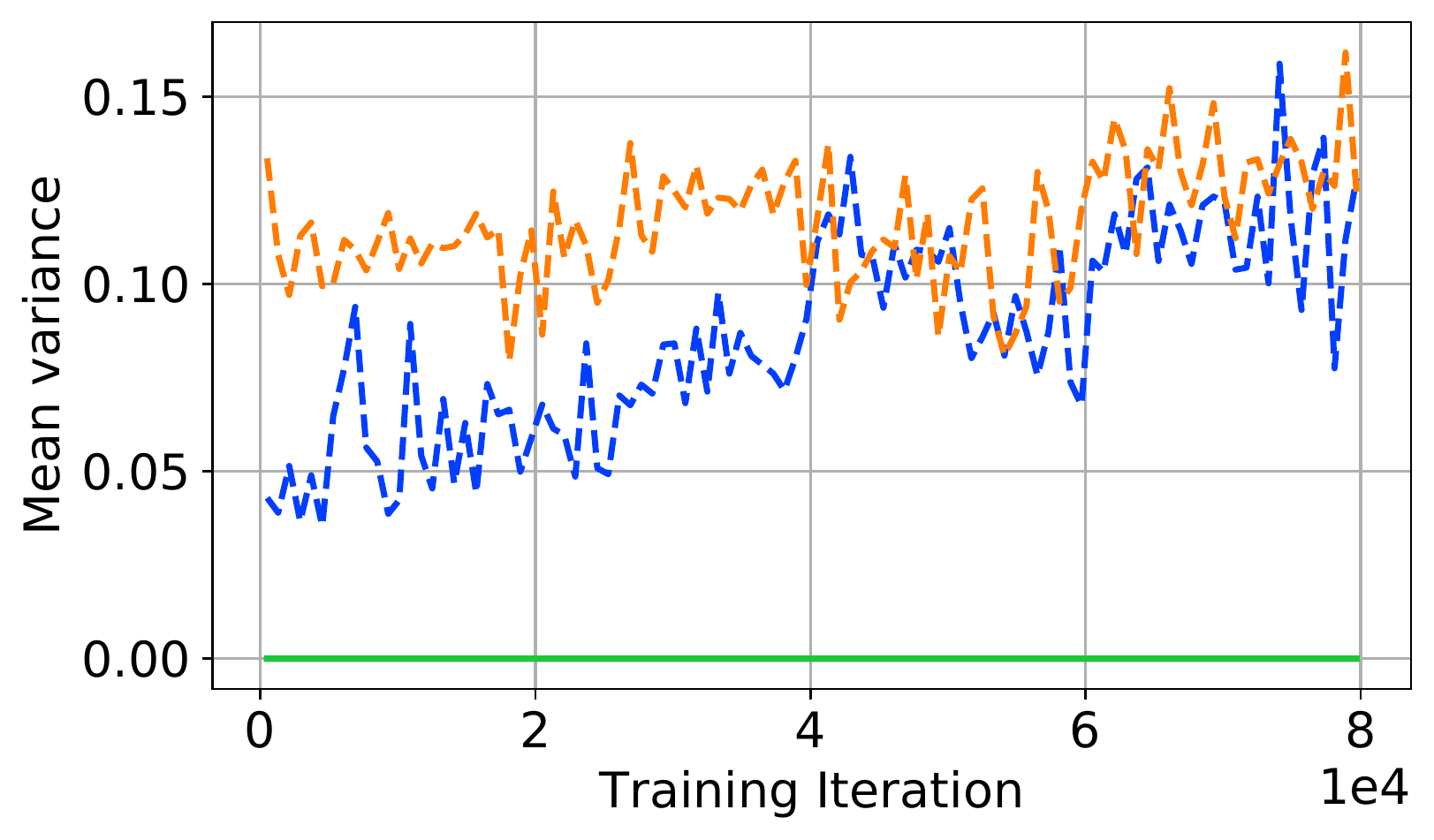}
    \caption{Comparison with SignSGD on CIFAR10. Training loss (left), training 
    accuracy (middle), and estimated variance (right) for training on $8$-GPUs.  
    Setting is similar to Figure~\ref{fig:loss}. SignSGD diverges with the 
    standard learning rate for ResNets. We multiply the standard learning rate 
    schedule by the constant mLR.}
    \label{fig:signsgd}
\end{figure*}

\begin{figure*}[t]
    \includegraphics[width=.32\textwidth]{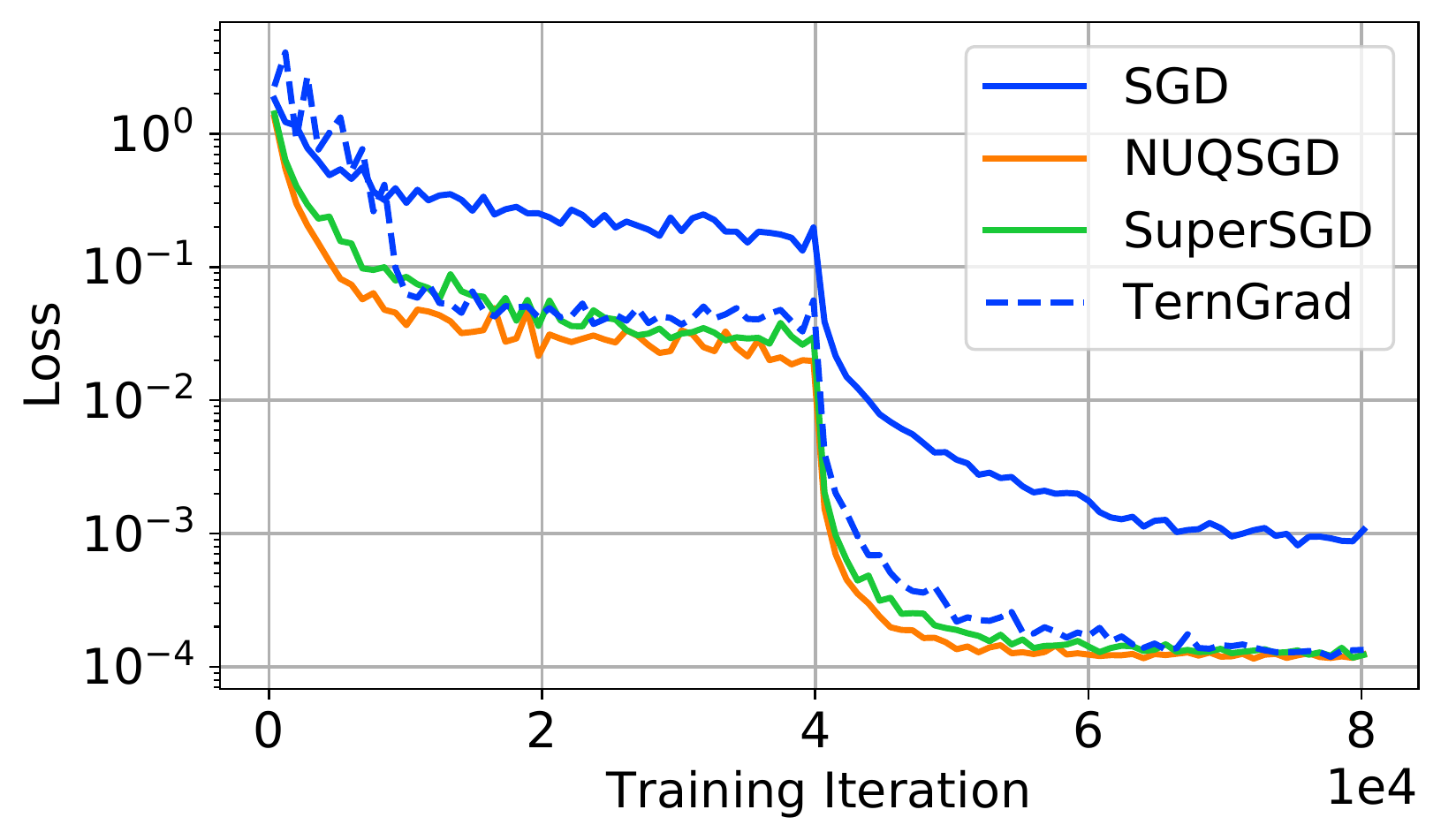}
    \hfill
    \includegraphics[width=.32\textwidth]{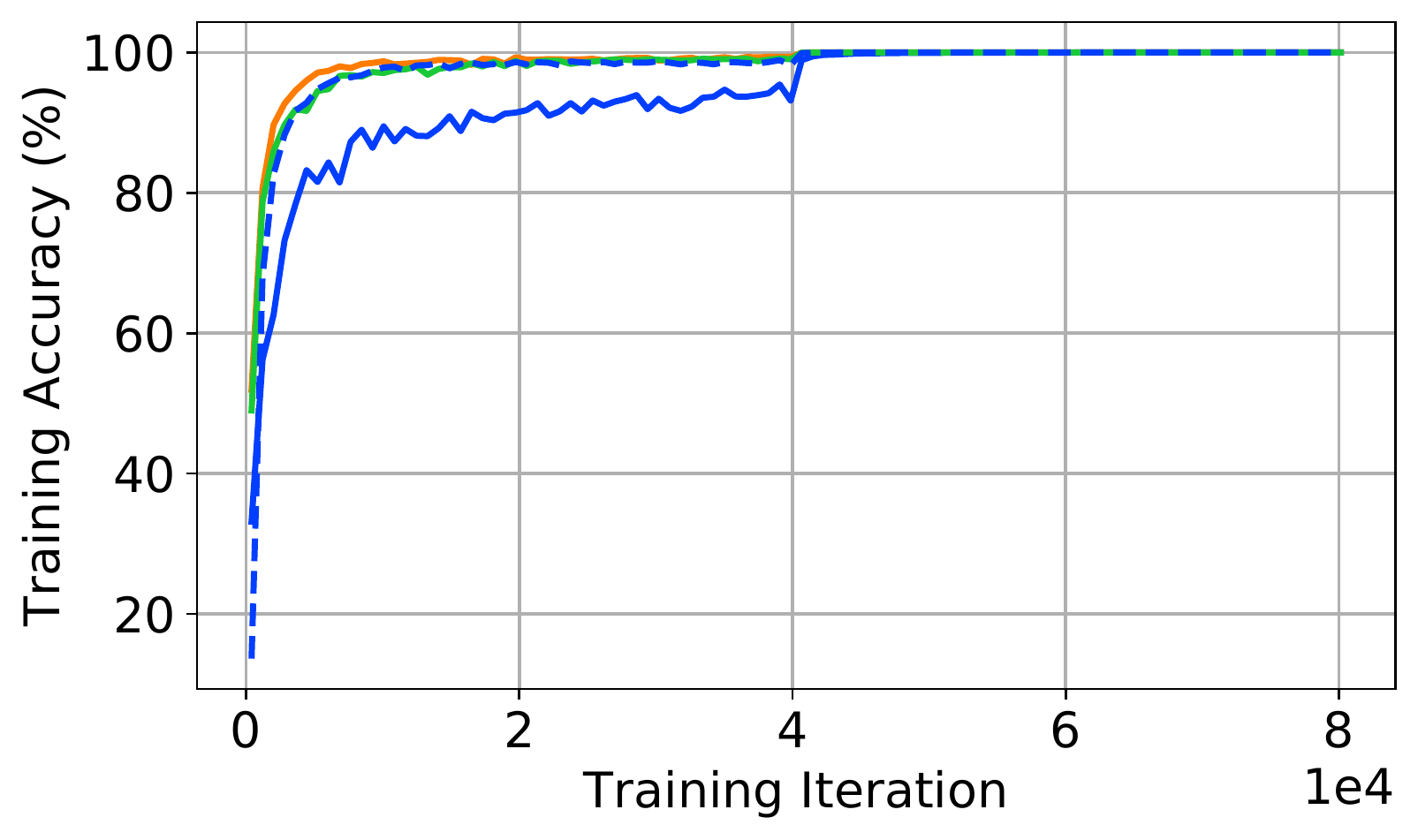}
    \hfill
    \includegraphics[width=.32\textwidth]{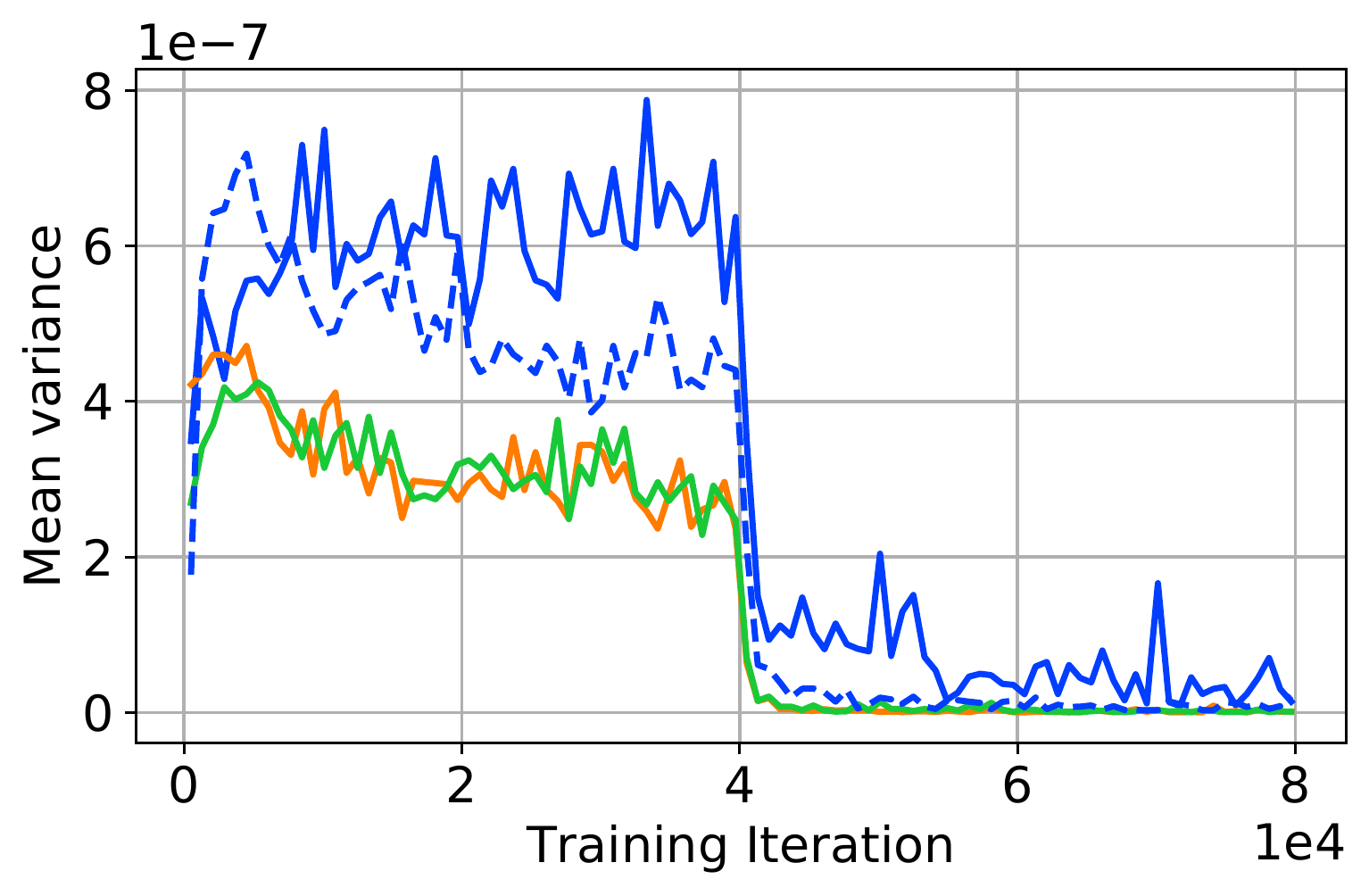}
    \caption{Comparison with TernGrad on CIFAR10. Training loss (left), 
    training accuracy (middle), and estimated variance (right) for training on 
    $8$-GPUs.  Setting is similar to Figure~\ref{fig:loss}. The performance 
    of TernGrad is inferior to NUQSGD.}
    \label{fig:signsgd}
\end{figure*}

\begin{figure*}[t]
    \includegraphics[width=.49\textwidth]{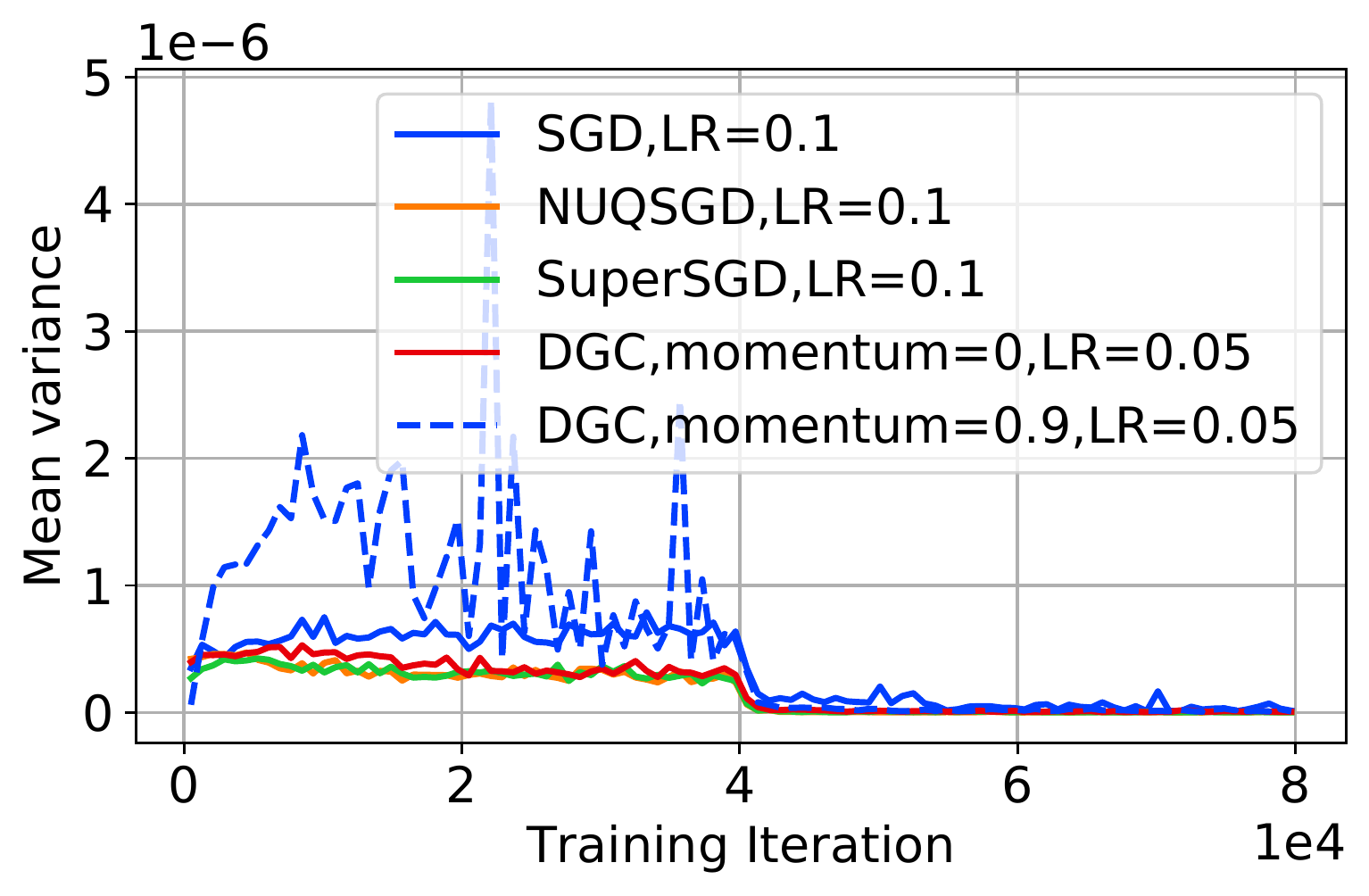}
    \hfill
    \includegraphics[width=.49\textwidth]{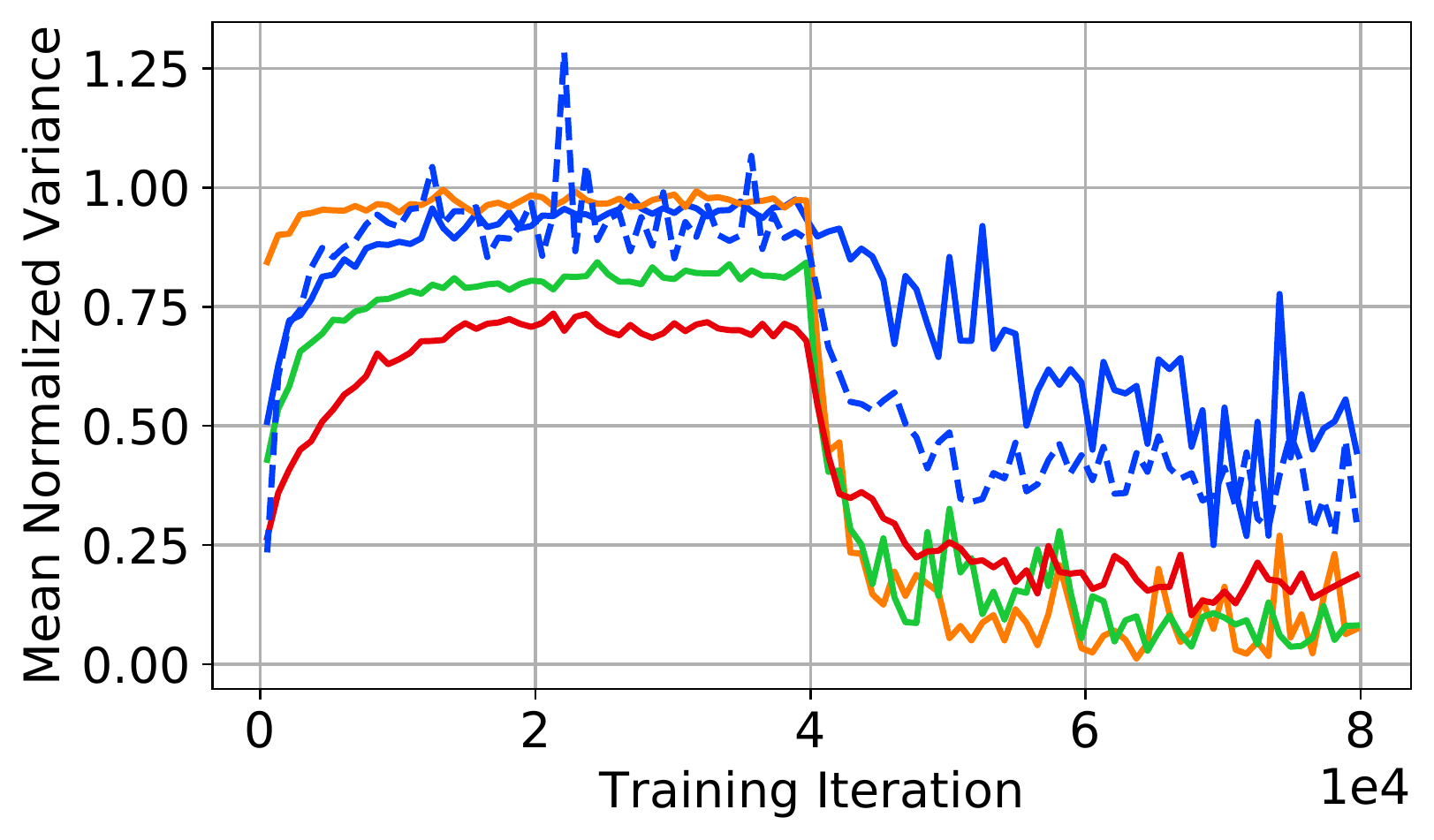}
    \caption{Estimated variance (left) and normalized variance (right) on CIFAR10 for ResNet110. 
    We set the ratio of compression for DGC to be roughly the same as NUQSGD.
    In particular, we compare compression methods at compression ratio = $4/32$, \ie NUQSGD with 
    4 bits and DGC at 12.5\% compression. At this rate, both methods will have approximately the same 
    communication cost, \ie comparison in simulation is representative of real-time performance. 
    We tune the learning rate and momentum for DGC and show its best performance.}
    \label{fig:DGCsimvar}
\end{figure*}

\begin{figure*}[t]
    \includegraphics[width=.49\textwidth]{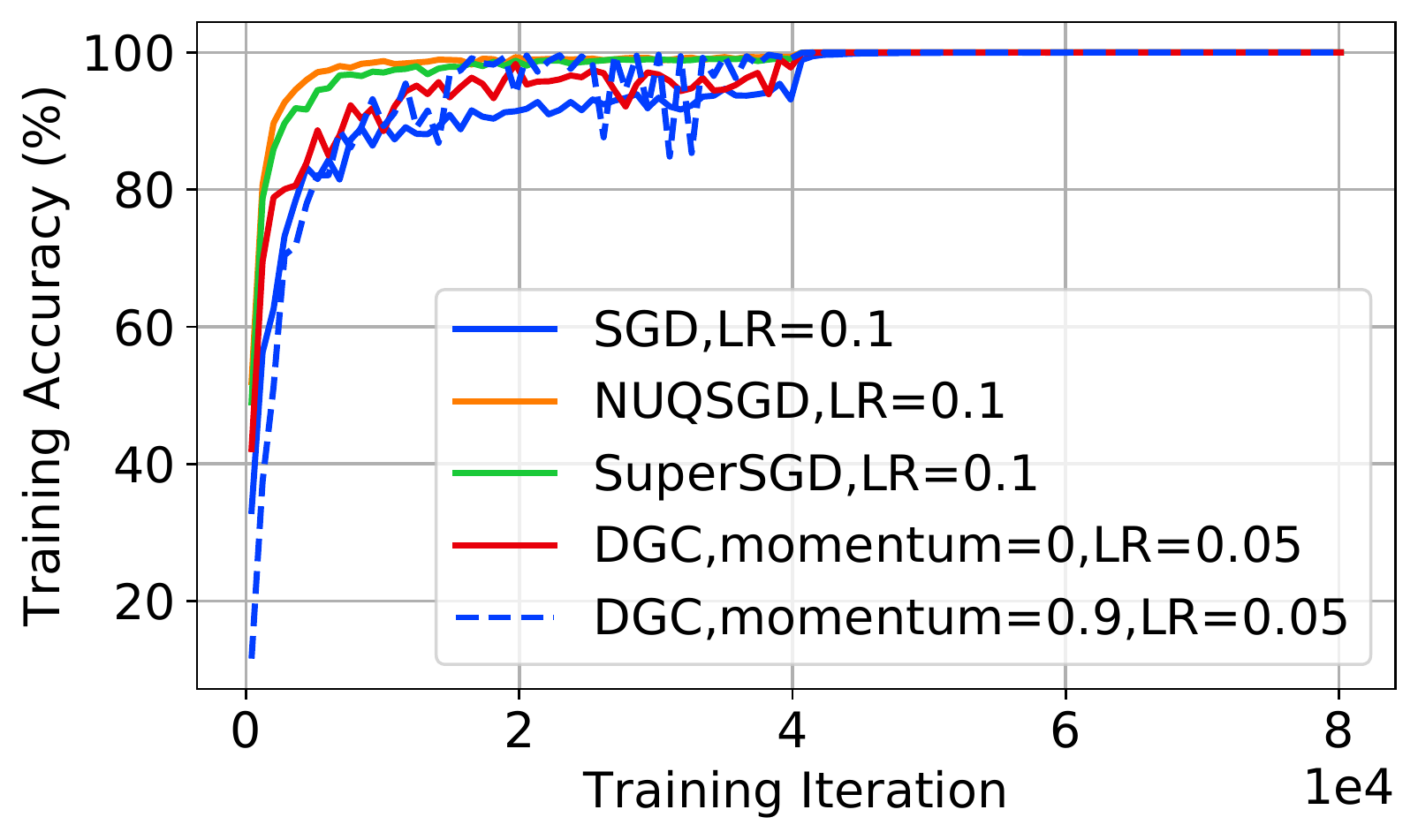}
    \hfill
    \includegraphics[width=.49\textwidth]{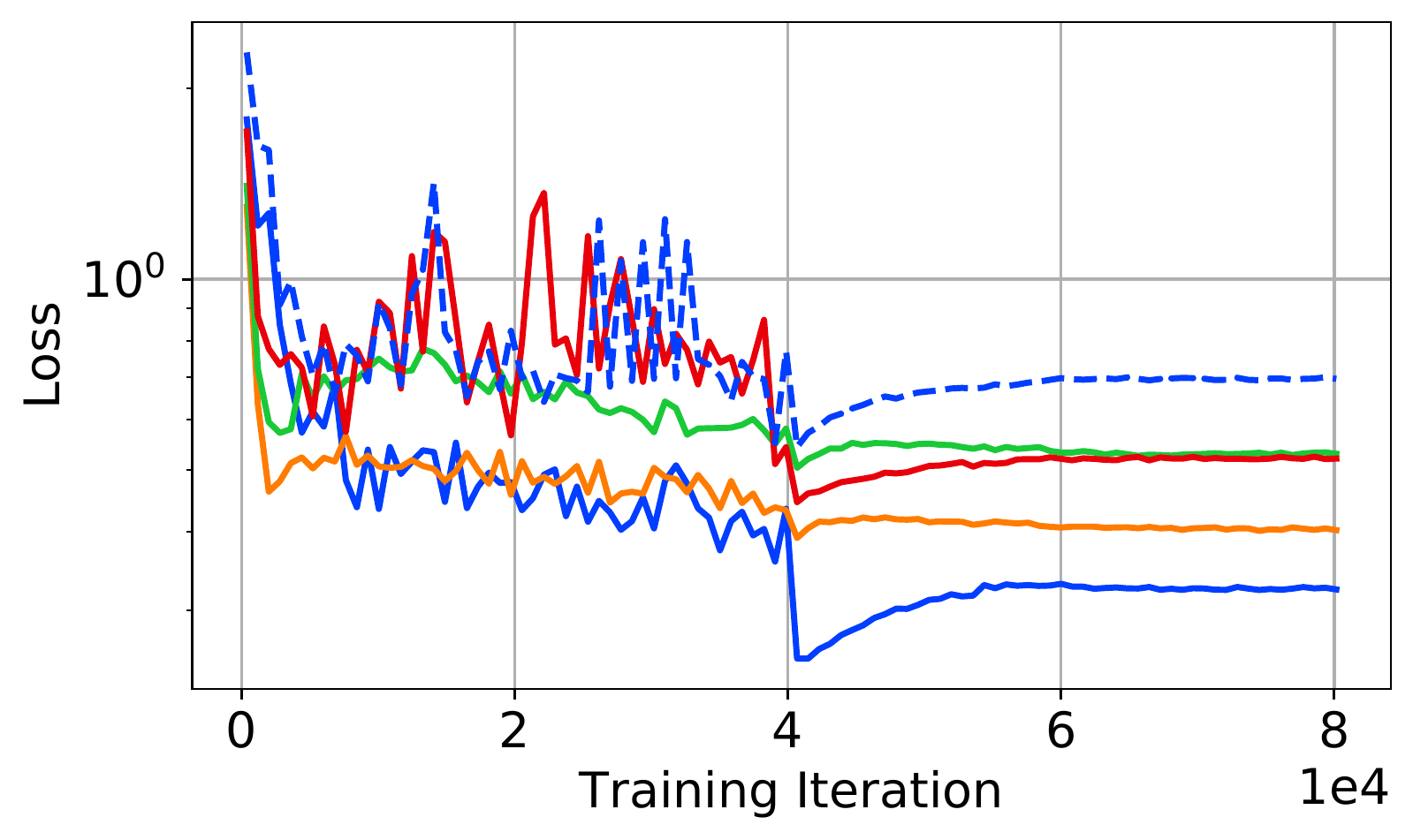}
	\caption{Training accuracy (left) and validation loss (right) on CIFAR10 for ResNet110. 
    We set the ratio of compression for DGC to be roughly the same as NUQSGD.
    In particular, we compare compression methods at compression ratio = $4/32$, \ie NUQSGD with 
    4 bits and DGC at 12.5\% compression. At this rate, both methods will have approximately the same 
    communication cost, \ie comparison in simulation is representative of real-time performance. 
    We tune the learning rate and momentum for DGC and show its best performance.}
    \label{fig:DGCsimvalloss}
\end{figure*}

\paragraph{Comparison with DGC.} 
In Figures~\ref{fig:DGCsimvar} and \ref{fig:DGCsimvalloss}, we make a comparison with DGC~\citep{DGC}.

\begin{figure}[t]
\begin{centering}
    \includegraphics[width=.5\linewidth]{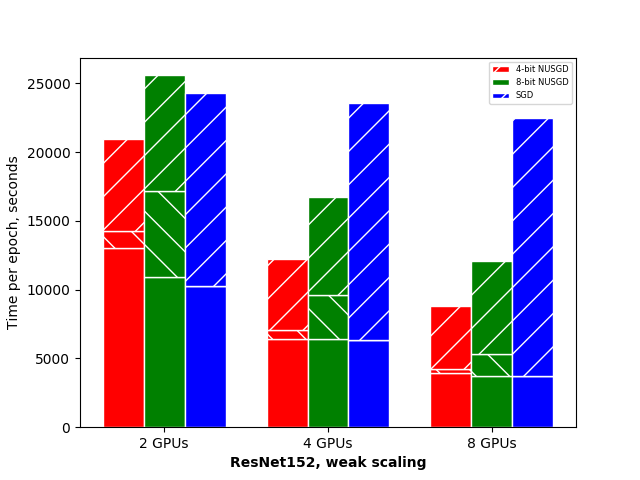}
    \caption{Scalability behavior for NUQSGD versus the full-precision baseline when training ResNet152 on ImageNet. }
    \label{fig:RN152}
\end{centering}
\end{figure}

\paragraph{ResNet152 Weak Scaling.} 
In Figure~\ref{fig:RN152}, we present the weak scaling results for ResNet152/ImageNet. 
Each of the GPUs receives a batch of size 8, and we therefore scale up the global batch size by the number of nodes. 
The results exhibit the same superior scaling behavior for NUQSGD relative to the uncompressed baseline.

\begin{figure}[t]
\begin{centering}
    \includegraphics[width=.5\linewidth]{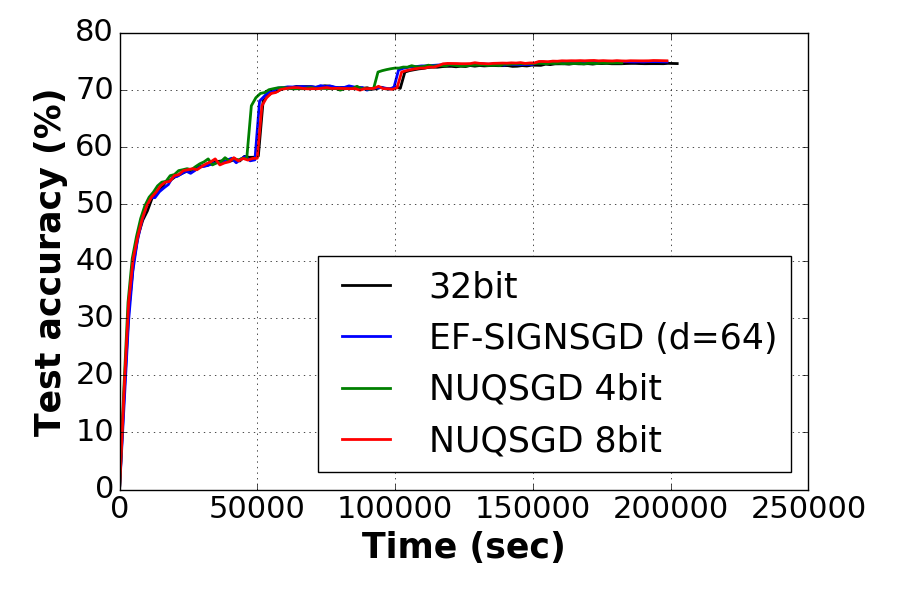}
    \caption{End-to-end training time for ResNet50/ImageNet for NUQSGD and EF-SignSGD versus the SGD baseline. }
    \label{fig:RN50_sign}
\end{centering}
\end{figure}

\paragraph{EF-SignSGD Convergence.} 
In Figure~\ref{fig:RN50_sign}, we present a performance comparison for NUQSGD variants (bucket size 512) and a convergent variant of EF-SignSGD with significant levels of parameter tuning for convergence. 
We believe this to be the first experiment to show convergence of the latter method at ImageNet scale, as the original paper only considers the CIFAR dataset. 
For convergence, we have tuned the choice of scaling factor and the granularity at which quantization is applied (bucket size). We have also considered learning rate tuning, but that did not appear to prevent divergence in the early stages of training for this model. We did not attempt warm start, since that would significantly decrease the practicality of the algorithm. 
We have found that bucket size 64 is the highest at which the algorithm will still converge on this model and dataset, and found 1-bit SGD scaling~\citep{Seide14}, which consists of taking sums over positives and over negatives for each bucket, to yield good results. 
The experiments are executed on a machine with 8 NVIDIA Titan X GPUs, and batch size 256, and can be found in Figure~\ref{fig:RN50_sign}.
Under these hyperparameter values the EF-SignSGD algorithm sends 128 bits per each bucket of 64 values (32 for each scaling factor, and 64 for the signs), doubling its baseline communication cost. 
Moreover, the GPU implementation is not as efficient, as error feedback must be computed and updated at every step, and there is less parallelism to leverage inside each bucket.  
This explains the fact that the end-to-end performance is in fact close to that of the 8-bit NUQSGD variant, and inferior to 4-bit NUQSGD.

\begin{figure*}[t]
    \includegraphics[width=.49\textwidth]{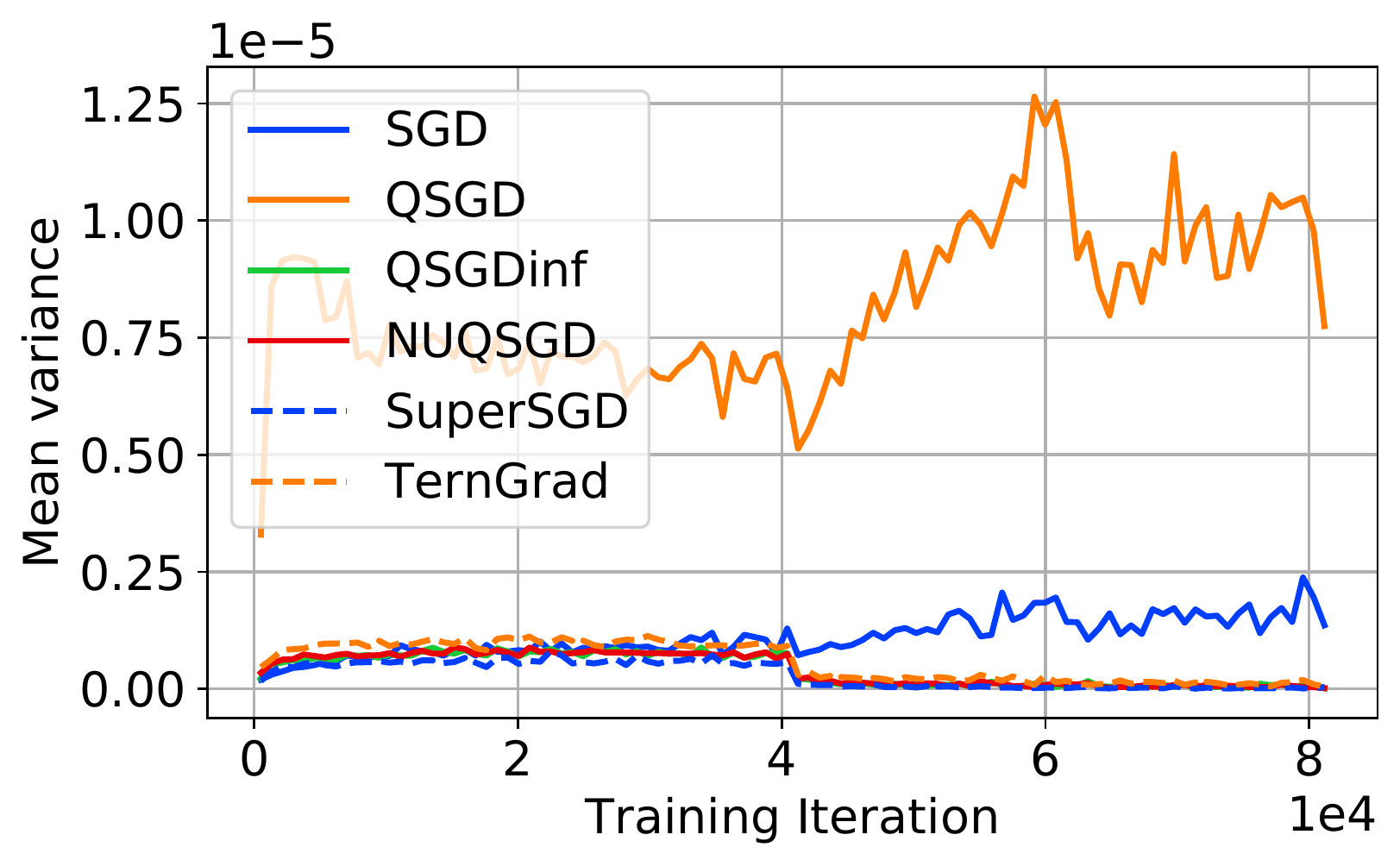}
    \hfill
    \includegraphics[width=.49\textwidth]{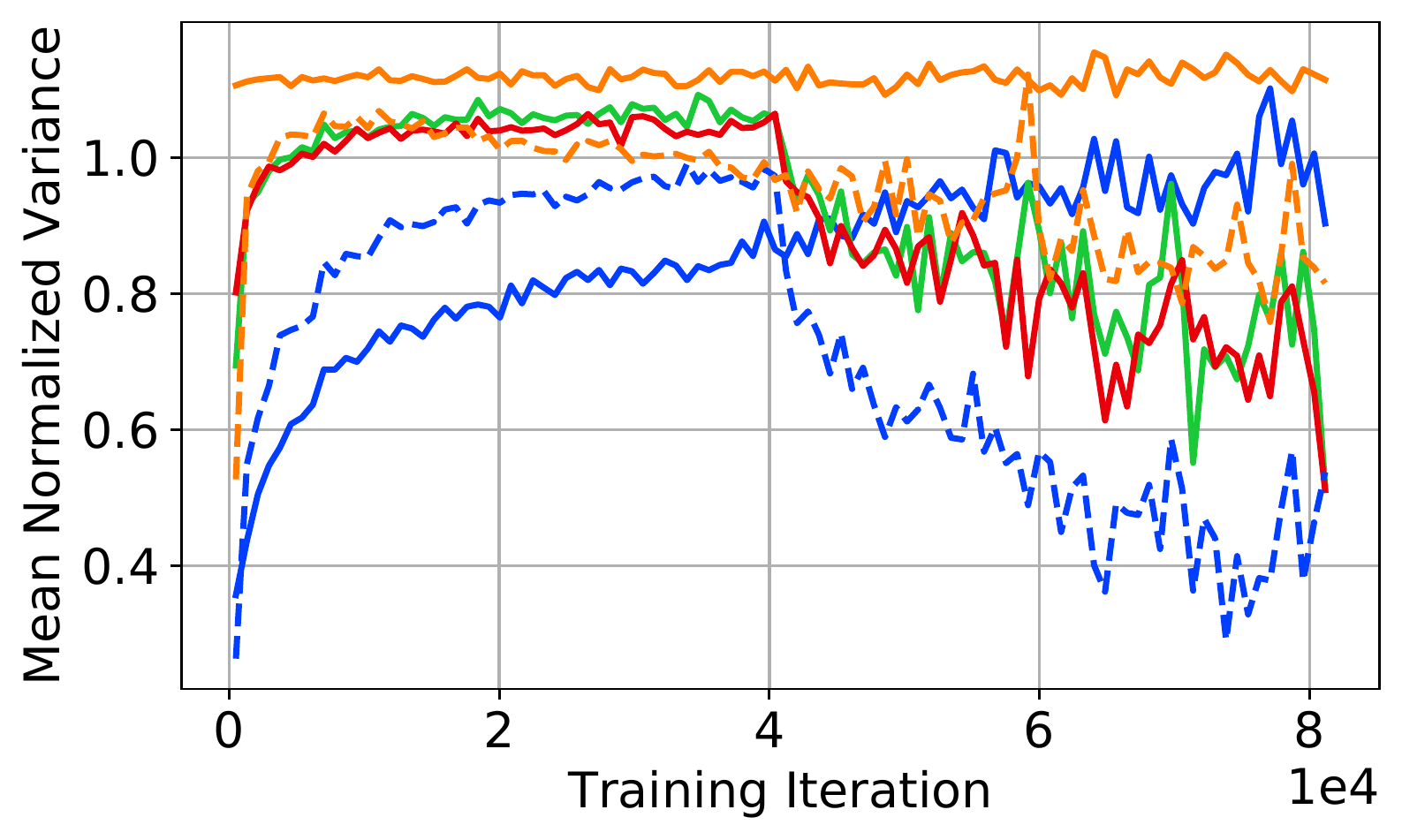}
    \caption{Estimated variance (left) and normalized variance (right) on CIFAR10 for 
    ResNet110 with mini-batch size 32 over 8 GPUs. The number of quantization bits is set to 4.}
    \label{fig:bs32var}
\end{figure*}

\begin{figure*}[t]
    \includegraphics[width=.49\textwidth]{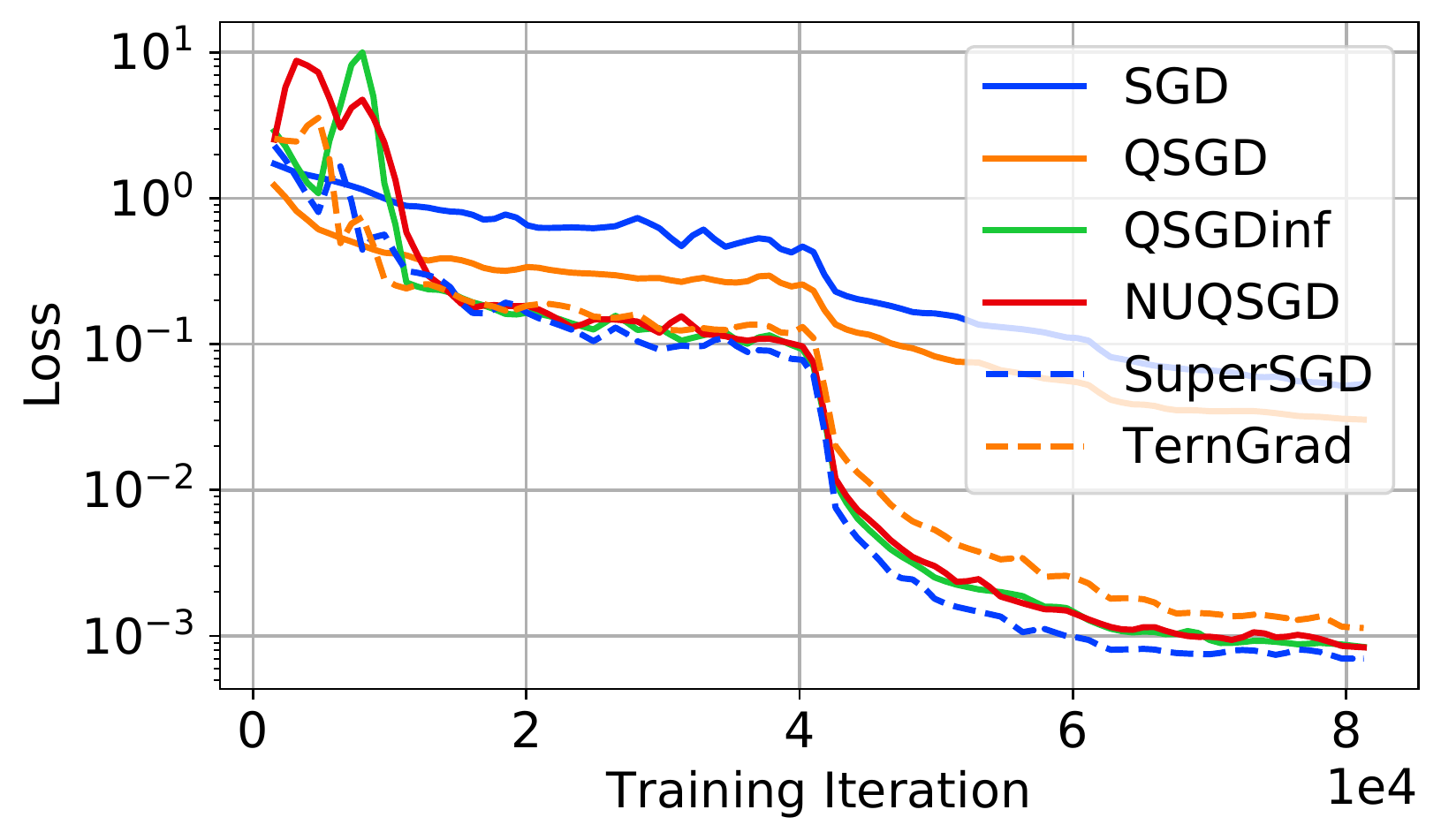}
    \hfill
    \includegraphics[width=.49\textwidth]{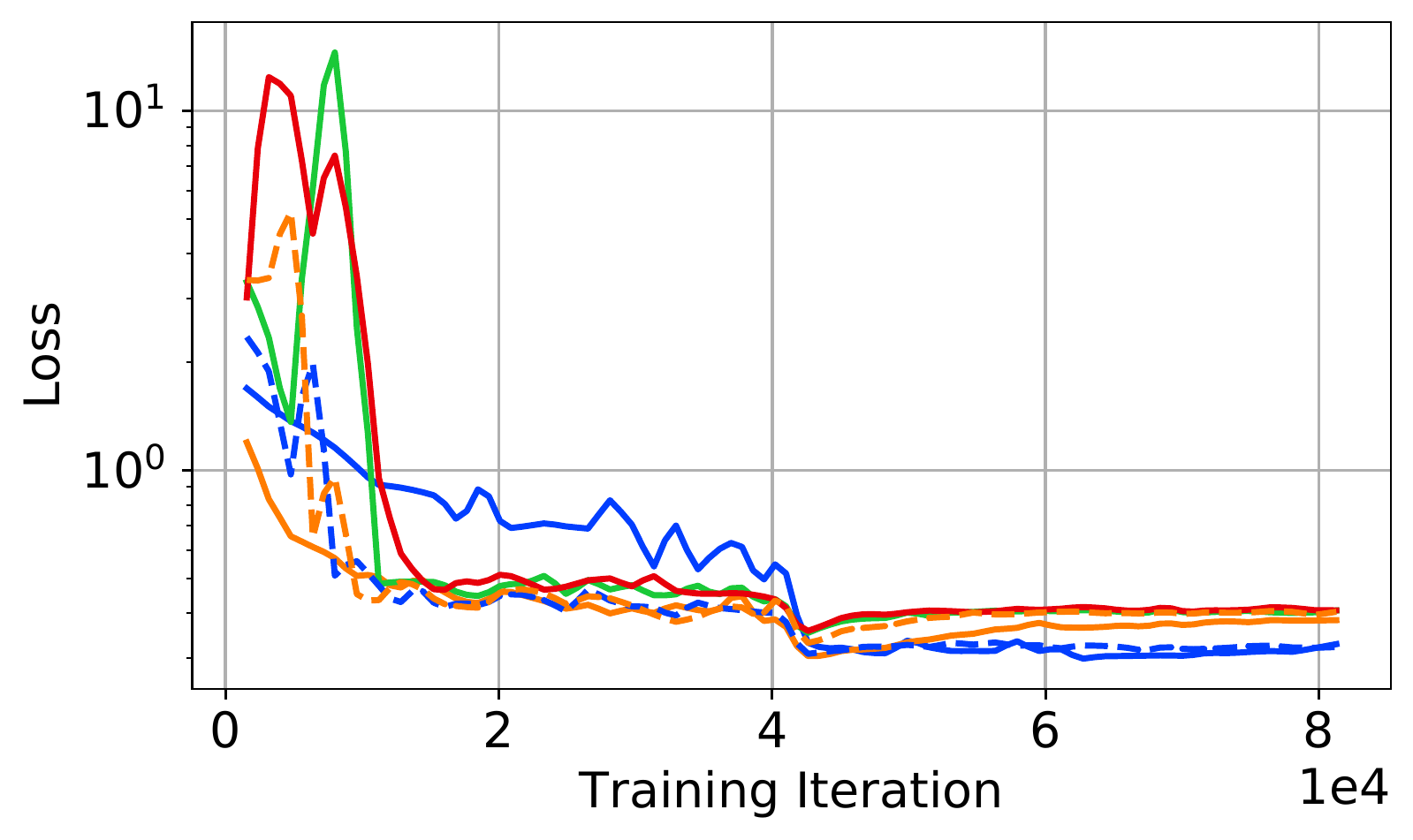}
	\caption{Training loss (left) and validation loss (right) on CIFAR10 for 
    ResNet110 with mini-batch size 32 over 8 GPUs. The number of quantization bits is set to 4.}
    \label{fig:bs32loss}
\end{figure*}

\begin{figure*}[t]
    \includegraphics[width=.49\textwidth]{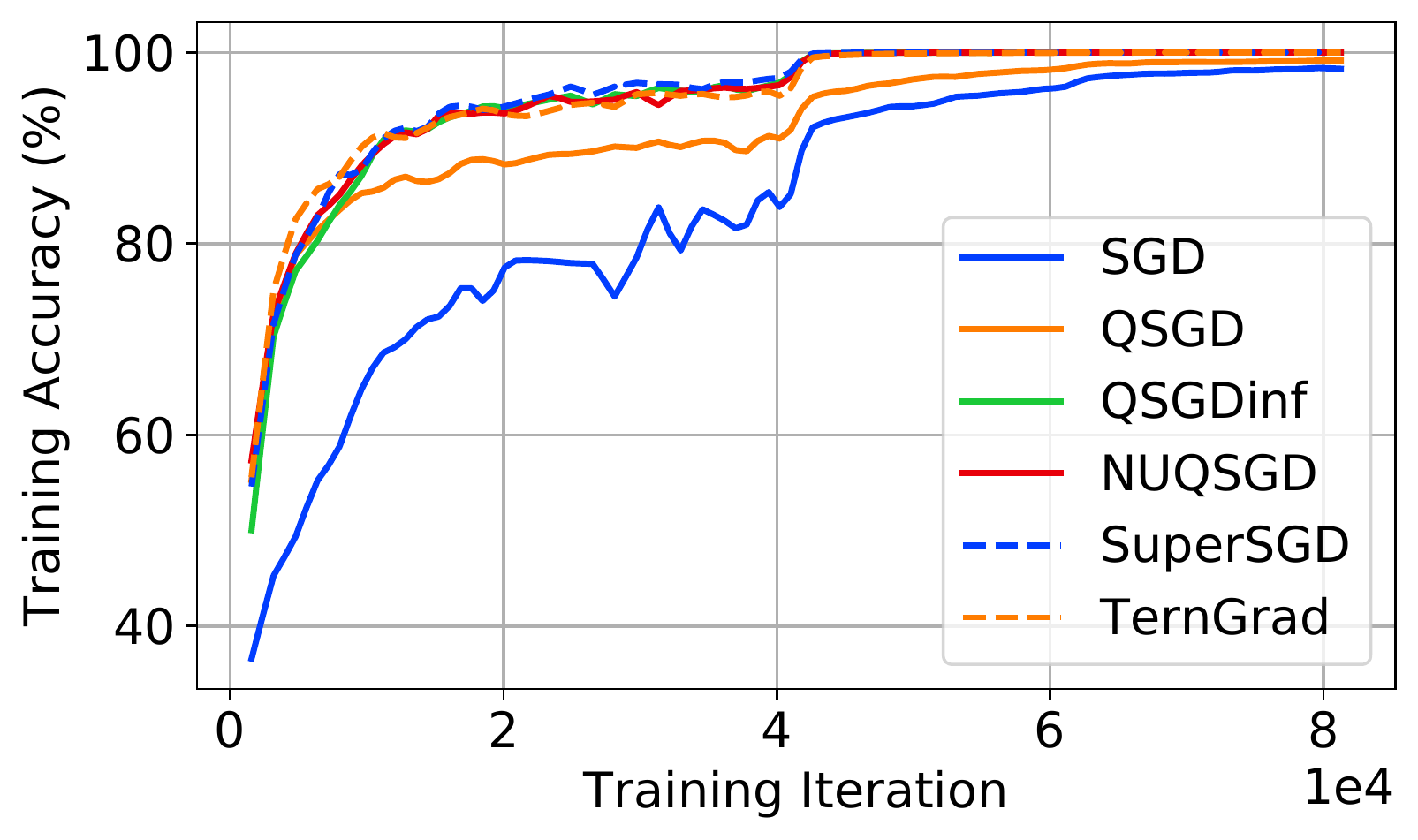}
    \hfill
    \includegraphics[width=.49\textwidth]{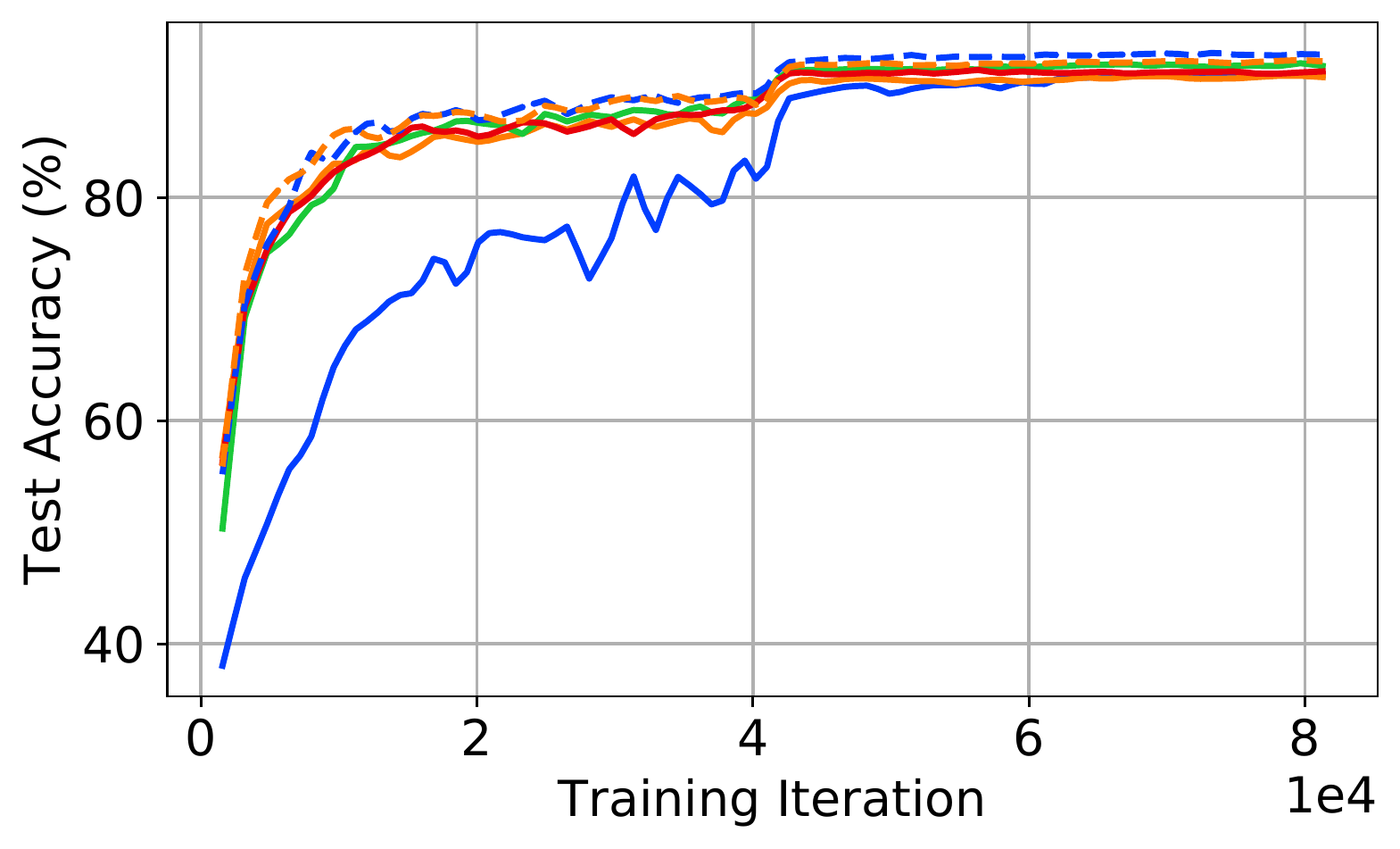}
	\caption{Training accuracy (left) and test accuracy (right) on CIFAR10 for 
    ResNet110 with mini-batch size 32 over 8 GPUs. The number of quantization bits is set to 4.}
    \label{fig:bs32acc}
\end{figure*}

\paragraph{Comparison under Small Mini-batch Size.} In Figures~\ref{fig:bs32var}, \ref{fig:bs32loss}, and \ref{fig:bs32acc}, we show the results when we train ResNet110 on CIFAR10 with mini-batch size 32 over 8 GPUs. 

The number of quantization bits is set to 4. We observe a significant gap between the variance and accuracy of QSGD with those of QSGDinf and NUQSGD. QSGDinf and NUQSGD perform similarly and slightly outperform TernGrad in this setting.
\end{appendices}
\end{document}